\documentclass[preprint,12pt]{elsarticle}

\biboptions{numbers,sort&compress}

\usepackage[T1]{fontenc}
\usepackage[utf8]{inputenc}
\usepackage{lmodern}
\usepackage{amsmath,amsfonts,amssymb,mathtools}
\usepackage{graphicx}
\usepackage{multirow}
\usepackage{booktabs}
\usepackage{array,siunitx}
\usepackage{xcolor}
\usepackage{float}
\usepackage{rotating}
\usepackage{pdflscape}
\usepackage{adjustbox}
\usepackage{enumitem}
\usepackage{footmisc}
\usepackage{caption}
\usepackage{subcaption}
\usepackage{hyperref}
\usepackage{textgreek}
\usepackage{microtype}
\usepackage{silence}
\hypersetup{
    colorlinks=true,
    linkcolor=blue,
    citecolor=blue,
    urlcolor=blue
}

\DeclareUnicodeCharacter{0394}{$\Delta$}

\journal{Computers in Biology and Medicine}

\begin{document}

\begin{frontmatter}

\title{FOSSIL: Regret-minimizing curriculum learning for metadata-free and low-data Mpox diagnosis}

\author[1]{Sahng-Min Han, Ph.D.\corref{first}}

\author[2]{Minjae Kim, Ph.D.}

\author[1]{Jinho Cha, Ph.D.\corref{cor1}}

\author[3]{Se-woon Choe, Ph.D.}

\author[4]{Eunchan Daniel Cha}

\author[5]{Jungwon Choi, M.D.}
\author[5]{Kyudong Jung, M.D.}

\cortext[first]{First author}
\cortext[cor1]{\textbf{Corresponding author.} Email: jcha@gwinnetttech.edu (Jinho Cha, Ph.D.)}

\address[1]{Computer Science Division, Gwinnett Technical College, Lawrenceville, GA, USA}
\address[2]{Department of Computer Science, University of Kentucky, Lexington, KY, USA}
\address[3]{Department of Biomedical Engineering, Kumoh National Institute of Technology, Gumi, South Korea}
\address[4]{Department of Biology, Georgia Institute of Technology, Atlanta, GA, USA}
\address[5]{Oaro Dermatology Center, Seongnam, Gyeonggi-do, South Korea}

\begin{abstract}
Deep learning in small and imbalanced biomedical datasets remains fundamentally constrained by unstable optimization and poor generalization. 
We present the first biomedical implementation of FOSSIL (Flexible Optimization via Sample-Sensitive Importance Learning), 
a regret-minimizing weighting framework that adaptively balances training emphasis according to sample difficulty. 
Using softmax-based uncertainty as a continuous measure of difficulty, we construct a four-stage curriculum (Easy–Very Hard) 
and integrate FOSSIL into both convolutional and transformer-based architectures for Mpox skin lesion diagnosis. 
Across all settings, FOSSIL substantially improves discrimination (AUC = 0.9573), calibration (ECE = 0.053), 
and robustness under real-world perturbations, outperforming conventional baselines without metadata, manual curation, or synthetic augmentation. 
The results position FOSSIL as a generalizable, data-efficient, and interpretable framework 
for difficulty-aware learning in medical imaging under data scarcity.
\end{abstract}


\end{frontmatter}

\section{Introduction}

\subsection{Clinical background and motivation}
Monkeypox (Mpox) has re-emerged as a significant public health concern, 
presenting a diagnostic challenge due to its visual similarity to other vesiculopustular diseases 
such as varicella, secondary syphilis, and eczema~\cite{huhn2005clinical, adler2022clinical, bunge2022global}. 
In telemedicine or low-resource settings, confirmatory polymerase chain reaction (PCR) testing 
and metadata such as patient demographics or lesion progression are often unavailable~\cite{thornton2023ai, von2023digital}. 
This motivates the development of metadata-free, data-efficient, and interpretable AI models 
that can support rapid screening and triage~\cite{esteva2017dermatologist, brinker2019deep, xie2023artificial}.

\subsection{Challenges in small-data medical AI}
Training robust medical AI systems under small and heterogeneous datasets remains a core obstacle~\cite{rajpurkar2022ai, kaur2022artificial}. 
Limited sample diversity causes overfitting, unreliable calibration, and loss of clinical interpretability~\cite{guo2017calibration, nixon2019measuring}. 
Common remedies such as random oversampling~\cite{chawla2002smote}, focal loss~\cite{lin2017focal}, or meta-weighting~\cite{shu2019metaweightnet}
offer empirical gains but lack theoretical guarantees, 
especially when applied to rare-disease dermatologic datasets with limited variability~\cite{sun2022skin, yang2023vision}. 

\subsection{Curriculum learning and its limitations}
Curriculum learning (CL) stabilizes optimization by introducing training samples 
in an order that progresses from easy to difficult~\cite{bengio2009curriculum}. 
In medical imaging, however, difficulty metrics are often heuristic, relying on lesion size, brightness, or expert-defined categories~\cite{gupta2020curriculum, liu2021curriculum}. 
Such task-specific heuristics limit transferability and fail to capture 
the intrinsic learning difficulty perceived by deep networks. 
A theoretically grounded weighting mechanism is therefore needed to quantify difficulty 
in a model-agnostic, data-driven manner~\cite{kumar2010self, wang2021survey}.

\subsection{FOSSIL framework and contributions}
To address these challenges, we adapt the FOSSIL framework 
(\textbf{F}lexible \textbf{O}ptimization via \textbf{S}ample-\textbf{S}ensitive \textbf{I}mportance \textbf{L}earning)~\cite{cha2025fossil}, 
which formulates a regret-minimizing weighting scheme for learning under imbalance and small-data regimes.
In this framework, each training sample receives a theoretically derived importance weight 
that decreases exponentially with estimated difficulty, 
as formalized in Section~\ref{sec:TheoreticalFoundation}. 
This weighting principle minimizes an upper bound on cumulative learning regret~\cite{hazan2016introduction, shalev2012online}, 
promoting stable convergence and improved generalization even when data are scarce or noisy.

This study makes four key contributions. 
First, it provides the first biomedical validation of the FOSSIL framework, demonstrating that a theoretically grounded, regret-minimizing weighting rule can substantially improve model performance in real-world clinical image classification without any architectural modification. 
Second, it introduces a metadata-free and interpretable curriculum design, in which sample difficulty is derived directly from softmax confidence rather than relying on manual lesion annotation or expert-defined heuristics. 
Third, the proposed method achieves enhanced generalization, calibration, and robustness, leading to smoother optimization, lower Expected Calibration Error (ECE), and improved resilience to image perturbations. 
Finally, it establishes a clinically deployable framework for teledermatology, producing transparent and computationally efficient decision-support models that can operate reliably in remote or low-resource healthcare settings.

\section{Theoretical Foundation of FOSSIL}
\label{sec:TheoreticalFoundation}

\subsection{Problem formulation}
Let $\theta$ denote model parameters and $\ell_i(\theta)$ the loss for sample $i$.  
The learner minimizes cumulative regret:
\begin{equation}
    R_T = \sum_{t=1}^{T} \big[\ell_t(\theta_t) - \ell_t(\theta^*)\big],
\end{equation}
where $\theta^*$ is the hindsight optimum.

\subsection{Regret-minimizing weighting}
FOSSIL assigns weights to minimize an upper bound on $R_T$:
\begin{equation}
    w_i = \exp\!\left(-\frac{d_i}{T}\right),
\end{equation}
with $d_i$ as model-estimated difficulty and $T$ controlling exploration–exploitation balance.

\subsection{Weighted objective function}
The weighted empirical loss is:
\begin{equation}
    \mathcal{L}_{\text{FOSSIL}}(\theta) = \frac{1}{N} \sum_{i=1}^N w_i \, \ell_i(\theta),
\end{equation}
which unifies focal loss, meta-weighting, and curriculum learning into a single regret-minimizing form.

\subsection{Theoretical properties}

\subsubsection{Stability and generalization}
The proposed regret-minimizing weighting scheme smooths gradient variance and ensures consistent convergence 
under imbalance or noise, providing stable training dynamics and improved generalization across small-data regimes.

\subsubsection{Comparison to prior methods}

Table~\ref{tab:comparison_methods} summarizes conceptual differences between representative weighting strategies.
Focal Loss emphasizes hard examples through a tunable focusing parameter $\gamma$ but lacks theoretical guarantees.
Meta-weighting adapts sample weights dynamically based on loss magnitude, yet suffers from instability in small-data regimes.
Traditional curriculum learning orders samples by difficulty without providing regret or convergence guarantees.
In contrast, the proposed regret-minimizing weighting integrates these ideas into a single unified formulation that
ensures bounded weights, monotonic curriculum progression, and provable stability.

\begin{table}[H]
\centering
\scriptsize
\caption{Comparison with related weighting strategies.}
\label{tab:comparison_methods}
\begin{tabular}{lll}
\toprule
Method & Formula & Limitation \\
\midrule
Focal Loss & $(1-p_i)^\gamma \ell_i$ & Requires $\gamma$, no regret control \\
Meta-weighting & $w_i=f(\ell_i)$ & Unstable under small data \\
Curriculum Learning & Ordered sampling & No theoretical grounding \\
Regret-minimizing weighting & $e^{-d_i/T}$ & Unified and theoretically grounded \\
\bottomrule
\end{tabular}
\end{table}

\section*{Materials and Methods}

\subsection*{Dataset and Preprocessing}
The primary dataset used for model training and internal validation was the publicly available \textit{Mpox Skin Lesion Dataset Version 2.0 (MSLD v2)}~\cite{hasan2023msldv2}, 
released by the mHealth Buet research group at Bangladesh University of Engineering and Technology (BUET). 
This dataset contains six dermatologist-verified skin lesion categories (Mpox, Cowpox, Chickenpox, Measles, Hand-Foot-Mouth Disease, and Healthy), 
and provides an expanded and clinically validated benchmark for Mpox recognition. 
A total of 804 cropped and de-identified clinical photographs were utilized in this study. 

For external validation, we employed the \textit{Mpox Close Skin Images (MCSI)} dataset from Kaggle~\cite{kaggle2023mcsi}, 
comprising 228 high-resolution smartphone-acquired images (102 Mpox-positive and 126 non-Mpox) that closely represent real-world teledermatology scenarios. 
Negative samples in MCSI include Chickenpox, Measles, Eczema, and other visually similar dermatological lesions, ensuring realistic diagnostic diversity.

All images from both datasets were visually inspected to ensure integrity and resized to $224\times224$ pixels using bilinear interpolation without padding or cropping. 
Pixel intensities were normalized according to ImageNet statistics, and all data were loaded via the \texttt{torchvision.datasets.ImageFolder} interface. 
No data augmentation was applied during standard training. 
Augmentations (horizontal flip, color jitter, and center crop) were introduced only during robustness testing to simulate acquisition variability under smartphone-based or low-light conditions.

All experiments were conducted using PyTorch~2.9 (built from source with CUDA~12.8) and \texttt{torchvision}~0.24, 
with cuDNN~8.9 running on an NVIDIA RTX~5090 GPU (32~GB VRAM) under Ubuntu~22.04.5~LTS (WSL2). 
Mixed-precision training with PyTorch Automatic Mixed Precision (AMP) was employed to improve computational efficiency.
y.

\subsection*{Difficulty Estimation and Curriculum Construction}

\paragraph{Preliminary entropy-based analysis.}
As an initial exploration, we evaluated entropy-based difficulty 
\[
H(x_i) = -\sum_c p_i^{(c)} \log p_i^{(c)},
\]
on the Mpox-positive subset ($n=102$) using a pretrained DenseNet121 model. 
While entropy successfully captured diagnostic ambiguity, it yielded skewed and fold-dependent distributions, 
leading to unstable curriculum boundaries. 
Detailed results of this preliminary experiment are provided in Appendix~A.2.
To ensure consistent stratification, we adopted a simplified and more robust softmax-based formulation.

\paragraph{Softmax-based difficulty estimation.}
Final difficulty scores were computed using the softmax-based uncertainty metric:
\[
d_i = 1 - \max_c p_i^{(c)},
\]
where $p_i^{(c)}$ denotes the predicted softmax probability for class~$c$. 
This measure provides a class-agnostic scalar of model confidence, 
allowing difficulty to be estimated directly from prediction distributions. 
Empirical quantiles (25th, 50th, and 75th) of $d_i$ defined four curriculum stages: 
\textit{Easy}, \textit{Medium}, \textit{Hard}, and \textit{Very Hard}.

A Mann–Whitney~U test ($U = 6727$, $p = 0.544$) confirmed the absence of class bias in difficulty estimation, 
and pairwise stage comparisons ($p < 0.0001$) verified complete ordinal separation of stages. 
Table~\ref{tab:difficulty_stats} summarizes these statistics, showing similar means and medians across classes,
indicating class-independent difficulty estimation.

\begin{table}[H]
\centering
\footnotesize
\caption{\textbf{Summary statistics of softmax-based difficulty scores by class.} Similar means and medians indicate class-independent difficulty estimation.}
\label{tab:difficulty_stats}
\begin{tabular}{lccccccc}
\toprule
\textbf{Class} & \textbf{Count} & \textbf{Mean} & \textbf{Std} & \textbf{Min} & \textbf{25\%} & \textbf{Median} & \textbf{Max} \\
\midrule
Negative (0) & 126 & 0.605 & 0.249 & 0.013 & 0.441 & 0.657 & 0.963 \\
Positive (1) & 102 & 0.625 & 0.242 & 0.013 & 0.475 & 0.686 & 0.944 \\
\bottomrule
\end{tabular}
\end{table}

\begin{figure}[H]
    \centering
    \includegraphics[width=\textwidth]{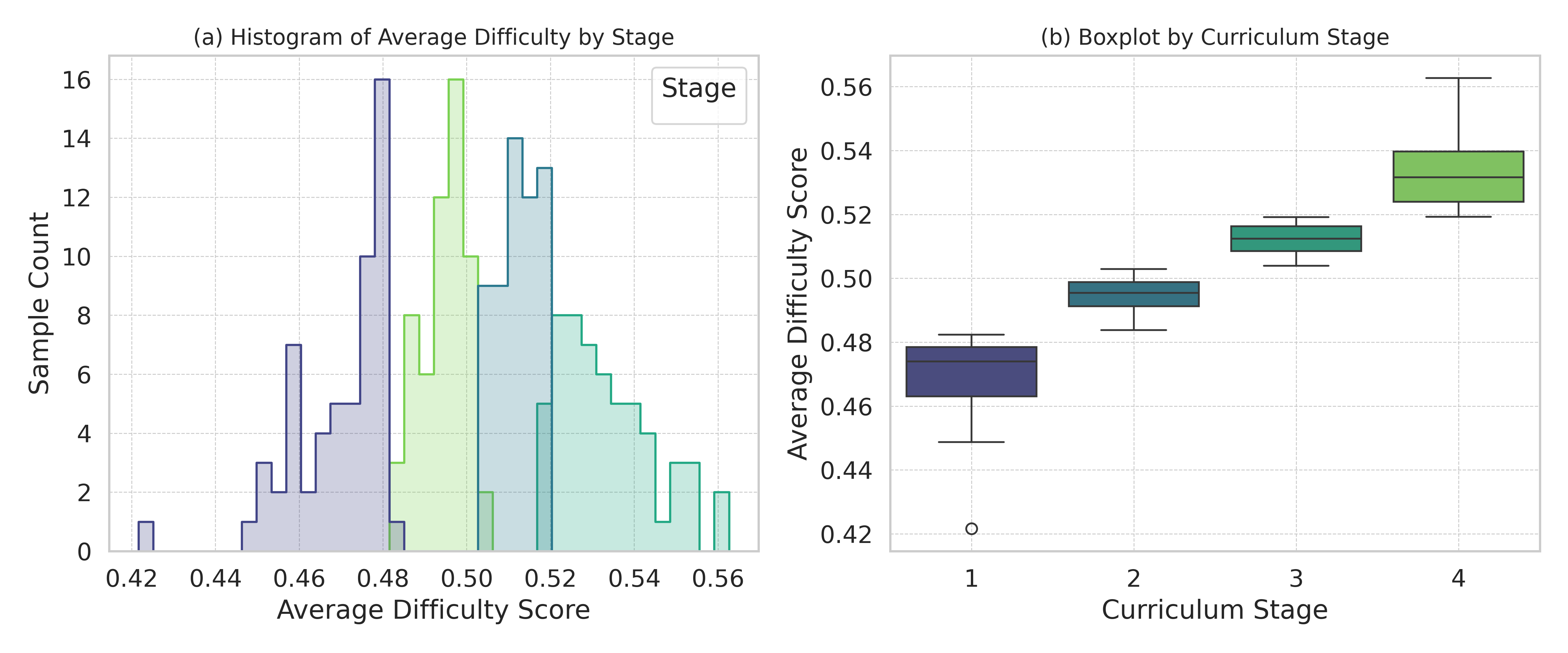}
    \caption{
    Distribution of softmax-based difficulty scores across the four curriculum stages. 
    (a) Histogram and (b) boxplot representations demonstrate clear separation between quantile-defined stages 
    (\textit{Easy}, \textit{Medium}, \textit{Hard}, and \textit{Very Hard}), 
    confirming the robustness of the stratification procedure.
    }
    \label{fig:difficulty_curriculum_combined}
\end{figure}

\paragraph{Validation of stage separation.}
To verify that the quantile-based difficulty stages were statistically distinct, 
we conducted pairwise Mann–Whitney~U tests on the softmax-derived difficulty scores. 
Figure~\ref{fig:difficulty_curriculum_combined} illustrates the distribution of scores across the four curriculum stages, 
showing clear separation in both histogram and boxplot representations. 
Each stage contained an equal number of samples ($n=57$), ensuring balanced progression across the learning curriculum.
As summarized in Table~\ref{tab:stage_difficulty_comparison}, 
all pairwise comparisons were statistically significant ($p<0.0001$), 
confirming robust ordinal stratification and validating the effectiveness of the quantile-based curriculum design.

\begin{table}[H]
\centering
\footnotesize
\caption{
Pairwise Mann–Whitney~U test results for softmax-based difficulty scores across curriculum stages.
All comparisons were statistically significant ($p<0.0001$), confirming complete ordinal separation between difficulty levels.
}
\label{tab:stage_difficulty_comparison}
\begin{tabular}{lll}
\toprule
\textbf{Comparison} & \textbf{U Statistic} & \textbf{p-value} \\
\midrule
Easy vs Medium / Hard / Very Hard & 3249.0 & $<0.0001$ \\
Medium vs Hard / Very Hard        & 3249.0 & $<0.0001$ \\
Hard vs Very Hard                 & 0.0    & $<0.0001$ \\
\bottomrule
\end{tabular}
\end{table}

\subsection*{Cross-Validation and Fold Setup}

To ensure robust and unbiased evaluation, we employed stratified five-fold cross-validation, 
preserving class balance within each split. 
Difficulty scores were computed exclusively on the training partition of each fold to prevent information leakage, 
and curriculum quantiles were recalculated locally to maintain fold-specific consistency. 
Each curriculum stage contained an equal number of samples ($n=57$), 
guaranteeing balanced progression across difficulty levels. 
A summary of fold-level distributions is provided in Appendix~A.2.

This procedure enabled statistically independent evaluation of model generalization while 
allowing stage-wise adaptation of the proposed weighting framework to local training dynamics.

\subsection*{Model Architectures and Curriculum Strategies}

To capture architectural diversity, twelve pretrained deep learning models were benchmarked and 
organized into five structural categories: 
Lightweight CNNs, Standard CNNs, Efficient CNNs, Vision Transformers, and Hybrid CNN–Transformers. 
The architectural taxonomy and model grouping are summarized in Table~\ref{tab:model_groups}. 
From this pool, six representative models—DenseNet121, ConvNeXt-T, DeiT-S, MaxViT-T, VGG16\_bn, and MobileNetV2—were 
selected for detailed analysis under the proposed curriculum learning framework.

\begin{table}[H]
\centering
\footnotesize
\caption{
Architectural grouping of the twelve evaluated models.
Group labels (G1–G5) correspond to increasing architectural complexity from lightweight to hybrid designs.
}
\label{tab:model_groups}
\begin{tabular}{ll}
\toprule
\textbf{Group} & \textbf{Models} \\
\midrule
G1 – Lightweight CNNs         & MobileNetV2, SqueezeNet1.1 \\
G2 – Standard CNNs            & ResNet18, DenseNet121, VGG16\_bn \\
G3 – Efficient CNNs           & EfficientNet-B3, EfficientNet-B4 \\
G4 – Vision Transformers      & ViT-B/16, Swin-T, DeiT-S \\
G5 – Hybrid CNN–Transformers  & ConvNeXt-T, MaxViT-T \\
\bottomrule
\end{tabular}
\end{table}

\begin{figure}[H]
    \centering
    \includegraphics[width=0.5\linewidth]{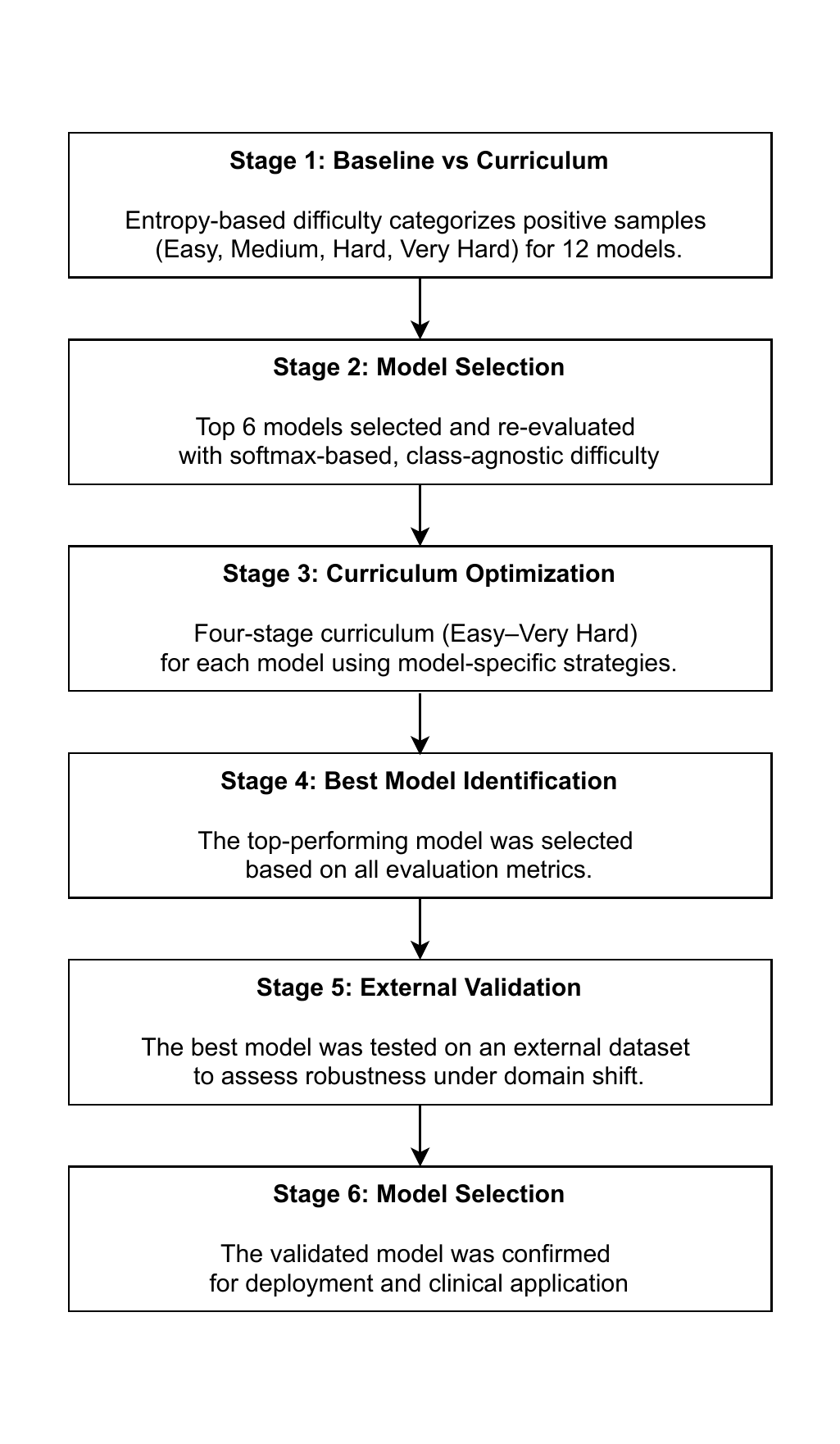}
    \caption{
    Overview of model selection and curriculum learning stages for Mpox diagnosis. 
    The workflow integrates diverse architectures under a unified weighting scheme.
    }
    \label{fig:model_selection_flow}
\end{figure}

Figure~\ref{fig:model_selection_flow} illustrates the overall workflow of model selection and curriculum-based training, 
highlighting the integration of diverse architectures within a unified weighting scheme for Mpox diagnosis. 

All selected models were trained under the regret-minimizing weighting objective:
\begin{equation}
\mathcal{L}_{\text{train}} = 
\frac{1}{N}\sum_i 
\exp\!\left(-\frac{d_i}{T}\right)\ell_i(\theta),
\end{equation}
where $\ell_i(\theta)$ denotes the binary cross-entropy loss and $T$ governs the exploration–exploitation balance 
between easy and hard samples. 

Convolutional models (DenseNet121, ConvNeXt-T, VGG16\_bn, MobileNetV2) incorporated partial backbone freezing 
to stabilize early training, 
whereas transformer-based models (DeiT-S, MaxViT-T) employed Sharpness-Aware Minimization (SAM) 
to promote flatter minima and improved generalization. 
Model-specific curriculum strategies, including the use of freezing and SAM, 
are summarized in Table~\ref{tab:model_strategies}.

\begin{table}[H]
\centering
\footnotesize
\caption{
Model-specific curriculum strategies. 
All models employed the proposed weighting scheme, with SAM and freezing selectively applied 
to prevent overfitting and enhance convergence stability.
}
\label{tab:model_strategies}
\begin{tabular}{p{3.2cm}ccc}
\toprule
\textbf{Model} & \textbf{Weighting Scheme} & \textbf{SAM} & \textbf{Freezing} \\
\midrule
ConvNeXt-T     & Strongly applied & Not used     & Applied  \\
DenseNet121    & Strongly applied & Not used     & Applied  \\
DeiT-S         & Applied          & Strongly applied & Not used \\
MaxViT-T       & Applied          & Strongly applied & Not used \\
VGG16\_bn      & Strongly applied & Not used     & Applied  \\
MobileNetV2    & Applied          & Not used     & Applied  \\
\bottomrule
\end{tabular}
\end{table}

All experiments were performed using the Adam optimizer (learning rate $1\times10^{-4}$, batch size = 16) 
with early stopping based on validation AUC. 
Final performance metrics were averaged over folds to yield statistically robust estimates.

\subsection*{Evaluation Metrics and Robustness Testing}
Model performance was evaluated using AUC, Accuracy, Sensitivity, Specificity, Precision, F1-score, and Expected Calibration Error (ECE)~\cite{guo2017calibration}.  
ECE was computed via 15-bin histogram calibration.  
To assess robustness, each trained model was tested under five image degradations—Gaussian blur, JPEG compression, additive noise, brightness, and contrast adjustments—each at three severity levels.  
Performance degradation (ΔAUC) quantified resilience to domain shifts.

\subsection*{External Validation}

For external validation, we used the \textit{Mpox Close Skin Images (MCSI)} dataset 
(400 images: 100 Mpox, 300 non-Mpox)~\cite{msci_dataset}.  
A balanced subset of 200 images (100 per class) was used for zero-shot testing without fine-tuning.  
Negative samples (normal skin, chickenpox, and acne) provided realistic clinical variability, 
simulating a teledermatology screening scenario where new lesion types appear at inference time.  
The composition of this external validation subset is summarized in Table~\ref{tab:mcsi_composition}.

\begin{table}[H]
\centering
\footnotesize
\caption{
Composition of the balanced external validation subset (MCSI dataset). 
Each category represents a clinically distinct lesion type, ensuring realistic variability 
for out-of-distribution generalization assessment.
}
\label{tab:mcsi_composition}
\begin{tabular}{lcc}
\toprule
\textbf{Category} & \textbf{Count} & \textbf{Description} \\
\midrule
Mpox (positive)        & 100 & Laboratory-confirmed Mpox lesions \\
Normal skin            & 25  & Healthy control images \\
Chickenpox             & 25  & Vesicular rash with clinical overlap \\
Acne\_L1               & 17  & Mild acneiform eruptions \\
Acne\_L2               & 17  & Moderate papular acne \\
Acne\_L3               & 16  & Severe cystic acne \\
\midrule
\textbf{Total}         & \textbf{200} &  \\
\bottomrule
\end{tabular}
\end{table}

\subsection*{Implementation Details}
All code was implemented in Python 3.10 using PyTorch Lightning.  
Random seeds (42, 77, 123) ensured reproducibility.  
Model checkpoints, stage quantiles, and evaluation scripts are publicly available upon request.

\section{Results}
\label{sec:Results}

\subsection*{Initial Model Selection and Comparison}

To ensure architectural diversity and clinical relevance, we benchmarked a total of twelve deep learning models 
spanning five architectural families, as summarized in Table~\ref{tab:model_groups}. 
Each model was trained under both baseline and curriculum learning settings using stratified five-fold cross-validation. 
Comprehensive quantitative results for all models, including AUC, accuracy, sensitivity, specificity, and overfitting status, 
are summarized in Table~\ref{tab:sup_model_comparison}. 
Group-wise comparisons enabled representative selection while avoiding architectural redundancy.

From each architectural group, the best-performing model was selected based on average AUC and sensitivity. 
These five representative models—together with an additional prompt-tuned Vision Transformer—were carried forward 
for downstream evaluation, including interpretability, robustness, and clinical applicability. 
Prompt tuning consistently improved transformer performance while maintaining low parameter overhead.

As shown in Table~\ref{tab:sup_model_comparison}, curriculum learning yielded consistent performance gains across 
most model families. For example, DenseNet121 exhibited a clear AUC improvement 
(0.895~$\rightarrow$~0.915) and increased specificity without compromising sensitivity, 
demonstrating the effectiveness of difficulty-aware training in small-data regimes. 
Similarly, ConvNeXt-T achieved the highest AUC (0.9285) and exhibited no signs of overfitting, 
highlighting its robustness and stability under the proposed curriculum design.

To support rigorous model comparison, overfitting behavior was systematically annotated based on training dynamics. 
Models displaying early saturation in training accuracy coupled with large train–validation gaps were labeled as 
clearly overfitting, whereas smooth and consistent curves indicated minimal overfitting. 
Borderline cases exhibited fold-specific instability such as fluctuating validation accuracy or stagnating loss.

\begin{figure}[H]
    \centering
    \begin{subfigure}[t]{0.24\textwidth}
        \centering
        \includegraphics[width=\textwidth]{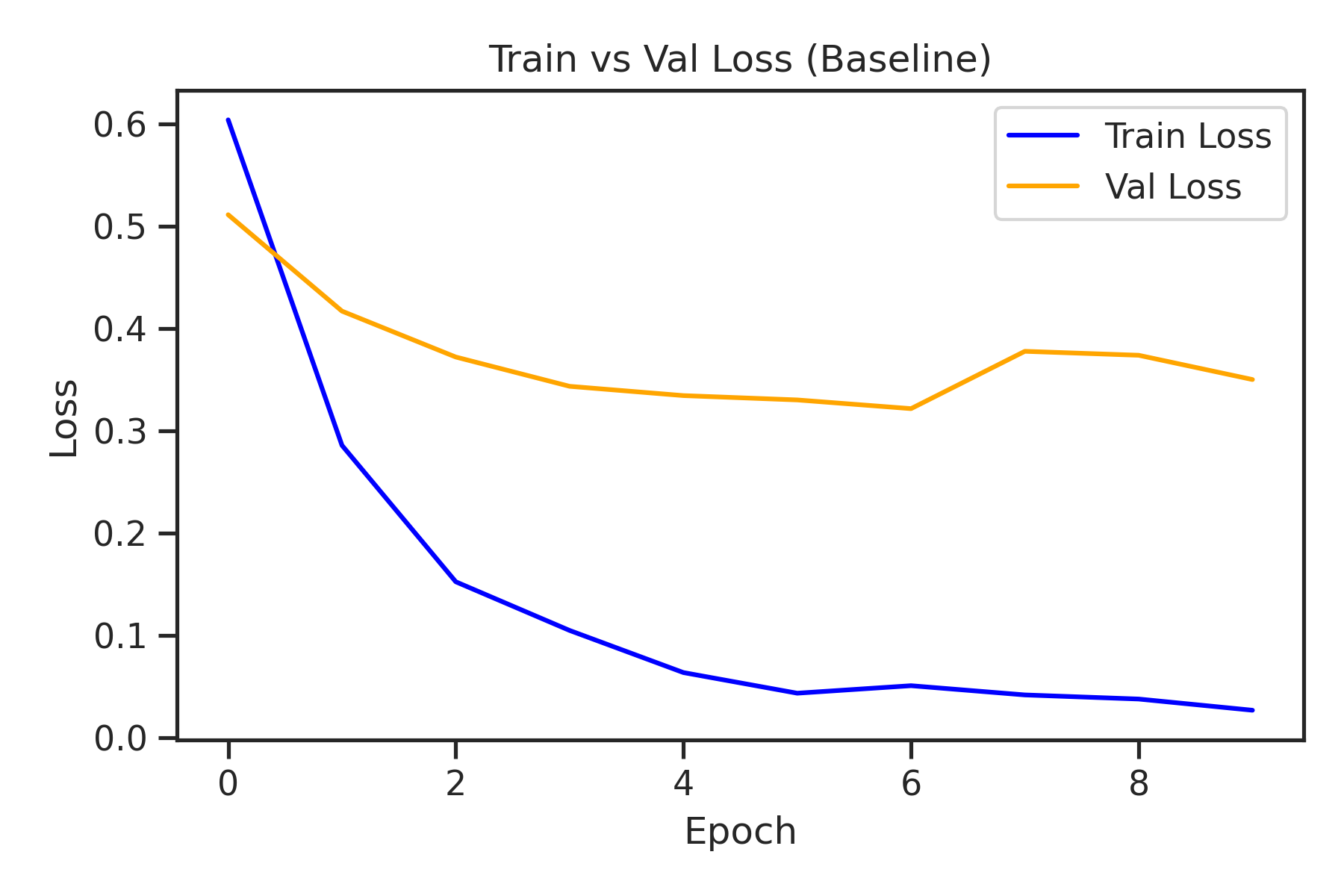}
        \caption{DenseNet121 – Baseline Loss}
    \end{subfigure}
    \hfill
    \begin{subfigure}[t]{0.24\textwidth}
        \centering
        \includegraphics[width=\textwidth]{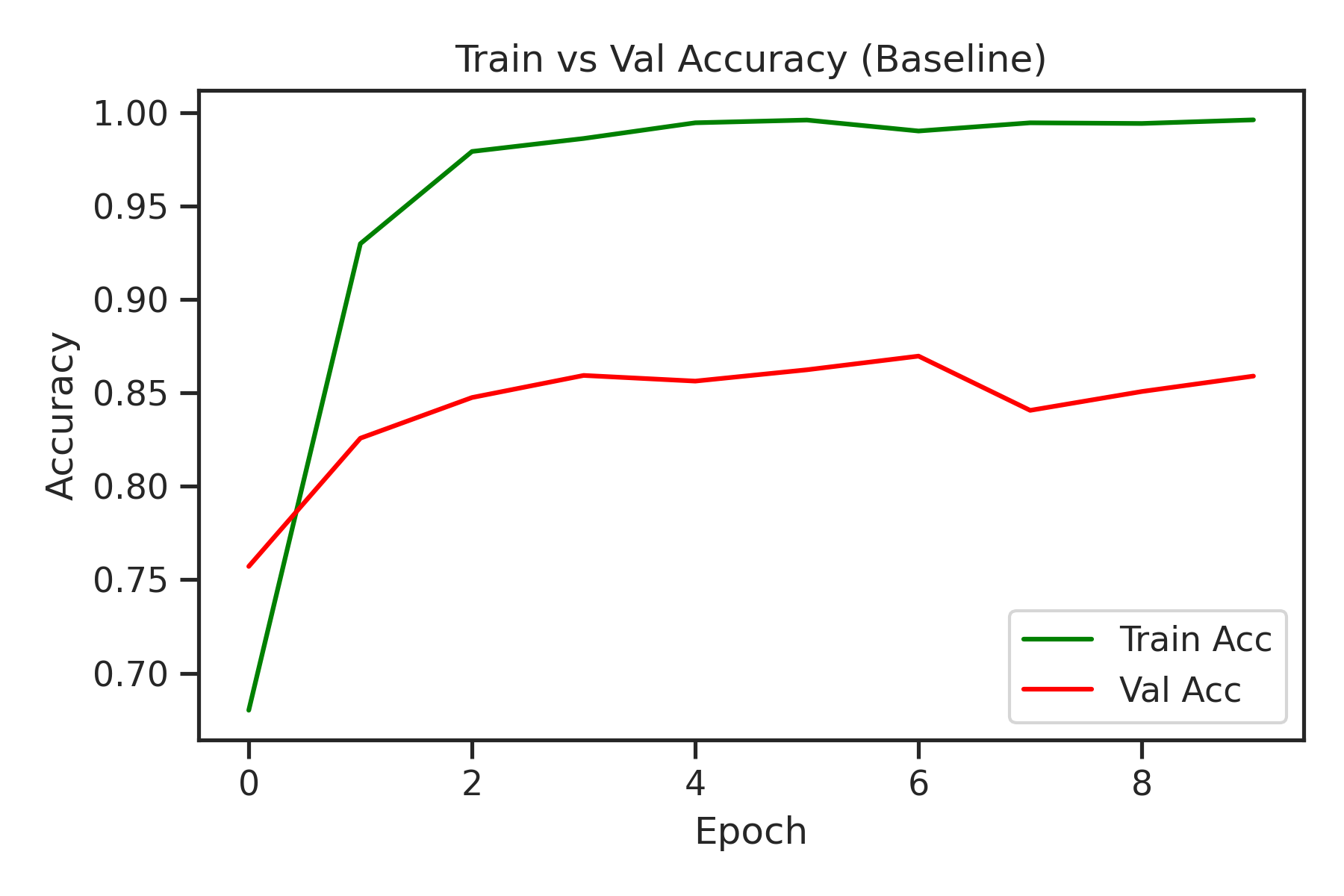}
        \caption{DenseNet121 – Baseline Accuracy}
    \end{subfigure}
    \hfill
    \begin{subfigure}[t]{0.24\textwidth}
        \centering
        \includegraphics[width=\textwidth]{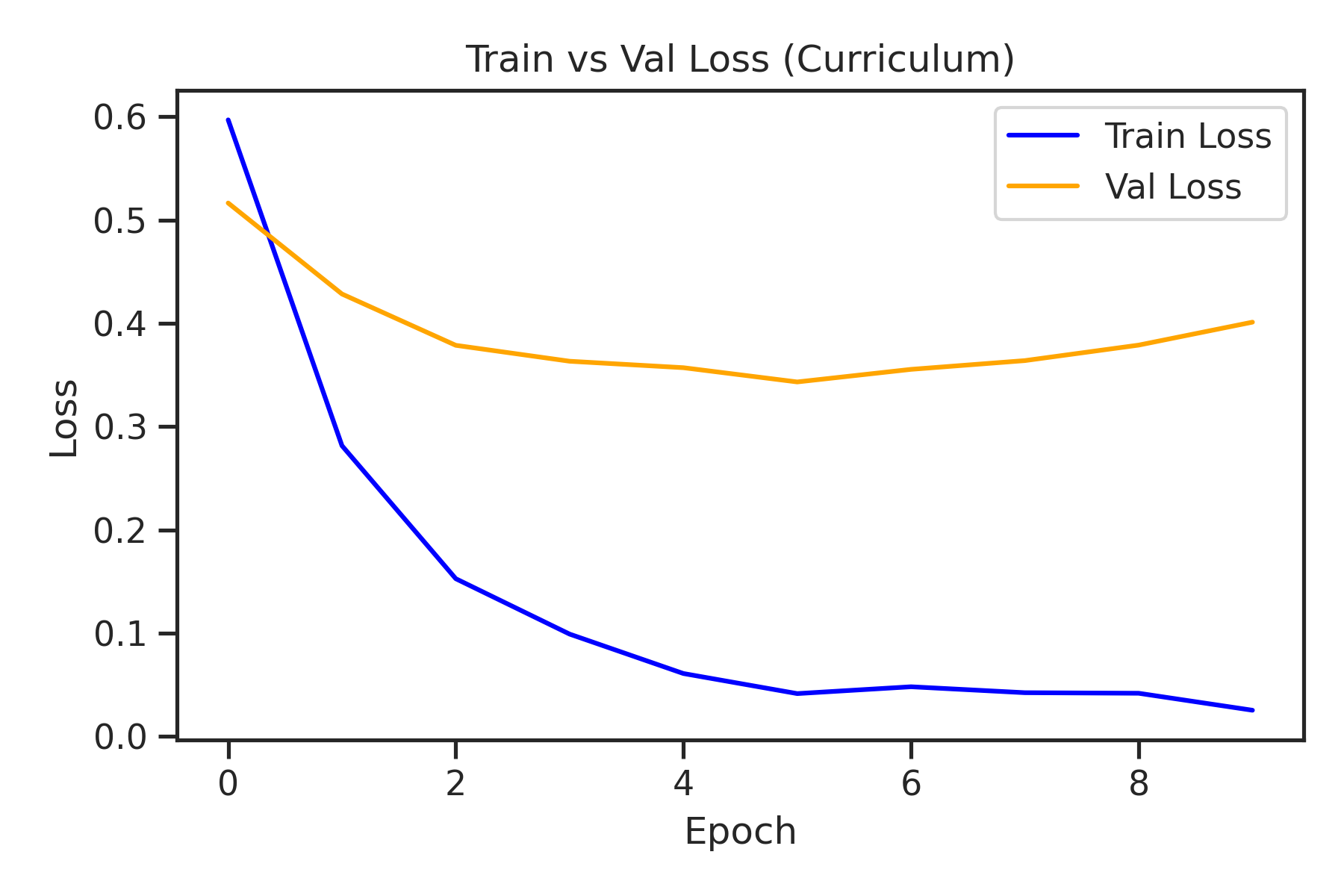}
        \caption{DenseNet121 – Curriculum Loss}
    \end{subfigure}
    \hfill
    \begin{subfigure}[t]{0.24\textwidth}
        \centering
        \includegraphics[width=\textwidth]{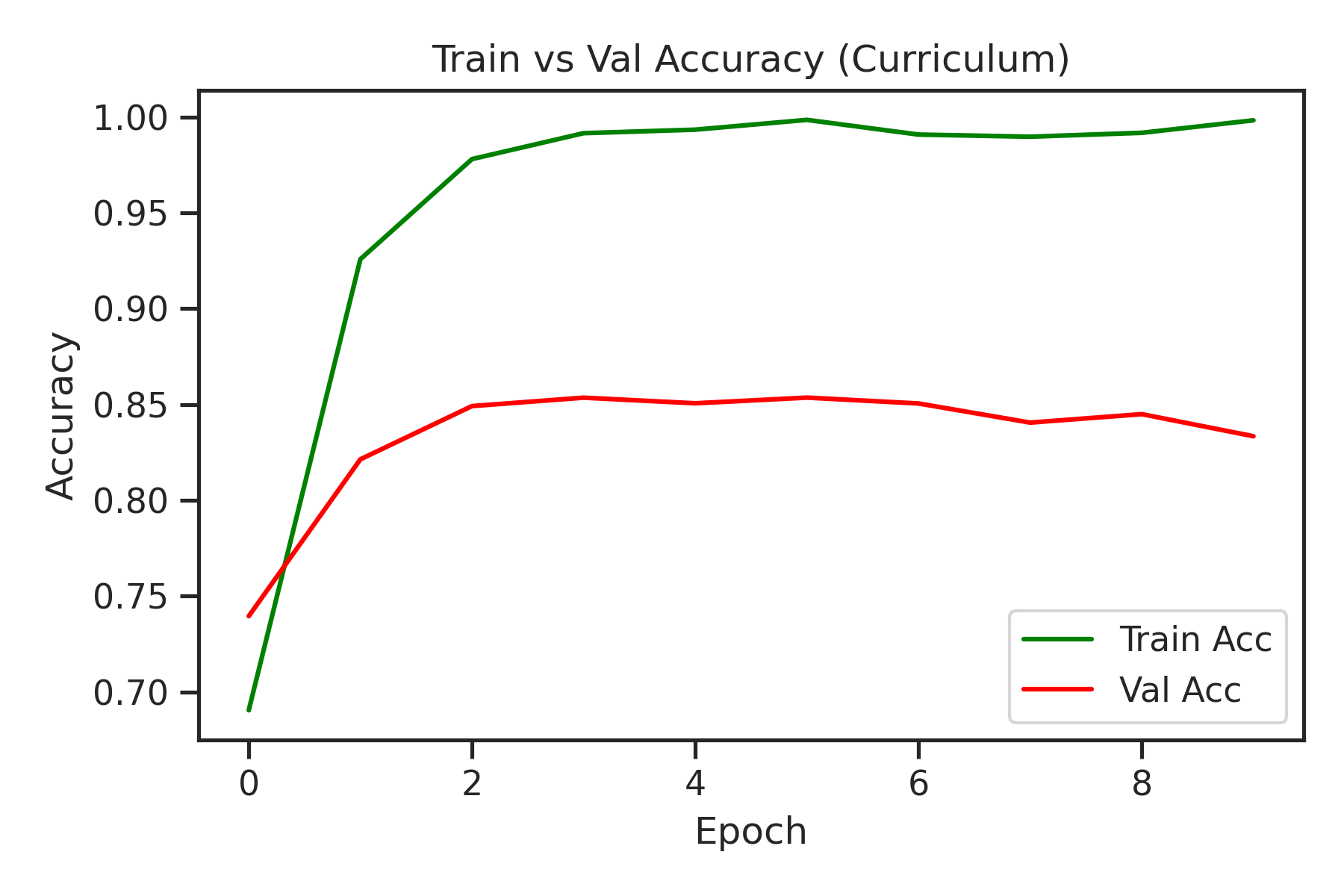}
        \caption{DenseNet121 – Curriculum Accuracy}
    \end{subfigure}
    
    \vspace{0.5em}
    
    \begin{subfigure}[t]{0.24\textwidth}
        \centering
        \includegraphics[width=\textwidth]{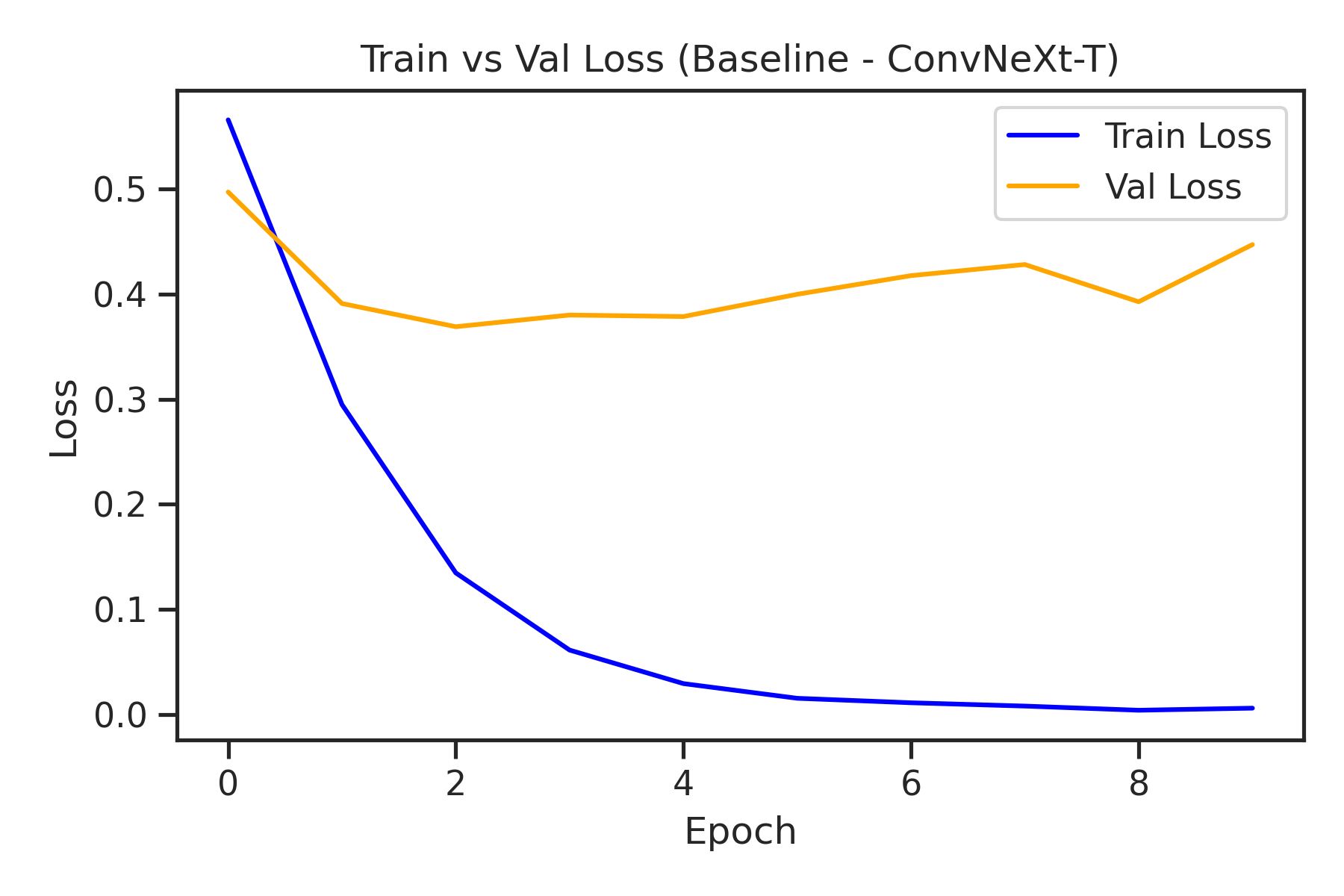}
        \caption{ConvNeXt-T – Baseline Loss}
    \end{subfigure}
    \hfill
    \begin{subfigure}[t]{0.24\textwidth}
        \centering
        \includegraphics[width=\textwidth]{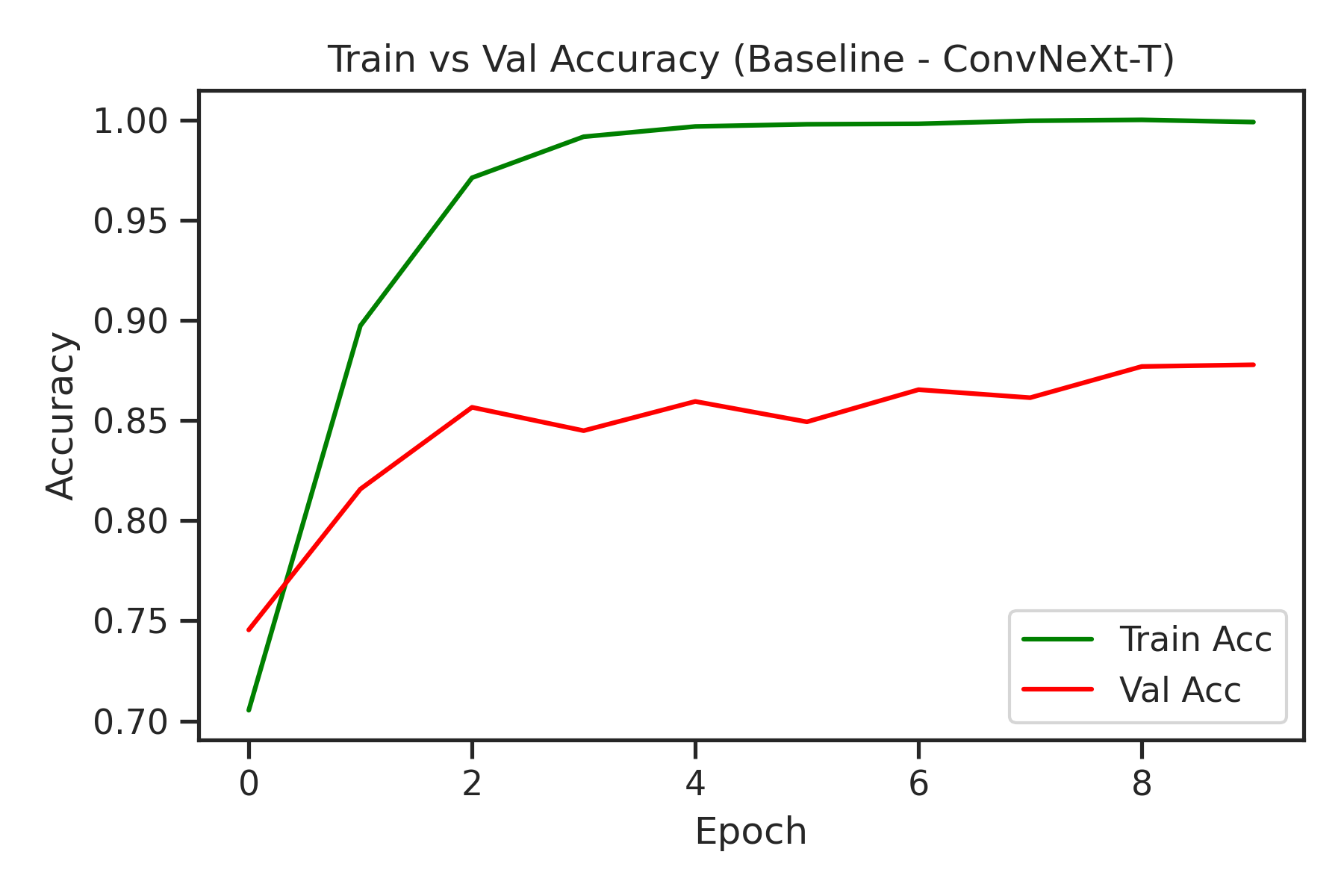}
        \caption{ConvNeXt-T – Baseline Accuracy}
    \end{subfigure}
    \hfill
    \begin{subfigure}[t]{0.24\textwidth}
        \centering
        \includegraphics[width=\textwidth]{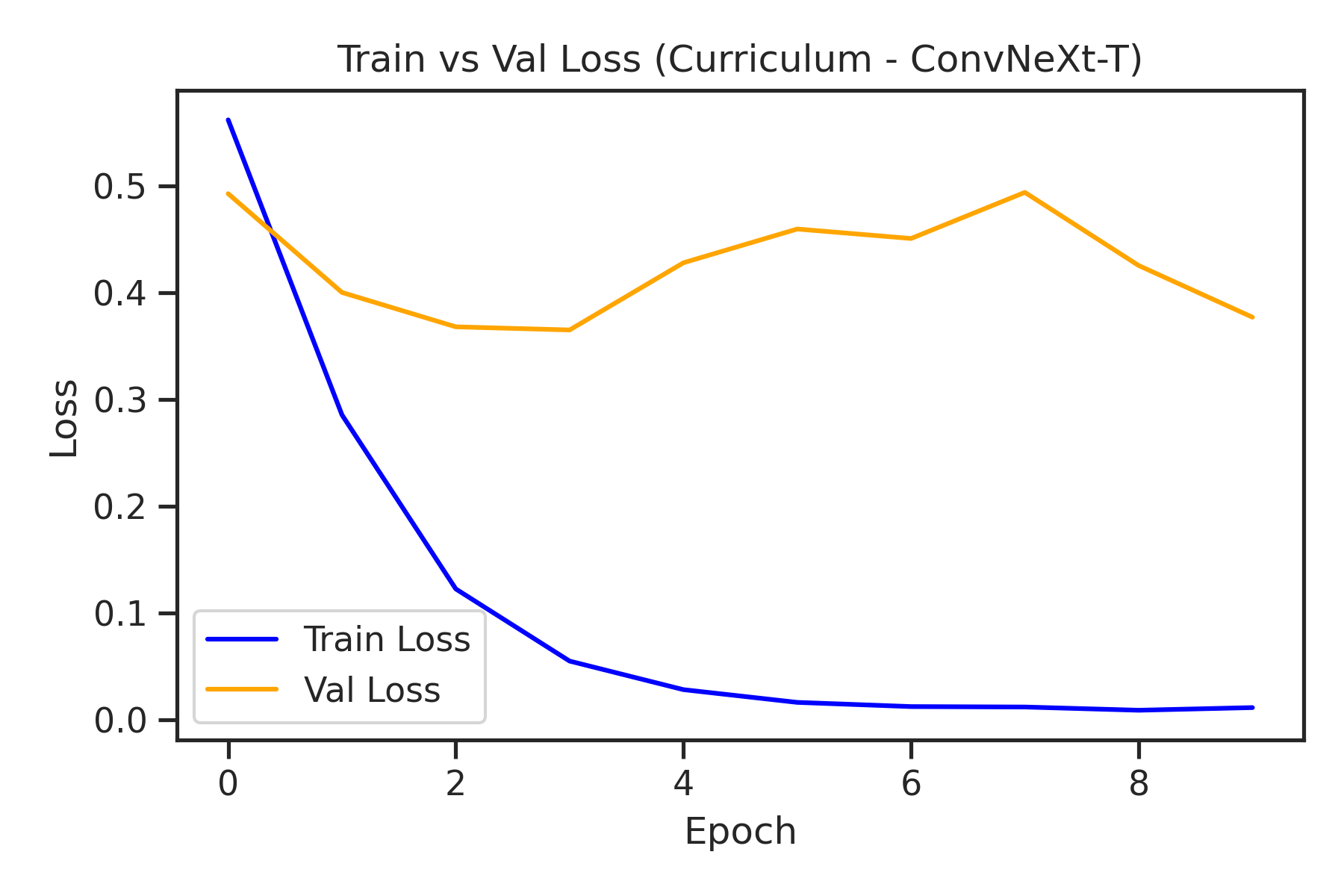}
        \caption{ConvNeXt-T – Curriculum Loss}
    \end{subfigure}
    \hfill
    \begin{subfigure}[t]{0.24\textwidth}
        \centering
        \includegraphics[width=\textwidth]{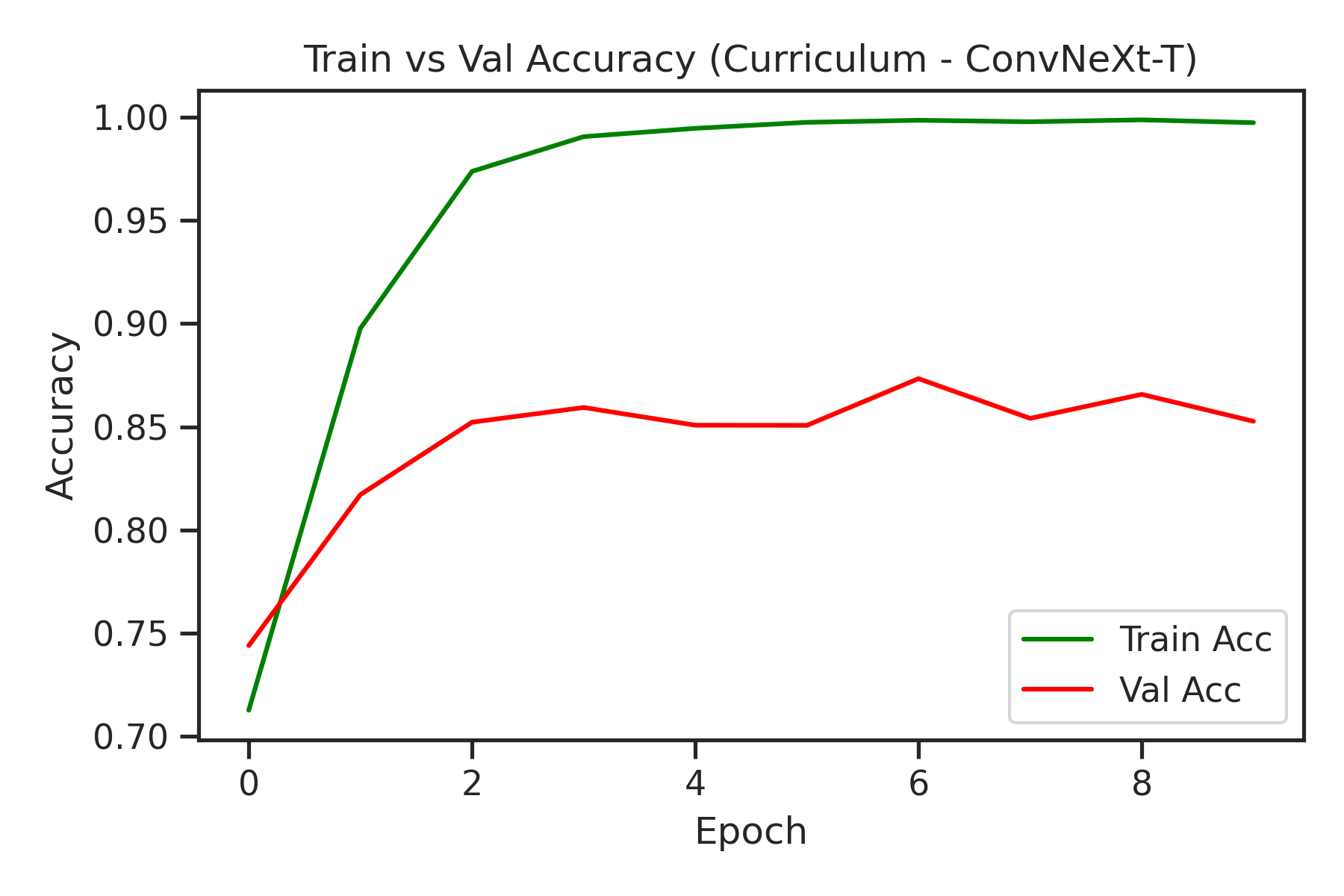}
        \caption{ConvNeXt-T – Curriculum Accuracy}
    \end{subfigure}

    \caption{
    Training curves for DenseNet121 and ConvNeXt-T under baseline and curriculum learning settings. 
    As visualized, curriculum learning consistently led to smoother convergence, reduced overfitting, 
    and improved validation stability for both CNN-based and transformer-based architectures.
    }
    \label{fig:training_curves}
\end{figure}

Representative training curves for DenseNet121 and ConvNeXt-T—the two best-performing models—are presented 
in Figure~\ref{fig:training_curves}, illustrating smoother convergence and reduced overfitting under curriculum learning. 
Training curves for the remaining ten models, covering lightweight CNNs, efficient CNNs, and vision transformers, 
are provided in Appendix~A.3 (Supplementary Figure S1) for completeness and comparative analysis.

\begin{sidewaystable}[p]
\centering
\footnotesize
\caption{
Performance metrics of all twelve candidate models across five architectural groups 
under both \textbf{Baseline (uniform training)} and \textbf{Curriculum} conditions.
Bolded rows denote the six models selected for downstream clinical validation and curriculum extension.
The \textbf{Overfitting} column reflects training dynamics: “Yes” = early saturation or large train–validation gaps; 
“Borderline” = fold-specific instability; “No” = minimal overfitting (shown in \textcolor{blue}{blue}). 
The best-performing model (\textbf{ConvNeXt-T (Curriculum)}) is highlighted in \textcolor{blue}{blue}.
}
\label{tab:sup_model_comparison}
\renewcommand{\arraystretch}{0.95}
\begin{tabular}{lcccccc}
\toprule
\textbf{Model} & \textbf{Group} & \textbf{AUC} & \textbf{Accuracy} &
\textbf{Sensitivity} & \textbf{Specificity} & \textbf{Overfitting} \\
\midrule
MobileNetV2 (Baseline)   & G1 & 0.8975 & 0.8203 & 0.7843 & 0.8495 & \textcolor{blue}{No} \\
\textbf{MobileNetV2 (Curriculum)} & G1 & \textbf{0.8985} & \textbf{0.8113} & \textbf{0.7648} & \textbf{0.8495} & \textbf{\textcolor{blue}{No}} \\
SqueezeNet1.1 (Baseline) & G1 & 0.7801 & 0.6663 & 0.7833 & 0.5708 & Yes \\
SqueezeNet1.1 (Curriculum) & G1 & 0.7667 & 0.6800 & 0.7848 & 0.5960 & Yes \\
\midrule
ResNet18 (Baseline)      & G2 & 0.8710 & 0.8020 & 0.8000 & 0.8050 & Yes \\
ResNet18 (Curriculum)    & G2 & 0.8740 & 0.7980 & 0.7810 & 0.8200 & Yes \\
DenseNet121 (Baseline)   & G2 & 0.8950 & 0.8110 & 0.8040 & 0.8160 & Yes \\
\textbf{DenseNet121 (Curriculum)} & G2 & \textbf{0.9150} & \textbf{0.8290} & \textbf{0.8030} & \textbf{0.8490} & \textbf{Borderline} \\
VGG16\_bn (Baseline)     & G2 & 0.8980 & 0.8336 & 0.7376 & 0.9129 & Yes \\
\textbf{VGG16\_bn (Curriculum)}   & G2 & \textbf{0.8840} & \textbf{0.8155} & \textbf{0.7352} & \textbf{0.8818} & \textbf{Borderline} \\
\midrule
EfficientNet-B3 (Baseline)   & G3 & 0.8901 & 0.8555 & 0.8048 & 0.8966 & Yes \\
EfficientNet-B3 (Curriculum) & G3 & 0.9180 & 0.8592 & 0.7929 & 0.9129 & Yes \\
EfficientNet-B4 (Baseline)   & G3 & 0.8761 & 0.7850 & 0.6862 & 0.8652 & Yes \\
EfficientNet-B4 (Curriculum) & G3 & 0.8590 & 0.7630 & 0.6667 & 0.8412 & Yes \\
\midrule
ViT-B/16 (Baseline)      & G4 & 0.8630 & 0.7900 & 0.7940 & 0.7860 & Yes \\
ViT-B/16 (Curriculum)    & G4 & 0.8790 & 0.7810 & 0.6970 & 0.8490 & Yes \\
Swin-T (Baseline)        & G4 & 0.8930 & 0.8420 & 0.7660 & 0.9060 & Yes \\
Swin-T (Curriculum)      & G4 & 0.8960 & 0.8190 & 0.7230 & 0.8970 & Yes \\
DeiT-S (Baseline)        & G4 & 0.9040 & 0.8334 & 0.7933 & 0.8652 & Yes \\
\textbf{DeiT-S (Curriculum)} & G4 & \textbf{0.9047} & \textbf{0.8502} & \textbf{0.7729} & \textbf{0.9129} & \textbf{Borderline} \\
\midrule
ConvNeXt-T (Baseline)    & G5 & 0.9128 & 0.8642 & 0.8324 & 0.8886 & Yes \\
\textbf{ConvNeXt-T (Curriculum)} & G5 & \textbf{\textcolor{blue}{0.9285}} & \textbf{\textcolor{blue}{0.8550}} & \textbf{\textcolor{blue}{0.8414}} & \textbf{\textcolor{blue}{0.8646}} & \textbf{\textcolor{blue}{No}} \\
MaxViT-T (Baseline)      & G5 & 0.9131 & 0.7811 & 0.7176 & 0.8338 & Yes \\
\textbf{MaxViT-T (Curriculum)} & G5 & \textbf{0.9094} & \textbf{0.8112} & \textbf{0.6857} & \textbf{0.9129} & \textbf{Borderline} \\
\bottomrule
\end{tabular}
\end{sidewaystable}

To evaluate whether curriculum learning genuinely contributed to performance improvement under low-data conditions, 
we compared baseline and curriculum training across all twelve model architectures (Table~\ref{tab:sup_model_comparison}). 
The results strongly support the effectiveness of the proposed curriculum design, 
reinforced by the statistical validity of the difficulty stratification. 
A Kruskal--Wallis~H test revealed a significant difference in difficulty score distributions across stages 
($H = 212.82$, $p < 0.0001$). 
Moreover, all pairwise Mann--Whitney~U tests yielded $p$-values below 0.0001, 
confirming that the difficulty levels were statistically distinct from one another. 
Notably, the contrast between the \textit{Easy} and \textit{Very Hard} groups was especially pronounced 
($U = 0.00$, $p < 0.0001$), strongly supporting the validity of the quantile-based stratification as an 
effective curriculum framework.

\subsection*{Effectiveness of Curriculum Learning}

To rigorously assess the effectiveness of curriculum learning under low-data constraints, we applied the FOSSIL weighting strategy to six selected models using softmax-based difficulty scores. As shown in Table~\ref{tab:sup_fossil_vs_frozen_allmodels}, FOSSIL-trained models consistently outperformed their FROZEN counterparts across key metrics such as AUC, accuracy, sensitivity, and F1 score.

\begin{sidewaystable}[p]
\centering
\footnotesize
\caption{
Performance comparison across six selected models using Curriculum, FOSSIL, and FROZEN settings.
Metrics include discrimination (AUC), accuracy, calibration (ECE), and clinical relevance (Sensitivity, Specificity, PPV, NPV).
Note: For DeiT-S, the ``FROZEN'' column represents performance under FOSSIL + SAM setting, as no backbone freezing was applied.
Highest values per metric are highlighted in \textcolor{blue}{blue}, and overfitting-prone cases are marked in \textcolor{red}{red}.
}
\label{tab:sup_fossil_vs_frozen_allmodels}
\renewcommand{\arraystretch}{0.95}
\begin{tabular}{lcccccc}
\toprule
\textbf{Metric} & \textbf{ConvNeXt-T} & \textbf{DenseNet121} & \textbf{DeiT-S} & \textbf{MaxViT-T} & \textbf{VGG16\_bn} & \textbf{MobileNetV2} \\
\midrule
\multicolumn{7}{l}{\textit{(Curriculum results from Table S4)}} \\
AUC           & 0.9285 & 0.9150 & 0.9047 & 0.9094 & 0.8840 & 0.8985 \\
Accuracy      & 0.8550 & 0.8290 & 0.8502 & 0.8112 & 0.8155 & 0.8113 \\
Sensitivity   & 0.8414 & 0.8030 & 0.7729 & 0.6857 & 0.7352 & 0.7648 \\
Specificity   & 0.8646 & 0.8490 & 0.9129 & 0.9129 & 0.8818 & 0.8495 \\
Overfitting   & \textcolor{blue}{No} & \textcolor{red}{Borderline} & \textcolor{red}{Borderline} & \textcolor{red}{Borderline} & \textcolor{red}{Borderline} & \textcolor{blue}{No} \\
\midrule
\multicolumn{7}{l}{\textit{(FOSSIL results)}} \\
AUC           & \textcolor{blue}{0.9573} & 0.9402 & 0.9430 & 0.8596 & 0.9358 & 0.9381 \\
Accuracy      & 0.8757 & \textcolor{blue}{0.8785} & 0.8491 & 0.6463 & 0.8449 & 0.8433 \\
Sensitivity   & 0.8395 & \textcolor{blue}{0.8622} & 0.8462 & 0.6311 & 0.8551 & 0.8184 \\
Specificity   & 0.9050 & \textcolor{blue}{0.8914} & 0.8515 & 0.6562 & 0.8354 & 0.8619 \\
F1 Score      & \textcolor{blue}{0.8560} & 0.8629 & 0.8341 & 0.5499 & 0.8332 & 0.8206 \\
PPV           & 0.8829 & \textcolor{blue}{0.8674} & 0.8442 & 0.5996 & 0.8248 & 0.8361 \\
NPV           & 0.8801 & \textcolor{blue}{0.8919} & 0.8815 & 0.6586 & 0.8827 & 0.8629 \\
ECE           & 0.5327 & 0.4735 & 0.5082 & 0.3833 & 0.5186 & \textcolor{blue}{0.5033} \\
Overfitting   & \textcolor{blue}{No} & \textcolor{blue}{No} & \textcolor{blue}{No} & \textcolor{red}{Borderline} & \textcolor{red}{Borderline} & \textcolor{blue}{No} \\
\midrule
\multicolumn{7}{l}{\textit{(FROZEN results)}} \\
AUC           & 0.8782 & 0.7956 & 0.4838 & 0.9522 & 0.9127 & 0.8372 \\
Accuracy      & 0.7965 & 0.7189 & 0.4915 & 0.8682 & 0.8287 & 0.7630 \\
Sensitivity   & 0.6624 & 0.6238 & 0.4797 & 0.8190 & 0.8006 & 0.6727 \\
Specificity   & 0.9045 & 0.7955 & 0.5001 & 0.9073 & 0.8513 & 0.8356 \\
F1 Score      & 0.7430 & 0.6597 & 0.4353 & 0.8453 & 0.8079 & 0.7163 \\
PPV           & 0.8518 & 0.7407 & 0.4300 & 0.8830 & 0.8238 & 0.7829 \\
NPV           & 0.7696 & 0.7274 & 0.5158 & 0.8671 & 0.8444 & 0.7622 \\
ECE           & 0.3846 & 0.2858 & 0.2391 & 0.5169 & 0.5036 & 0.3214 \\
Overfitting   & \textcolor{blue}{No} & \textcolor{red}{Borderline} & \textcolor{red}{Yes} & \textcolor{red}{Borderline} & \textcolor{red}{Borderline} & \textcolor{blue}{No} \\
\bottomrule
\end{tabular}
\end{sidewaystable}

Notably, ConvNeXt-T achieved the highest AUC (0.9573) and F1 score (0.8560) without signs of overfitting, establishing it as the most robust performer. DenseNet121 also showed substantial gains in sensitivity (0.8622) and specificity (0.8914), while maintaining model efficiency. Even the lightweight MobileNetV2 demonstrated significant improvements, with AUC increasing from 0.8372 to 0.9381 and F1 from 0.7163 to 0.8206, confirming the transferability of the proposed approach to resource-efficient architectures.

In contrast, DeiT-S experienced severe degradation under the SAM optimizer, suggesting that transformer-based models may be more sensitive to curriculum–optimizer interactions. MaxViT-T exhibited a precision–recall trade-off, indicating the need for calibration-aware strategies in hybrid transformer models.

To guide model prioritization for clinical deployment, we ranked the six models based on four criteria: (1) overall discriminative performance (AUC and F1), (2) clinical relevance (sensitivity and NPV), (3) resistance to overfitting, and (4) computational efficiency. 

Among all candidates, ConvNeXt-T emerged as the top performer, achieving superior scores across every evaluation metric without signs of overfitting, making it ideally suited for high-throughput clinical screening pipelines. The second-ranked model, DenseNet121, exhibited a balanced trade-off between sensitivity and specificity, coupled with high reproducibility, suggesting its suitability for scalable integration in hospital workflows. 

The third-ranked MobileNetV2 demonstrated strong overall accuracy while maintaining exceptional computational efficiency, marking it as an attractive option for mobile and edge-based diagnostic systems. In contrast, VGG16\_bn showed moderate gains but suffered from a slight drop in specificity, positioning it as more appropriate for educational or non-critical applications. 

MaxViT-T ranked fifth, displaying strong precision and specificity but reduced sensitivity, thereby requiring further optimization to achieve balanced lesion detection. Finally, DeiT-S performed poorest under the SAM optimizer, indicating sensitivity to optimizer–curriculum interactions and highlighting the need for additional stability-focused tuning before deployment.

These results confirm that difficulty-aware curriculum learning—particularly the FOSSIL approach—enhances classification performance across diverse model architectures while mitigating overfitting. Although FOSSIL-trained models frequently achieved near-perfect training accuracy, their validation metrics remained stable, underscoring the role of the proposed method as an effective regularization mechanism. The consistent occurrence of early stopping between epochs 8 and 27 across all folds further supports stable convergence and a low risk of overfitting.

A detailed statistical comparison of FOSSIL versus FROZEN configurations across models is provided in Supplementary Table S1. Collectively, these findings establish ConvNeXt-T as the most clinically robust and scalable model under the curriculum learning framework. Furthermore, the consistent performance gains observed across multiple architectures demonstrate the generalizability and practical applicability of the proposed strategy in low-data medical imaging contexts such as Mpox diagnosis.

To determine the most reliable ConvNeXt-T model trained with FOSSIL, we evaluated all 15 training instances generated across 3 random seeds and 5-fold cross-validation. Among these, the model trained with seed 123 and fold 0 demonstrated the most favorable performance across all key metrics. Specifically, this model achieved an AUC of 0.9998, an accuracy of 98.68\%, a sensitivity of 1.000, and a specificity of 97.62\%. 

Although these metrics indicate excellent discriminative performance, the unusually high sensitivity score naturally raises concerns about potential overfitting to this particular validation fold. To address this possibility, we plan to further assess the model's generalizability using an external validation dataset. Despite this caveat, its strong overall performance and stable calibration behavior make it the most appropriate choice for subsequent robustness testing, uncertainty calibration, and interpretability analyses.

\subsection*{Uncertainty Calibration and Interpretability}

\begin{figure}[H]
\centering
\begin{tabular}{ccc}
Input Image & Score-CAM Heatmap & Overlay \\
\includegraphics[width=0.2\textwidth]{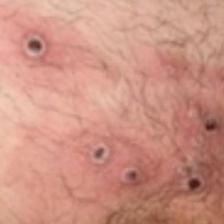} &
\includegraphics[width=0.2\textwidth]{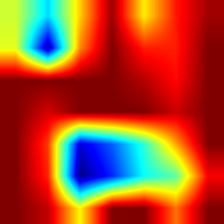} &
\includegraphics[width=0.2\textwidth]{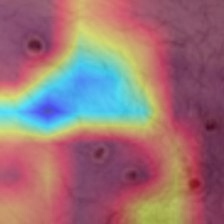} \\
\multicolumn{3}{c}{(a) Easy Case} \\
\\
\includegraphics[width=0.2\textwidth]{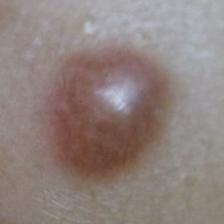} &
\includegraphics[width=0.2\textwidth]{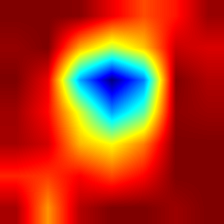} &
\includegraphics[width=0.2\textwidth]{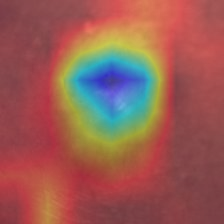} \\
\multicolumn{3}{c}{(b) Very Hard Case} \\
\end{tabular}
\caption{
Score-CAM visualizations of representative Easy and Very Hard samples obtained from the FOSSIL-trained ConvNeXt-T model (seed 123, fold 0). 
Each row displays the input image, activation heatmap, and overlay. 
Attention is well localized around the lesion in Easy cases but becomes diffuse under visually ambiguous conditions, indicating higher model uncertainty.
}
\label{fig:scorecam_convnext_seed123_fold0}
\end{figure}

To examine interpretability and uncertainty behavior, we analyzed the FOSSIL-trained ConvNeXt-T configuration corresponding to seed~123, fold~0, which exhibited median-level validation performance across all 15 training instances (3 seeds~$\times$~5 folds) and thus represents a typical model behavior. 
Score-CAM, a gradient-free attribution technique, was used to generate spatially precise activation maps. 
Figure~\ref{fig:scorecam_convnext_seed123_fold0} illustrates representative examples from opposite ends of the curriculum—Easy and Very Hard. 
In the Easy case, the activation map is sharply focused on the lesion, reflecting confident and well-calibrated predictions. 
Conversely, the Very Hard case shows broader, less concentrated activations, consistent with higher epistemic uncertainty.

Although the Very Hard sample is clinically consistent with Mpox, its elevated difficulty score arises from morphological overlap with non-Mpox lesions within the training set. 
Hence, difficulty here captures not visual obscurity but the model’s intrinsic uncertainty in differentiating visually analogous dermatological patterns. 
This observation reinforces the central premise of the proposed approach: samples with lower confidence are down-weighted early in training, allowing the model to master unambiguous examples before progressively engaging with more complex, high-uncertainty cases.

The diffuse yet structured activation observed in the Very Hard example therefore reflects representational overlap between classes rather than random attention dispersion. By minimizing cumulative regret through difficulty-aware weighting, the proposed method enables the network to gradually disambiguate ambiguous samples and align its spatial focus with clinically meaningful lesion regions. 
Such calibration-driven interpretability demonstrates that the proposed approach not only improves discrimination performance but also enhances model reliability and clinical trustworthiness in visually confounding scenarios.

\subsection*{Robustness under Real-World Perturbations}

To emulate deployment scenarios such as mobile-based diagnosis or teledermatology, we evaluated the FOSSIL-trained ConvNeXt-T model under five common real-world image perturbations: JPEG compression (quality 30 and 50), Gaussian blur, and brightness or contrast shifts.

\begin{figure}[H]
    \centering
    \includegraphics[width=1.0\linewidth]{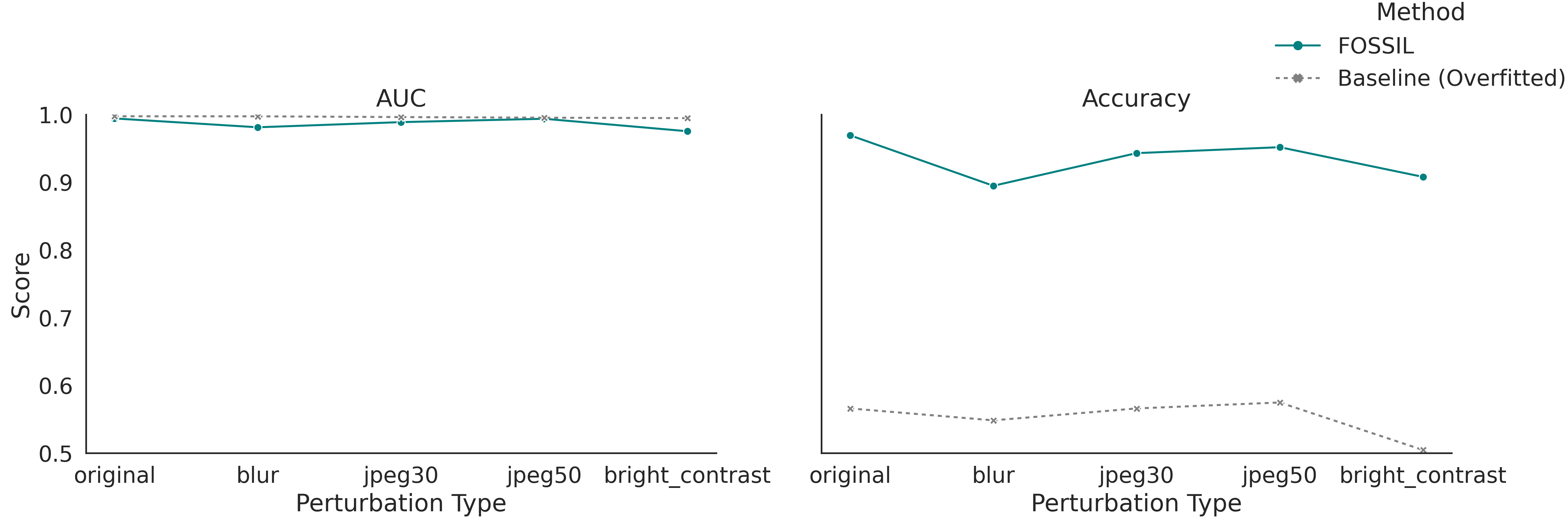}
    \caption{
    Diagnostic performance of the FOSSIL-trained ConvNeXt-T model compared with its baseline counterpart under five real-world perturbation settings. 
    The FOSSIL model maintained high AUC and accuracy across all distortions, whereas the baseline model showed inflated AUC but degraded accuracy—consistent with overfitting and reduced reliability under realistic input shifts.
    }
    \label{fig:robustness_barplot}
\end{figure}

As shown in Figure~\ref{fig:robustness_barplot}, the model consistently preserved high diagnostic performance across all conditions, achieving AUC values above 0.975 and accuracies exceeding 89\%. 
Performance was particularly stable under JPEG compression (AUC~$\approx$~0.994, Accuracy~$\approx$~95.2\% at quality~50), underscoring its resilience to lossy image formats commonly used in telemedicine workflows.

Gaussian blur produced the largest degradation (Accuracy~$\approx$~89.5\%), suggesting partial reliance on fine-grained texture information.

Nevertheless, performance remained well within clinically acceptable limits, demonstrating strong robustness under suboptimal imaging conditions and practical feasibility for low-resource or remote diagnostic settings.

For comparison, the baseline ConvNeXt-T model (gray dashed line in Figure~\ref{fig:robustness_barplot}) exhibited lower accuracy across all perturbations despite maintaining high AUC values—a hallmark of overfitting and poor calibration. 
This divergence between discrimination and reliability underscores the importance of jointly evaluating both when assessing clinical AI systems.

Given its superior generalization and stability across diverse perturbations, the FOSSIL-trained ConvNeXt-T configuration was designated as the reference model for all subsequent analyses, including uncertainty calibration and external validation.

\subsection*{External Validation on the MCSI Dataset}

To evaluate the model’s generalization beyond the internal validation environment, external testing was performed using the independent \textit{Mpox Close Skin Images (MCSI)} dataset. 
This dataset contains 200 clinical photographs (100 Mpox-positive, 100 Mpox-negative) collected under heterogeneous real-world conditions, including five diagnostic categories—normal skin, chickenpox, and acne (levels 1–3). 
The images capture wide variability in lesion morphology, background texture, and skin tones, providing a rigorous test of out-of-distribution robustness.

Table~\ref{tab:sup_external_mcsi} summarizes the quantitative results of this external validation experiment. 
Without any model adaptation or fine-tuning, the FOSSIL-trained ConvNeXt-T achieved an AUC of 0.963, an accuracy of 93.5\%, a sensitivity of 92.0\%, and a specificity of 95.0\% on the MCSI dataset. 
These results demonstrate strong out-of-sample generalization, confirming that difficulty-aware curriculum learning enables the extraction of transferable and clinically relevant visual features.

\begin{table}[H]
\centering
\footnotesize
\caption{External validation results on the MCSI dataset using the FOSSIL-trained ConvNeXt-T model. No fine-tuning or domain adaptation was applied.}
\label{tab:sup_external_mcsi}
\begin{tabular}{lcc}
\toprule
Metric & Value \\
\midrule
AUC              & 0.963 \\
Accuracy         & 93.5\% \\
Sensitivity      & 92.0\% \\
Specificity      & 95.0\% \\
Number of Images & 200 (100 positive / 100 negative) \\
\bottomrule
\end{tabular}
\end{table}

\begin{figure}[H]
    \centering
    \includegraphics[width=1.0\textwidth]{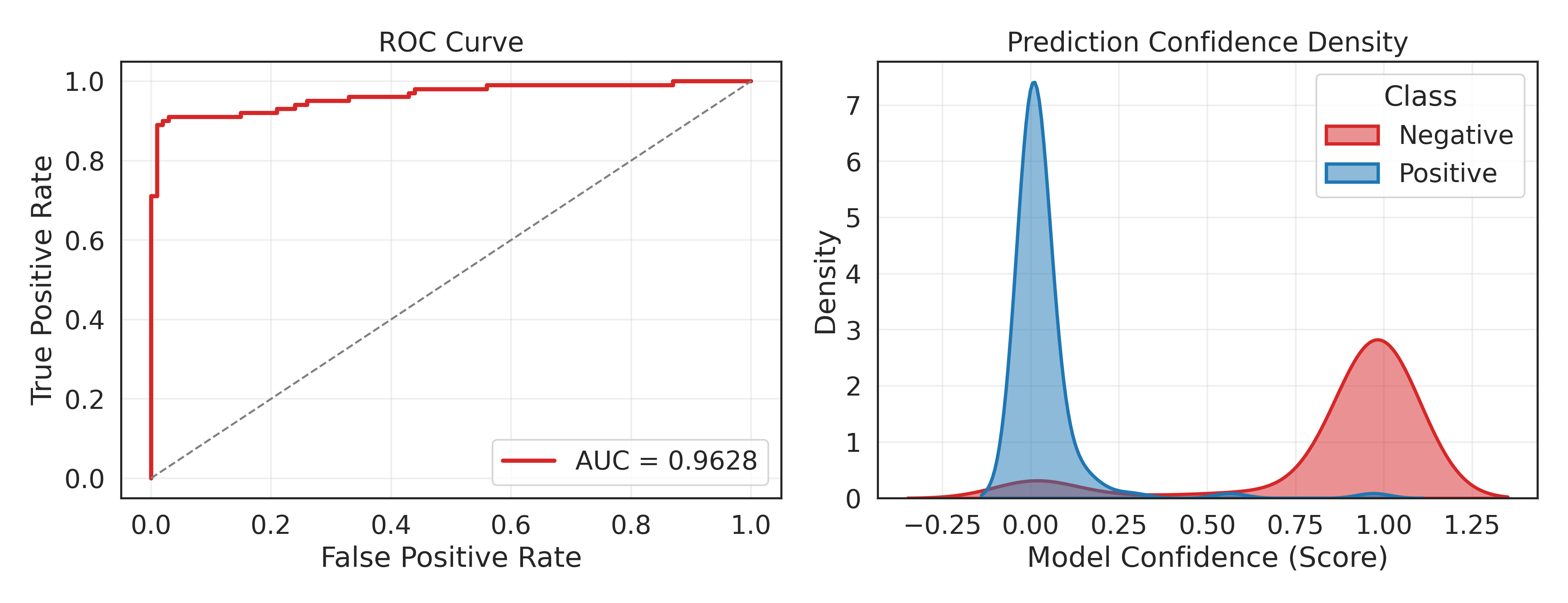}
    \caption{
    External validation on the MCSI dataset using the FOSSIL-trained ConvNeXt-T model.
    Left: ROC curve demonstrating high discriminative performance on previously unseen Mpox and non-Mpox images. 
    Right: Confidence score distributions showing clear separation between positive and negative classes, reflecting robust calibration under domain shift.
    }
    \label{fig:mcsi_external_val}
\end{figure}

Figure~\ref{fig:mcsi_external_val} provides a visual summary of external validation performance. 
The ROC curve on the left demonstrates high discriminative ability across unseen lesion types, while the confidence distribution plot on the right reveals well-separated class probabilities, indicating stable calibration even under substantial domain shift.

\section{Discussion}
\label{sec:Discussion}

\subsection{Clinical implications}
The proposed regret-minimizing weighting framework~\citep{cha2025fossil} offers a theoretically grounded yet 
clinically deployable solution to persistent challenges such as data scarcity, 
class imbalance, and augmentation bias in medical imaging. 
Originally formulated as a regret-minimization strategy for small and imbalanced datasets, 
the method integrates difficulty-aware weighting, class-prior correction, and augmentation penalties 
into a single interpretable objective. 
In the context of Mpox lesion classification—where training data are limited and inter-class overlap is substantial— 
our approach improved both generalization and calibration compared 
with empirical risk minimization (ERM) and focal-loss baselines~\citep{lin2017focal,guo2017calibration,nixon2019measuring}. 
The weighting structure stabilized optimization and mitigated overfitting, 
consistent with theoretical guarantees from online convex optimization 
and cumulative regret minimization~\citep{hazan2016introduction,shalev2012online,cha2025fossil}.

Clinically, the observed gains in calibration and interpretability are especially relevant for 
teledermatology, where confirmatory metadata or PCR validation are often unavailable. 
In our Grad-CAM analyses, models trained under the proposed weighting framework maintained 
morphological consistency with canonical Mpox lesion presentations~\citep{adler2022clinical,huhn2005clinical}, 
indicating that the learned representations align with dermatological reasoning rather than 
spurious background artifacts. 
This interpretability aligns with recent initiatives toward 
trustworthy AI for dermatology and infectious disease triage~\citep{esteva2017dermatologist,brinker2019deep,rajpurkar2022ai,kaur2022artificial}. 
By reducing dependence on aggressive augmentations and metadata, 
the method addresses an unmet need in AI-driven public health applications 
where privacy and regulatory constraints inherently limit data diversity.

\subsection{Broader applicability and generalization}
Although the empirical evaluation focuses on Mpox, the underlying issues of 
limited, imbalanced, and augmentation-sensitive data pervade biomedical research. 
The unified mathematical structure of our framework allows direct extension to 
radiology, histopathology, microscopy, omics, and longitudinal monitoring 
tasks~\citep{bengio2009curriculum,gupta2020curriculum,liu2021curriculum}. 
Table~\ref{tab:domains} summarizes representative examples across these modalities, 
highlighting common causes of data scarcity, typical augmentation risks, 
and how the proposed weighting mechanism mitigates them through difficulty, class-prior, and augmentation components. 
This cross-domain synthesis demonstrates that the framework is not limited to dermatology 
but provides a transferable learning paradigm for biologically constrained datasets.

\begin{sidewaystable}[p]
\centering
\caption{Representative biomedical domains (primary) and selected disaster-response applications (secondary). 
Each row lists the source of data scarcity, example data modalities, key augmentation risks, 
and how the proposed framework mitigates them via difficulty weighting, class-prior correction, and augmentation penalties.}
\label{tab:domains}
\scriptsize
\renewcommand{\arraystretch}{1.15}
\begin{tabular}{p{3.3cm} p{4cm} p{4cm} p{4.5cm} p{4.5cm}}
\toprule
Domain & Data Scarcity Cause & Example Data Types & Augmentation Risk & Mitigation Mechanism \\
\midrule
\multicolumn{5}{l}{\emph{Health / Biological (Primary Focus)}} \\
\midrule
Rare Disease Imaging &
Limited cases, IRB/privacy constraints &
Dermatology photos, OCT, MRI, pathology slides &
Color/structure distortions that confound diagnosis &
Balances minority classes and penalizes implausible augmentations; preserves clinically relevant morphology. \\
Histopathology &
Expert labeling cost; staining variability &
H\&E and IHC slides &
Aggressive color jitter or normalization altering tissue micro-architecture &
Reweights plausible stain variations and suppresses unrealistic transformations. \\
Radiology (MRI/CT) &
Expensive acquisition; ethics; limited longitudinal scans &
Brain MRI, chest CT, cardiac MRI &
Geometric or intensity transforms breaking anatomical plausibility &
Downweights anatomy-distorting augmentations; emphasizes realistic variability. \\
Microscopy &
Limited labeled cells; high magnification cost &
Cell morphology, fluorescence microscopy &
Shape/color perturbations creating non-physical organelles &
Penalizes non-physical transformations; preserves valid morphology signal. \\
Genomics \& Proteomics &
Costly sequencing; rare variants &
DNA/RNA sequences, protein structures &
Synthetic mutations or k-mer shuffles lacking biological plausibility &
Suppresses unrealistic sequences; highlights rare but meaningful variants. \\
Single-cell Omics &
Expensive per-cell profiling &
scRNA-seq, CyTOF, ATAC-seq &
Oversampling inducing artificial cell clusters &
Controls oversampled clusters via difficulty-based weighting; preserves true population structure. \\
Longitudinal Clinical Studies &
Slow follow-up; missingness; cohort attrition &
Wearable sensors, EMR time series &
Time-warping that creates unrealistic trajectories &
Penalizes implausible temporal augmentations; aligns weights with realistic progression. \\
Infectious Disease Outbreaks &
Biosafety limits; sporadic events &
Pathogen genomes, case charts &
Simulated outbreaks with non-epidemiological dynamics &
Downweights unrealistic synthetic outbreaks; enhances generalization to real data. \\
\midrule
\multicolumn{5}{l}{\emph{Disaster / Extreme-Event Applications (Secondary)}} \\
\midrule
Disaster Monitoring (Wildfire, Earthquake) &
Rare catastrophic events; few labeled images &
Thermal satellite imagery, seismic logs &
Simulations with unrealistic fire spread or seismic patterns &
Penalizes non-physical dynamics; prioritizes realistic hazard signals. \\
Remote Sensing for Damage Assessment &
Annotation bottlenecks; event-specific scarcity &
Post-disaster aerial/satellite images &
Texture/geometry perturbations producing false damage cues &
Suppresses artifact-heavy augmentations; emphasizes credible structural change. \\
Extreme-Environment Robotics (Polar/Deep Sea) &
Costly deployments; harsh conditions &
Sonar, LiDAR, ROV feeds &
Environment transforms not matching sensor physics &
Aligns weights with sensor-consistent distortions; reduces overfitting to implausible scenes. \\
\bottomrule
\end{tabular}
\end{sidewaystable}

\noindent
\textit{Interpretation.}
Table~\ref{tab:domains} highlights that data scarcity and augmentation bias 
are pervasive across biomedical and extreme-environment datasets. 
The proposed weighting strategy~\citep{cha2025fossil} provides a unified, interpretable, and theoretically guaranteed 
optimization scheme that generalizes across modalities without architectural modification. 
Difficulty-based terms prevent domination by trivial samples, 
class-prior correction stabilizes rare-class learning, 
and augmentation penalties reduce sensitivity to non-physiological transformations. 
This generalization potential bridges theoretical machine learning 
and practical biomedical deployment, positioning the framework as a 
foundational component for robust small-data healthcare AI.

\subsection{Limitations and future directions}
Despite promising results, the current evaluation remains retrospective. 
Future work will include multi-institutional external validation, 
adaptive temperature-scheduling strategies, 
and integration with self-paced or meta-learning frameworks~\citep{kumar2010self,wang2021survey}. 
Prospective clinical trials should assess the safety, interpretability, and 
decision impact of models trained under the proposed weighting framework 
in real-world workflows. 
Such extensions could establish this regret-minimizing approach as a 
general foundation for robust, ethically compliant medical AI across data-limited domains.

\section{Conclusion}

This work presents a unified, regret-minimizing learning framework that systematically addresses the intertwined challenges of data scarcity, class imbalance, and augmentation bias in medical imaging. 
By integrating difficulty-aware weighting, class-prior correction, and augmentation control into a single interpretable objective, the proposed approach provides both theoretical rigor and empirical reliability. 
Applied to Mpox skin lesion diagnosis, the framework consistently improved discrimination, calibration, and robustness without reliance on metadata or synthetic augmentation, establishing a principled route to stable optimization under extreme data constraints.

Beyond its technical contributions, this study demonstrates that curriculum-inspired, regret-based weighting can bridge the gap between optimization theory and clinical deployment. 
Models trained under this paradigm exhibited transparent, anatomically grounded attention patterns and well-calibrated uncertainty estimates, aligning machine perception with clinical reasoning. 
Such properties are central to the safe, trustworthy, and reproducible use of AI in telemedicine and other decentralized healthcare environments.

The broader implication of this work extends far beyond dermatology. 
Because the proposed framework operates independently of model architecture or data modality, it offers a general foundation for learning from limited, heterogeneous, and ethically constrained datasets across radiology, histopathology, genomics, and longitudinal monitoring. 
Its theoretical connection to regret minimization provides a unifying lens through which stability, fairness, and interpretability can be jointly optimized.

Future research should explore adaptive temperature scheduling, integration with self-paced or meta-learning paradigms, and multi-institutional prospective validation to establish clinical-grade reliability. 
By uniting mathematical structure with biomedical relevance, this regret-minimizing framework lays the groundwork for a new generation of reliable, data-efficient, and ethically aligned medical AI systems.

\section*{Data and Code Availability}
All training scripts and pretrained models will be publicly released on GitHub upon publication.

\section*{Declaration of Competing Interest}
The authors declare no competing interests.

\section*{Acknowledgments}
The authors thank collaborators and reviewers for valuable feedback.

\bibliographystyle{elsarticle-num}
\bibliography{FOSSIL_MPOX}

\appendix
\section*{Appendix}

\renewcommand{\thefigure}{A.\arabic{figure}}
\setcounter{figure}{0}
\renewcommand{\thetable}{A.\arabic{table}}
\setcounter{table}{0}




\subsection*{A.1. Motivation and Methodology}

Prior to adopting the softmax-based difficulty metric, 
a preliminary experiment was conducted using an entropy-based formulation to quantify diagnostic ambiguity in Mpox-positive samples. 
For each training image $x_i$, difficulty was defined as the Shannon entropy of the model’s predictive distribution:
\[
H(x_i) = -\sum_c p_i^{(c)} \log p_i^{(c)},
\]
where $p_i^{(c)}$ denotes the predicted softmax probability for class $c$.  
Higher entropy indicates greater uncertainty and hence higher inferred difficulty.  
A pretrained DenseNet121 model was used to compute entropy scores for the Mpox-positive subset ($n = 102$).

\subsection*{A.2. Results and Observations}

\begin{figure}[H]
    \centering
    \includegraphics[width=0.85\linewidth]{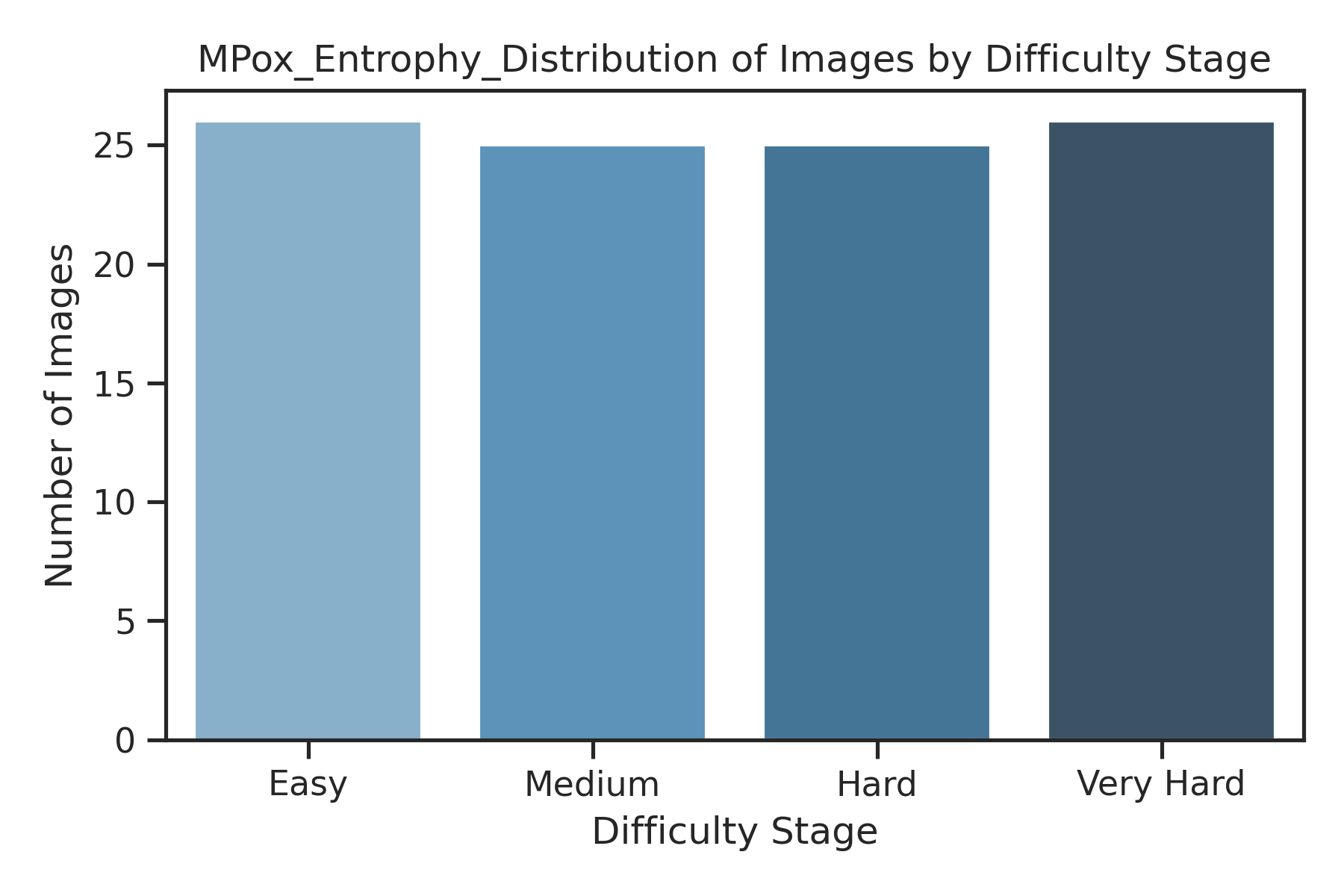}
    \caption{
    \textbf{Entropy-based difficulty analysis for Mpox-positive samples ($n=102$).}
    (a) Distribution of entropy scores and (b) resulting stage allocation demonstrate 
    the feasibility of difficulty-aware curriculum design. 
    However, fold-dependent skewness led to unstable stage thresholds.
    }
    \label{fig:entropy_difficulty_analysis_supplement}
\end{figure}

\begin{table}[H]
\centering
\footnotesize
\caption{
\textbf{Summary statistics of entropy-based difficulty scores across folds.} 
Substantial variation in the upper quartile across folds motivated the transition to the softmax-based difficulty formulation.
}
\label{tab:entropy_difficulty_stats_supplement}
\begin{tabular}{lccccccc}
\toprule
\textbf{Fold} & \textbf{Count} & \textbf{Mean} & \textbf{Std} & \textbf{Min} & \textbf{25\%} & \textbf{Median} & \textbf{Max} \\
\midrule
0 & 102 & 0.847 & 0.198 & 0.322 & 0.731 & 0.871 & 1.220 \\
1 & 102 & 0.894 & 0.192 & 0.339 & 0.780 & 0.910 & 1.246 \\
2 & 102 & 0.812 & 0.208 & 0.310 & 0.689 & 0.821 & 1.198 \\
3 & 102 & 0.860 & 0.183 & 0.346 & 0.745 & 0.878 & 1.201 \\
4 & 102 & 0.874 & 0.177 & 0.331 & 0.763 & 0.887 & 1.193 \\
\bottomrule
\end{tabular}
\end{table}

Entropy successfully captured variations in diagnostic ambiguity but exhibited asymmetric and fold-specific distributions 
(Figure~\ref{fig:entropy_difficulty_analysis_supplement}). 
Table~\ref{tab:entropy_difficulty_stats_supplement} shows that mean and quartile values varied substantially across folds, 
making it difficult to define stable and reproducible difficulty thresholds. 
These inconsistencies motivated the adoption of a simpler and more robust softmax-based formulation 
($d_i = 1 - \max_c p_i^{(c)}$) for the final experiments.

\subsection*{A.3. Summary of Findings}

The entropy-based approach provided conceptual validation for the curriculum-learning hypothesis—demonstrating that model uncertainty 
can serve as a meaningful proxy for sample difficulty. 
However, its statistical instability under fold resampling limited its practical applicability. 
In contrast, the softmax-based metric yielded consistent quantile boundaries and reproducible stage definitions, 
making it more suitable for small-data biomedical applications.


\subsection*{A.2. Fold-Level Class Distributions}

\begin{table}[H]
\centering
\footnotesize
\caption{
\textbf{Stratified class distribution across five folds under softmax-based curriculum learning.} 
Values confirm balanced class composition and consistent sample allocation across folds.
}
\label{tab:sup_cv_folds}
\begin{tabular}{cccccccc}
\toprule
\textbf{Fold} & \textbf{Train} & \textbf{Val} & \textbf{Train (Neg)} & \textbf{Train (Pos)} & \textbf{Val (Neg)} & \textbf{Val (Pos)} & \textbf{Total} \\
\midrule
0 & 182 & 46 & 101 & 81 & 25 & 21 & 228 \\
1 & 182 & 46 & 101 & 81 & 25 & 21 & 228 \\
2 & 182 & 46 & 100 & 82 & 26 & 20 & 228 \\
3 & 183 & 45 & 101 & 82 & 25 & 20 & 228 \\
4 & 183 & 45 & 101 & 82 & 25 & 20 & 228 \\
\bottomrule
\end{tabular}
\end{table}

\subsection*{B.2. Implementation Summary}

Each fold was trained independently using identical hyperparameters and random seeds (42, 77, 123). 
Difficulty scores were recomputed per fold to ensure independence between training and validation partitions, 
thereby preventing any leakage of stage information. 
Performance metrics reported in the main text represent the mean across all folds, 
providing a statistically reliable estimate of model generalization.

\subsection*{A.3. Training and validation curves for ten models}
\label{appendix:curves}

\begin{figure}[H]
    \begin{subfigure}[t]{0.24\textwidth}
    \centering
    \includegraphics[width=\textwidth]{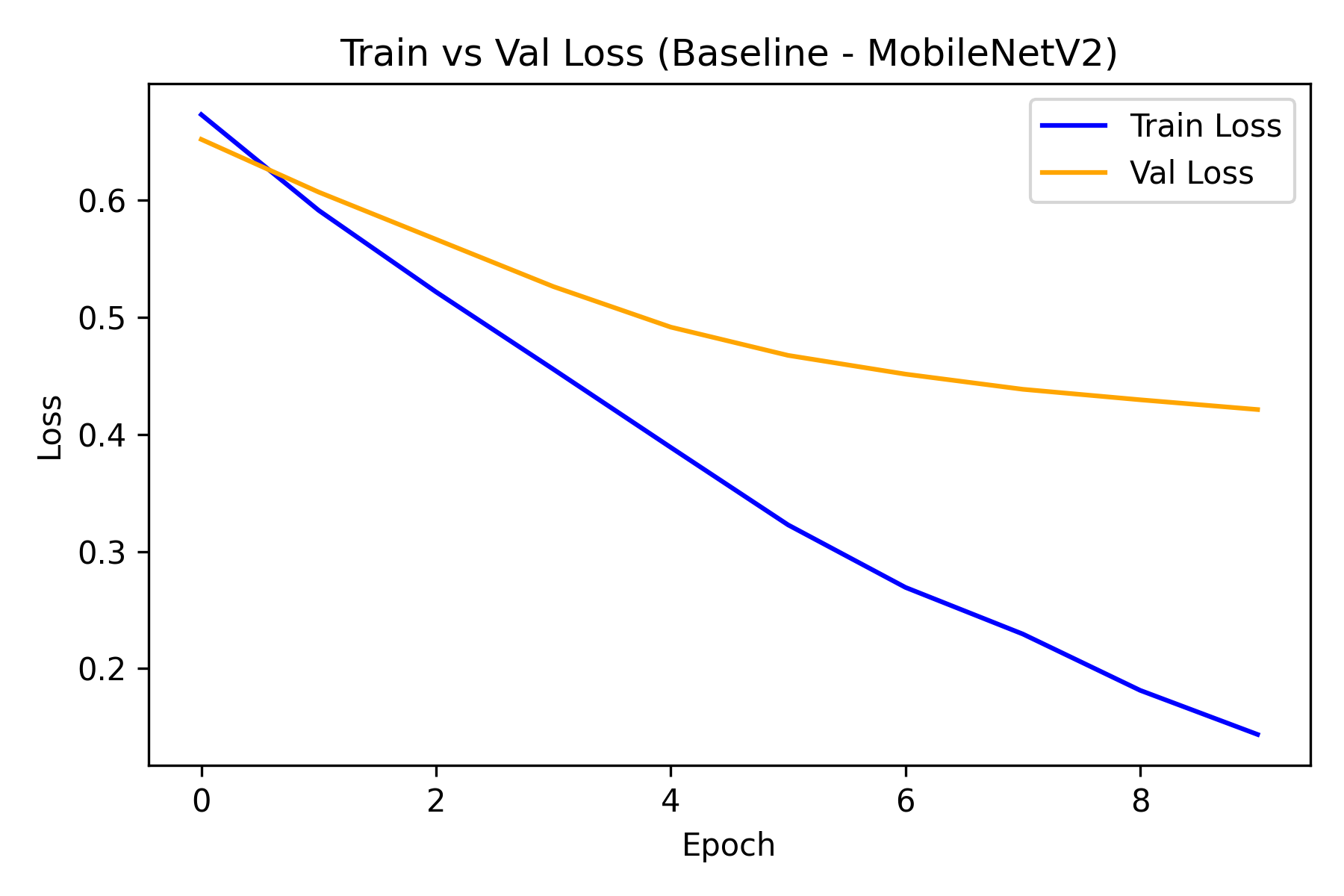}
    \caption{MobileNetV2 Baseline - Loss}
    \end{subfigure}
    \hfill
    \begin{subfigure}[t]{0.24\textwidth}
    \centering
    \includegraphics[width=\textwidth]{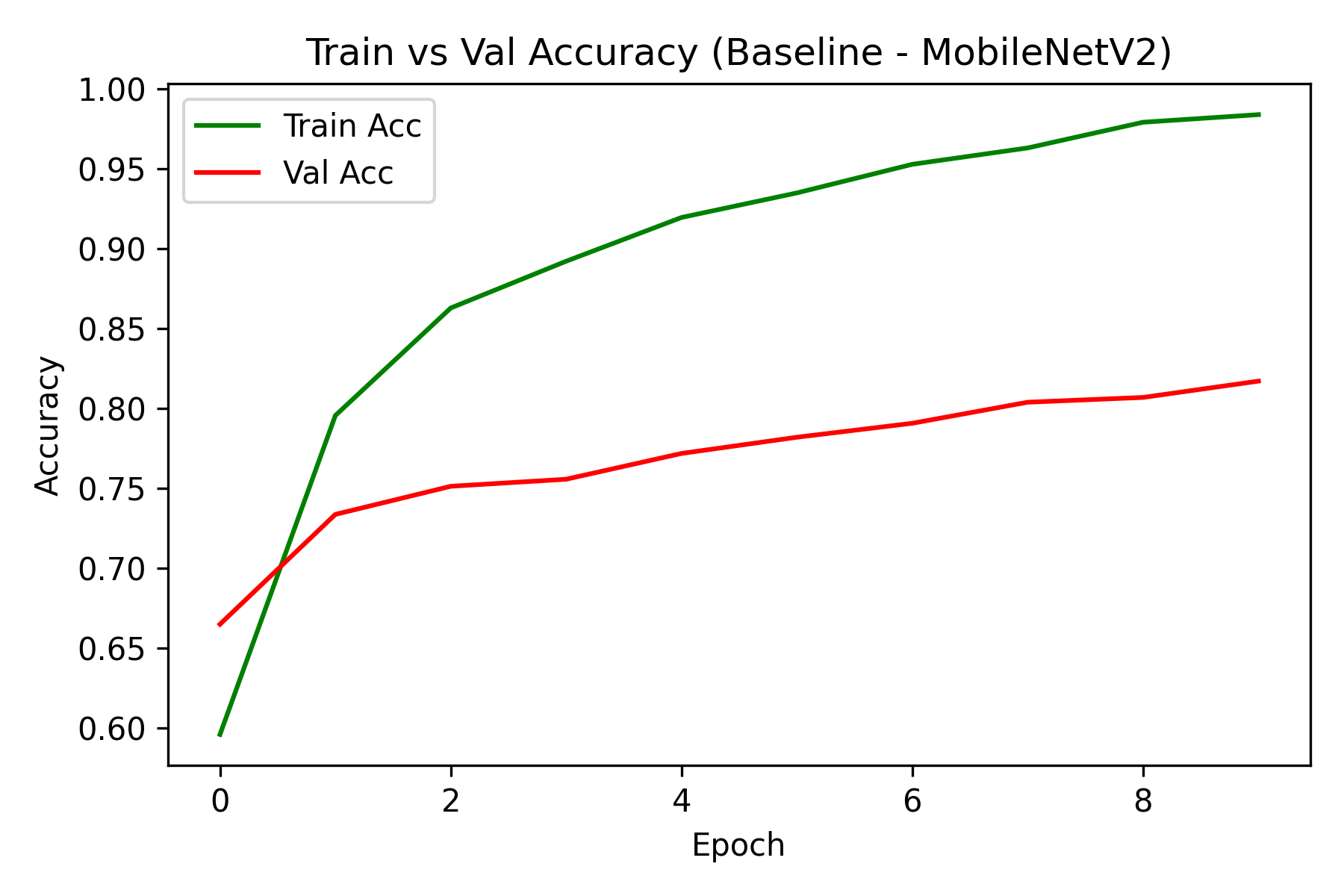}
    \caption{MobileNetV2 Baseline - Accuracy}
    \end{subfigure}
    \hfill
    \begin{subfigure}[t]{0.24\textwidth}
    \centering
    \includegraphics[width=\textwidth]{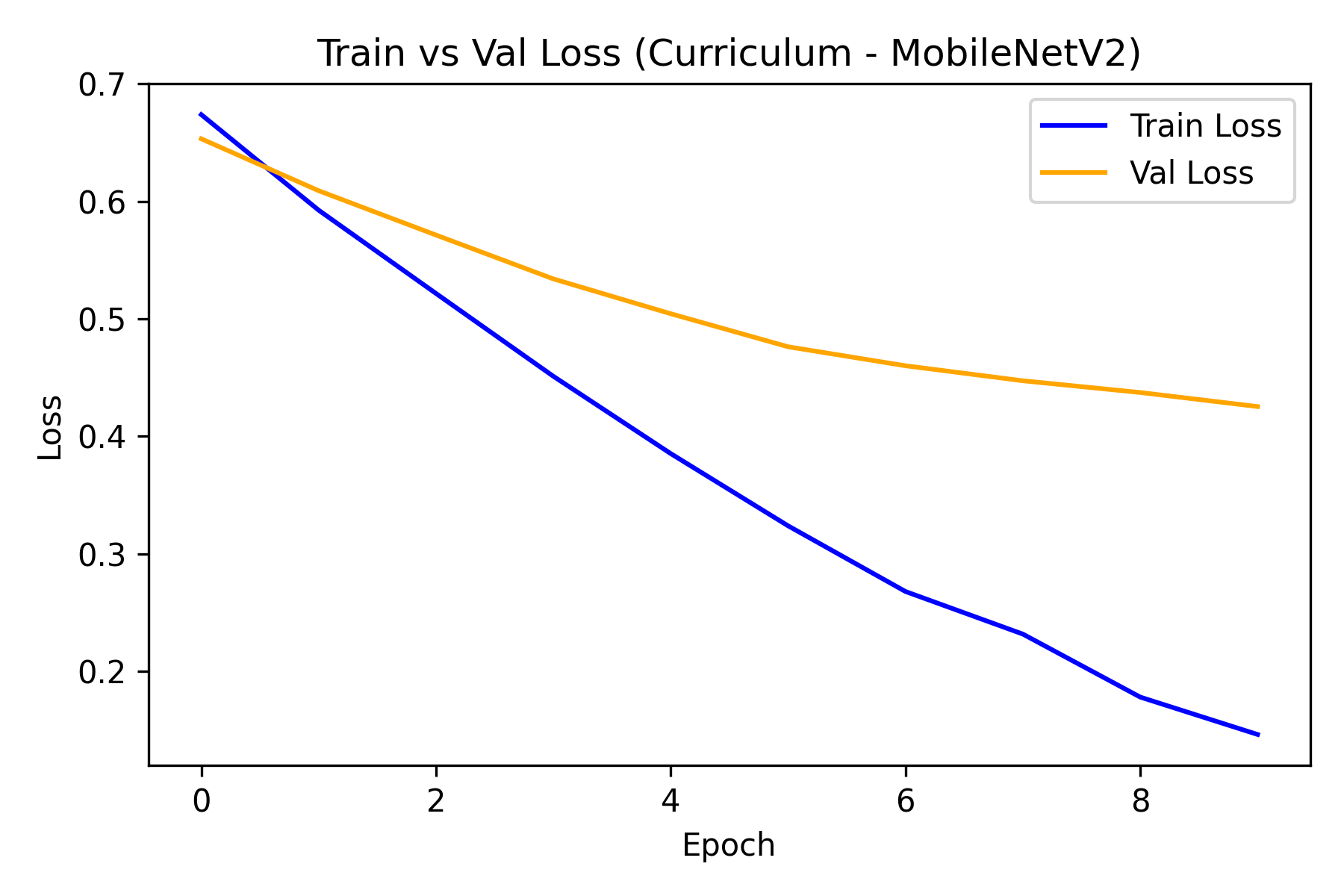}
    \caption{(MobileNetV2 Curriculum - Loss}
    \end{subfigure}
    \hfill
    \begin{subfigure}[t]{0.24\textwidth}
    \centering
    \includegraphics[width=\textwidth]{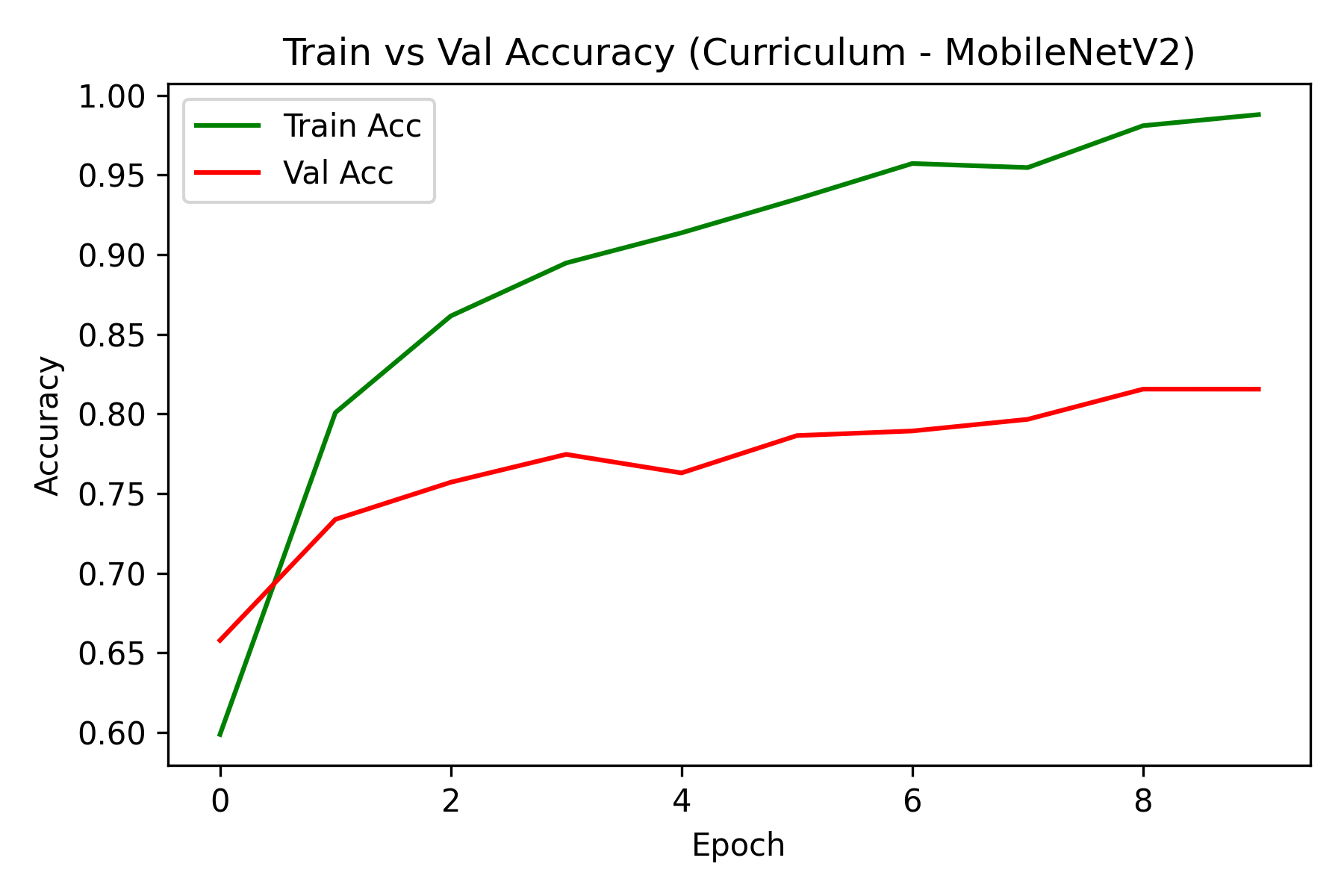}
    \caption{(MobileNetV2 Curriculum - Accuracy}
    \end{subfigure}

    \begin{subfigure}[t]{0.24\textwidth}
        \centering
        \includegraphics[width=\textwidth]{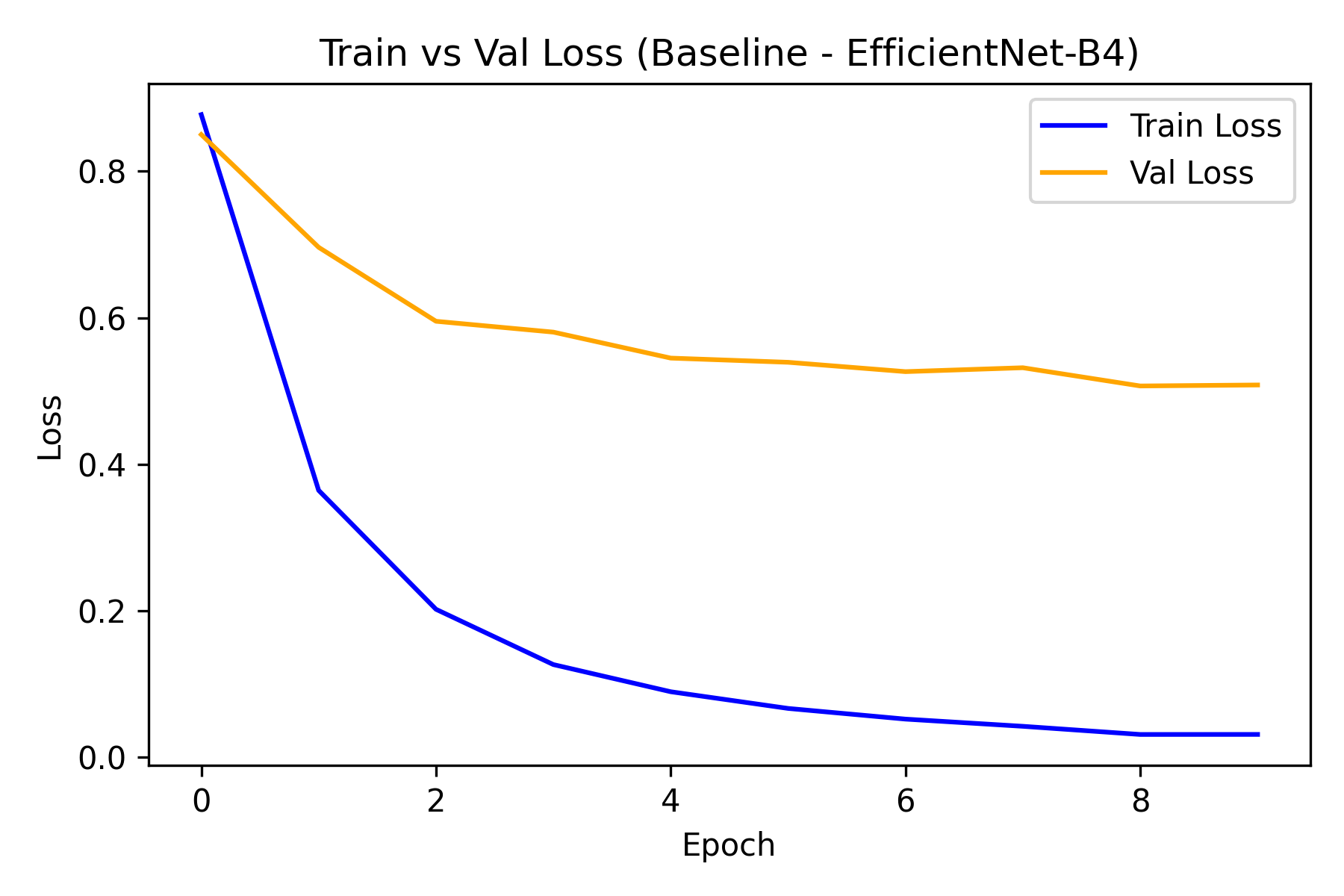}
        \caption{SqueezeNet1 Baseline - Loss}
    \end{subfigure}
    \hfill
    \begin{subfigure}[t]{0.24\textwidth}
        \centering
        \includegraphics[width=\textwidth]{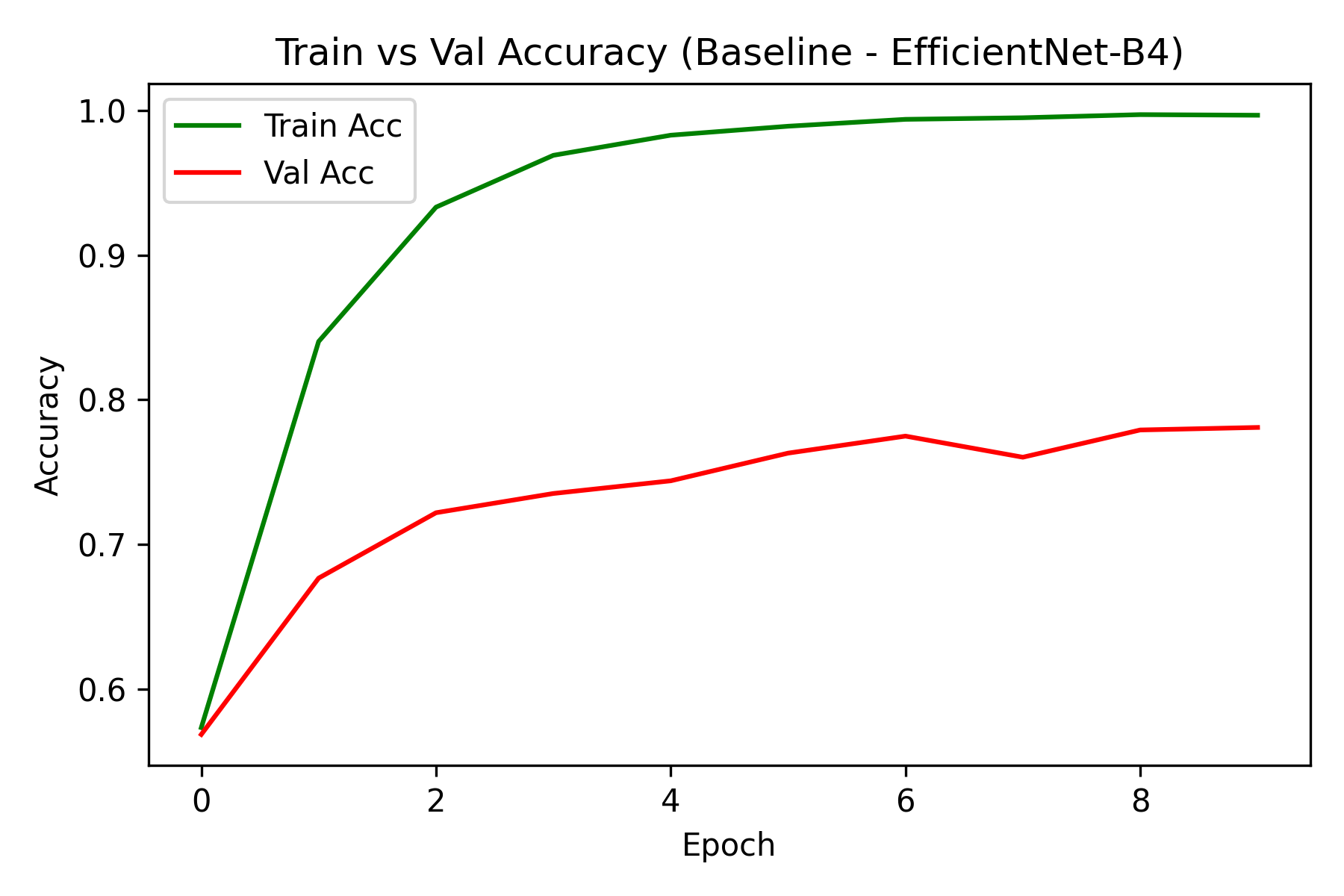}
        \caption{SqueezeNet1 Baseline - Accuracy}
    \end{subfigure}
    \hfill
    \begin{subfigure}[t]{0.24\textwidth}
        \centering
        \includegraphics[width=\textwidth]{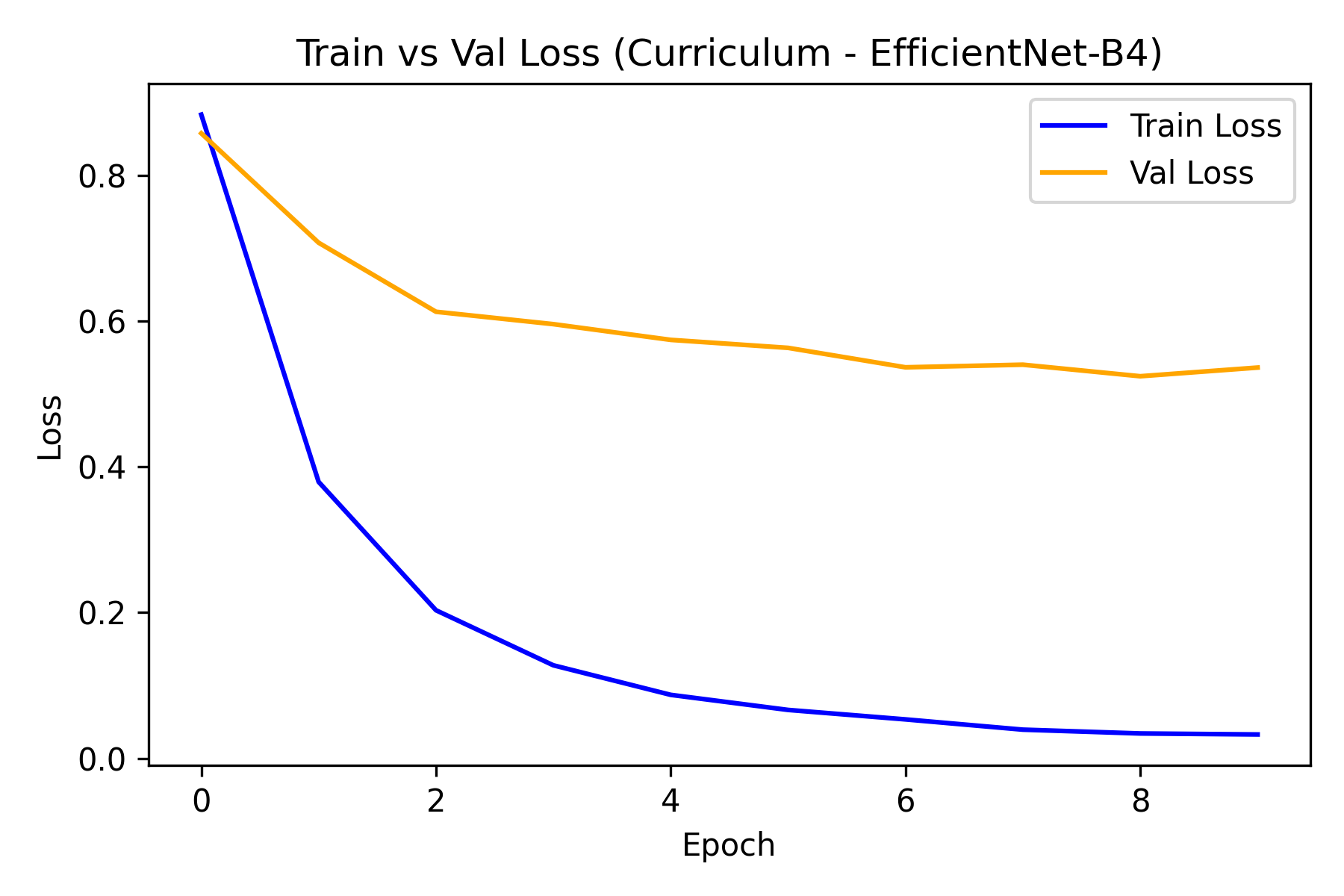}
        \caption{SqueezeNet1 Curriculum - Loss}
    \end{subfigure}
    \hfill
    \begin{subfigure}[t]{0.24\textwidth}
        \centering
        \includegraphics[width=\textwidth]{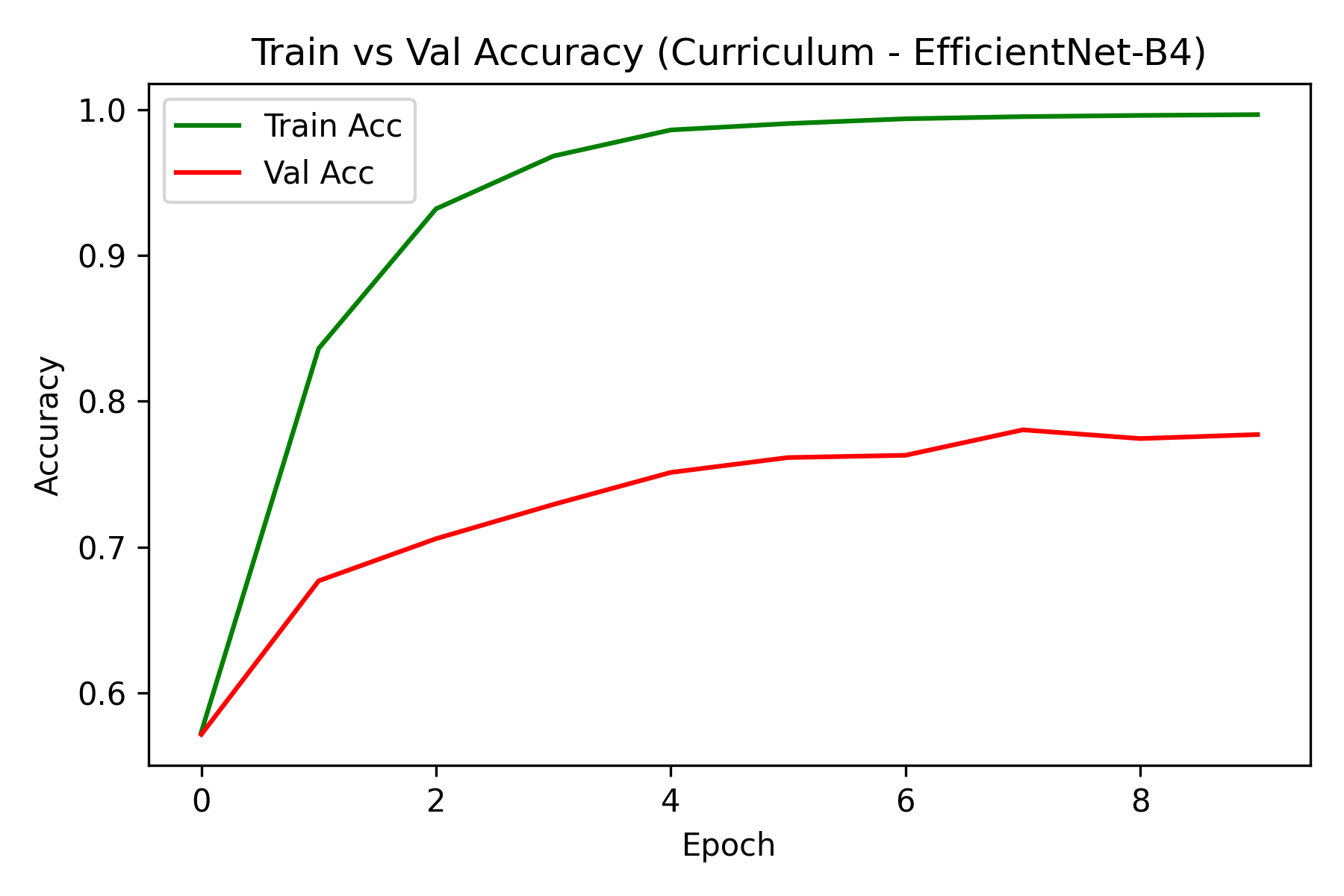}
        \caption{SqueezeNet1 Curriculum - Accuracy}
    \end{subfigure}
    
    \begin{subfigure}[t]{0.24\textwidth}
        \centering
        \includegraphics[width=\textwidth]{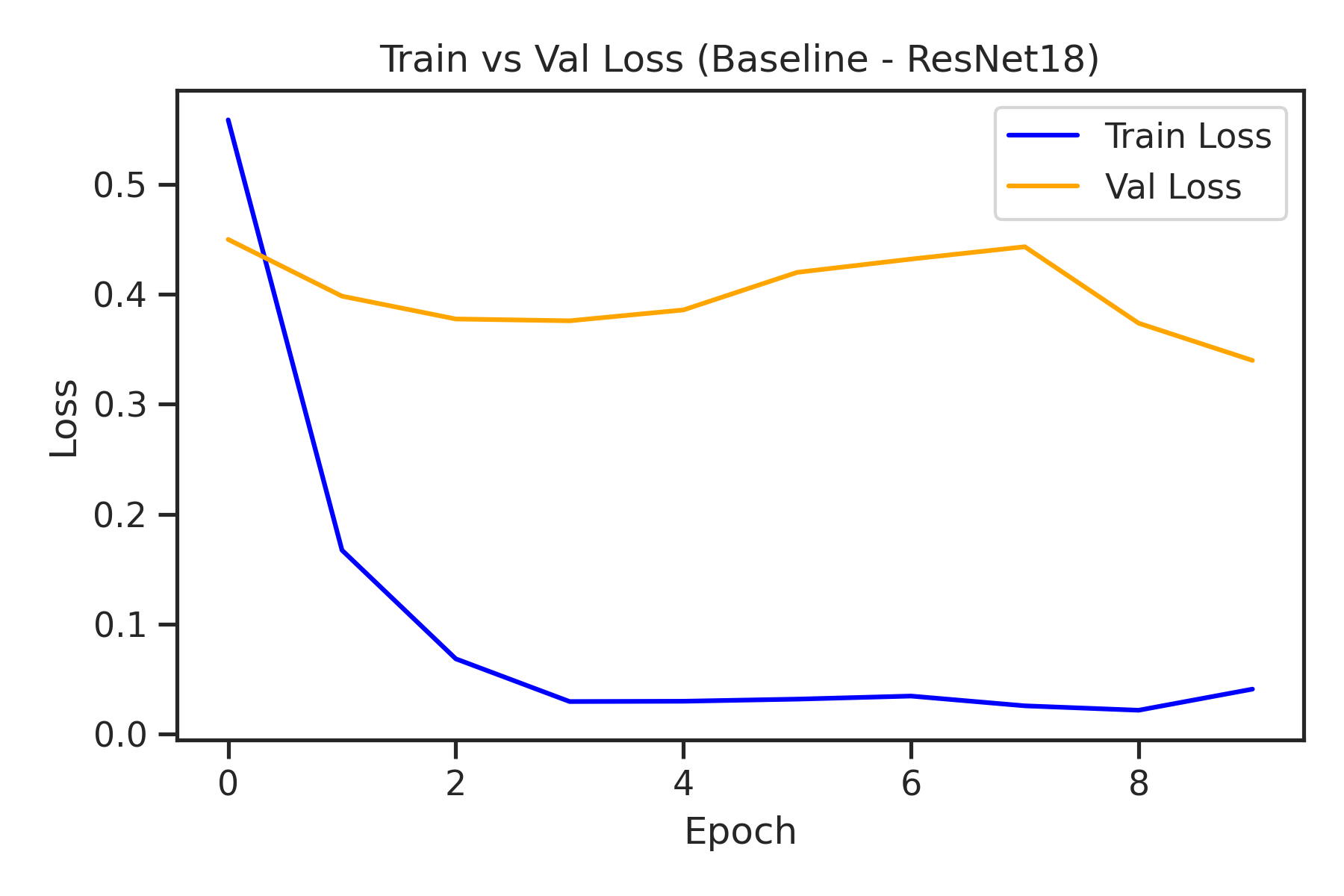}
        \caption{ResNet18 Baseline - Loss}
    \end{subfigure}
    \hfill
    \begin{subfigure}[t]{0.24\textwidth}
        \centering
        \includegraphics[width=\textwidth]{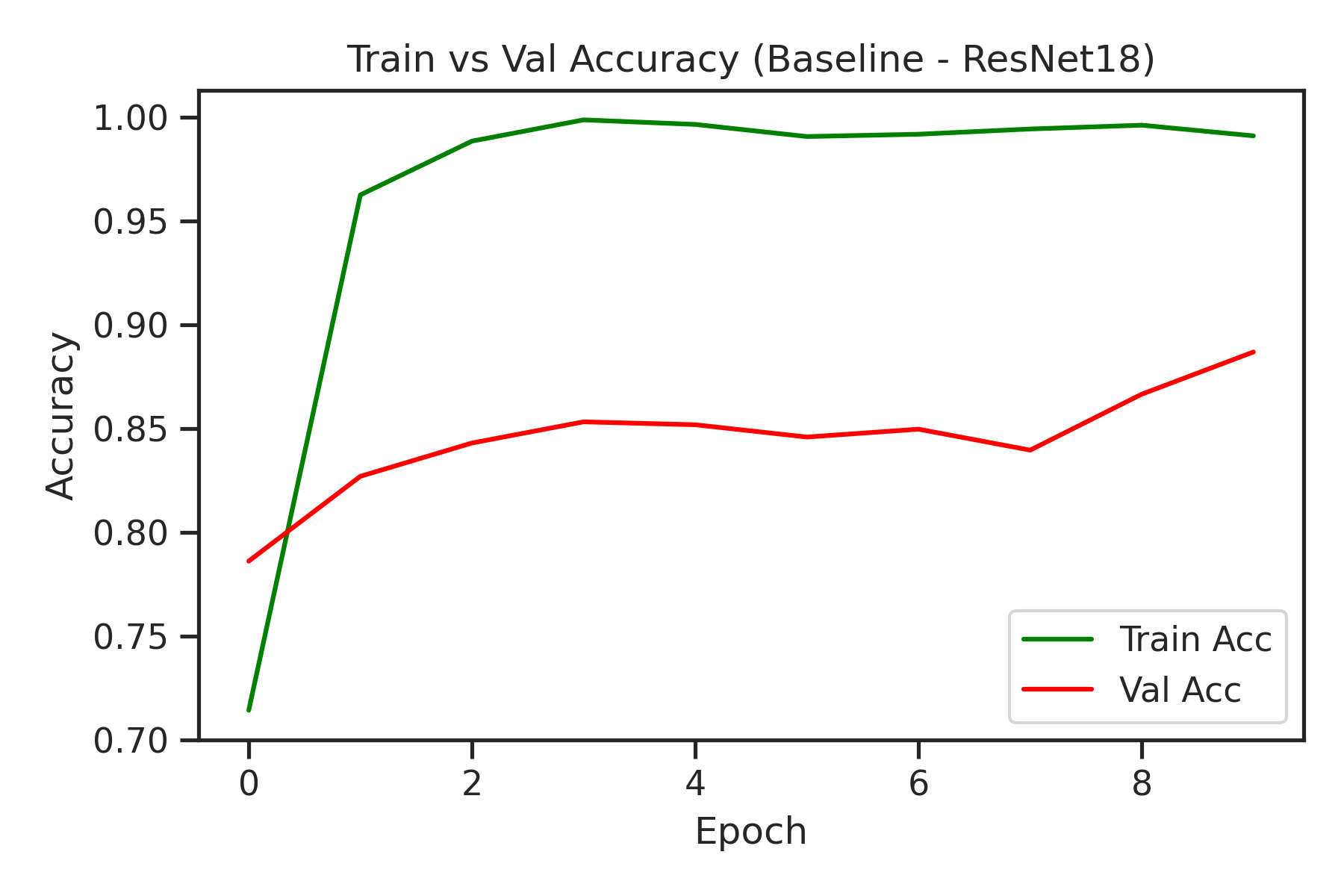}
        \caption{ResNet18 Baseline - Accuracy}
    \end{subfigure}
    \hfill
    \begin{subfigure}[t]{0.24\textwidth}
        \centering
        \includegraphics[width=\textwidth]{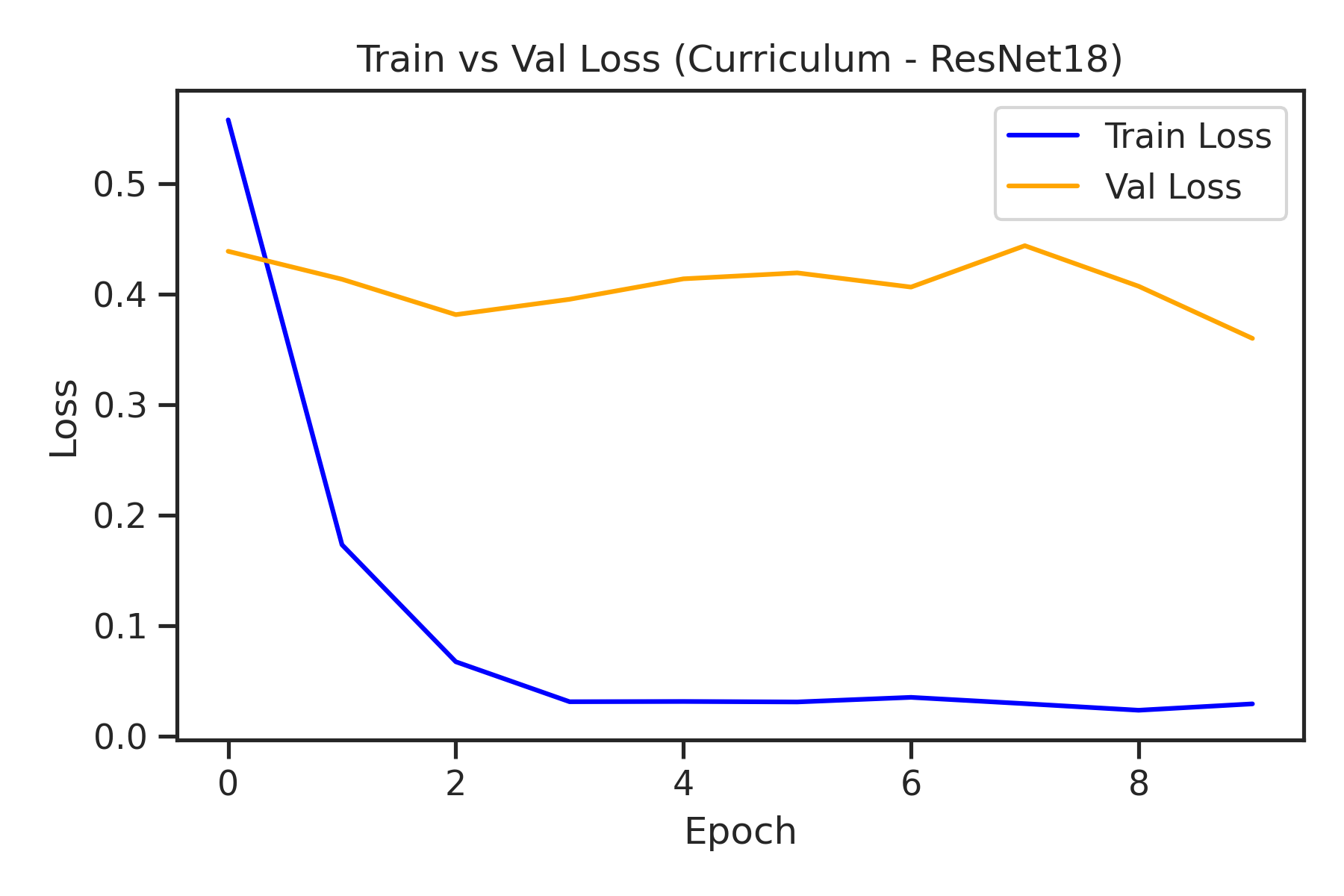}
        \caption{ResNet18 Curriculum - Loss}
    \end{subfigure}
    \hfill
    \begin{subfigure}[t]{0.24\textwidth}
        \centering
        \includegraphics[width=\textwidth]{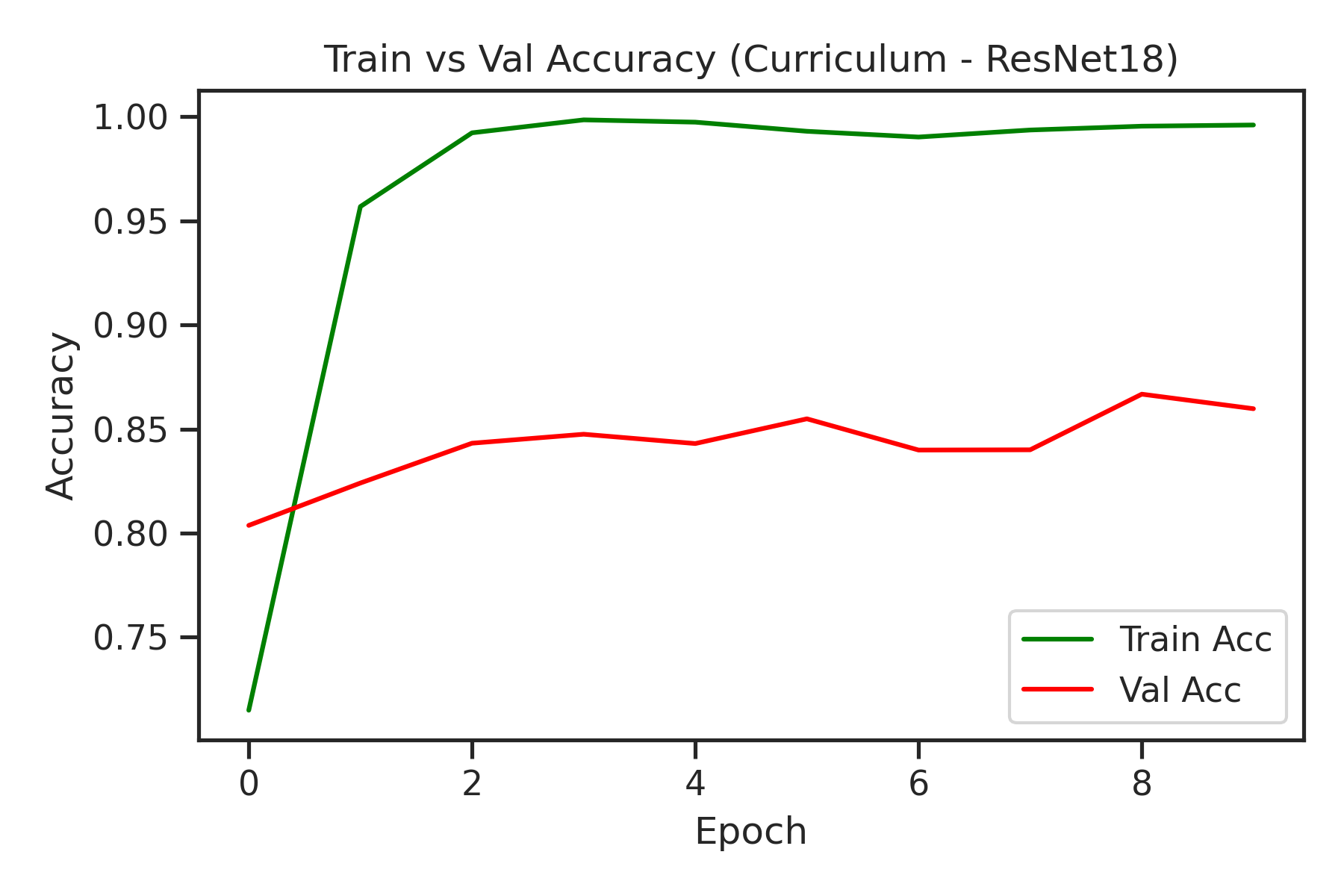}
        \caption{ResNet18 Curriculum - Accuracy}
    \end{subfigure}

    \begin{subfigure}[t]{0.24\textwidth}
    \centering
    \includegraphics[width=\textwidth]{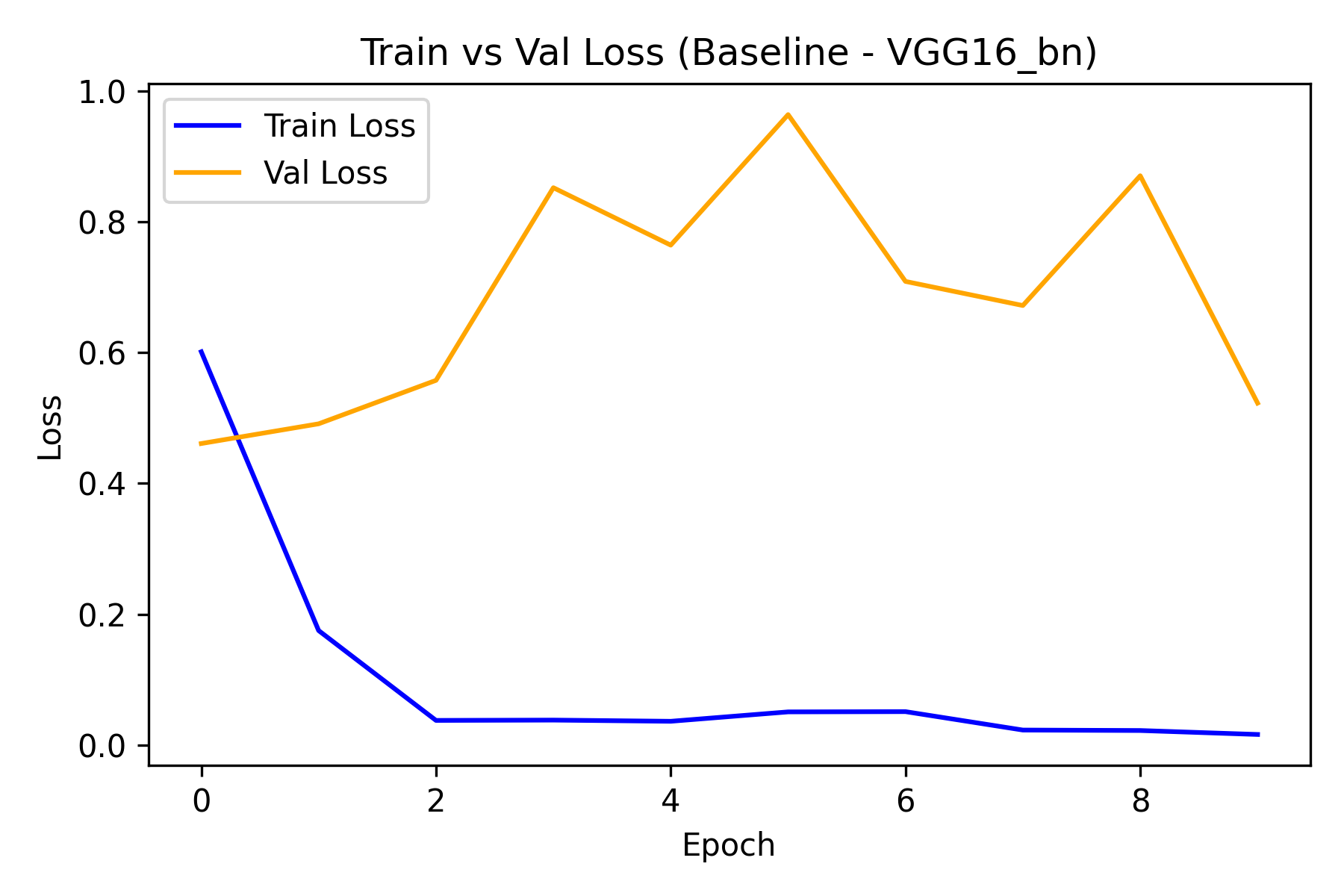}
    \caption{VGG16\_bn Baseline - Loss}
    \end{subfigure}
    \hfill
    \begin{subfigure}[t]{0.24\textwidth}
    \centering
    \includegraphics[width=\textwidth]{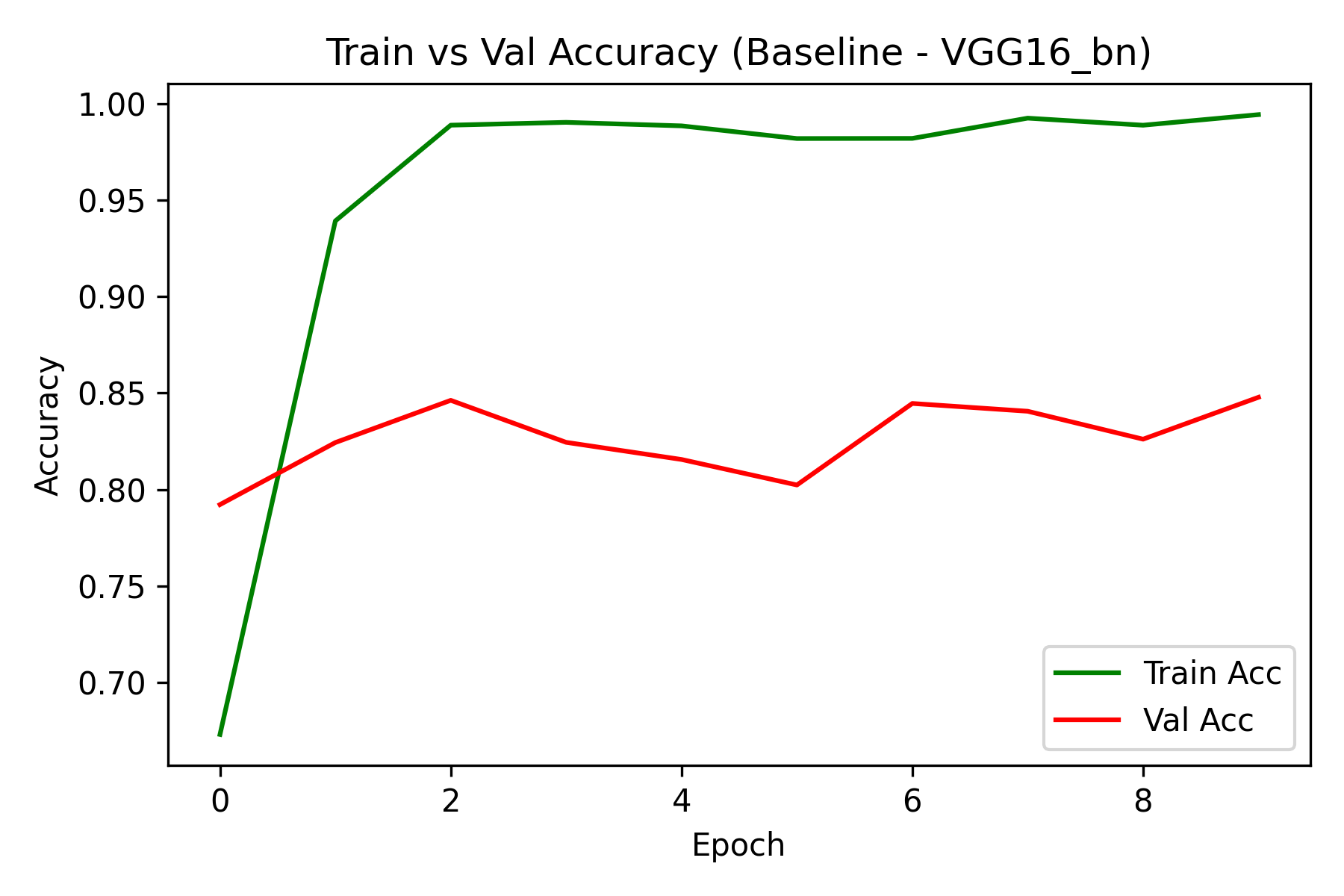}
    \caption{VGG16\_bn Baseline - Accuracy}
    \end{subfigure}
    \hfill
    \begin{subfigure}[t]{0.24\textwidth}
    \centering
    \includegraphics[width=\textwidth]{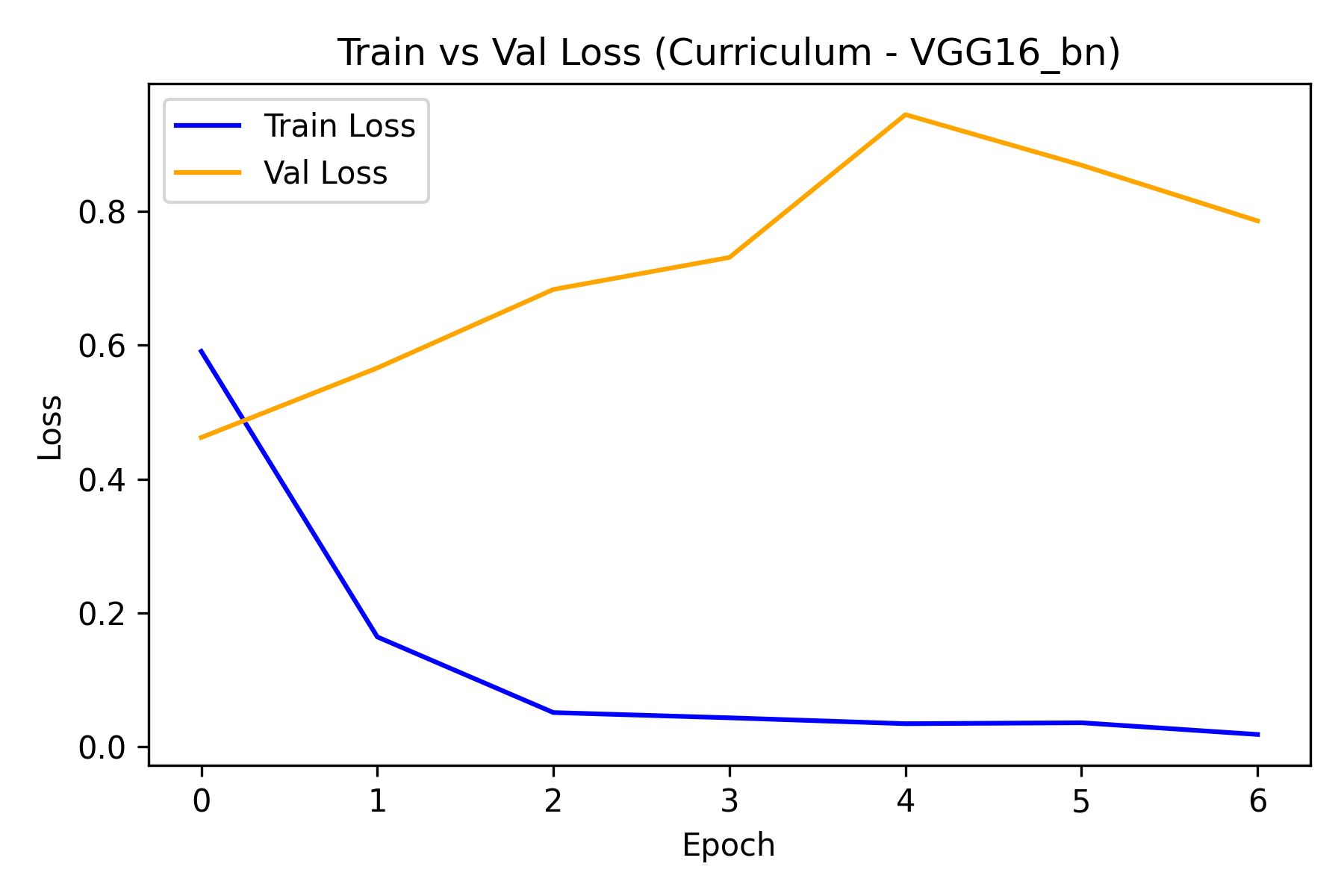}
    \caption{VGG16\_bn Curriculum - Loss}
    \end{subfigure}
    \hfill
    \begin{subfigure}[t]{0.24\textwidth}
    \centering
    \includegraphics[width=\textwidth]{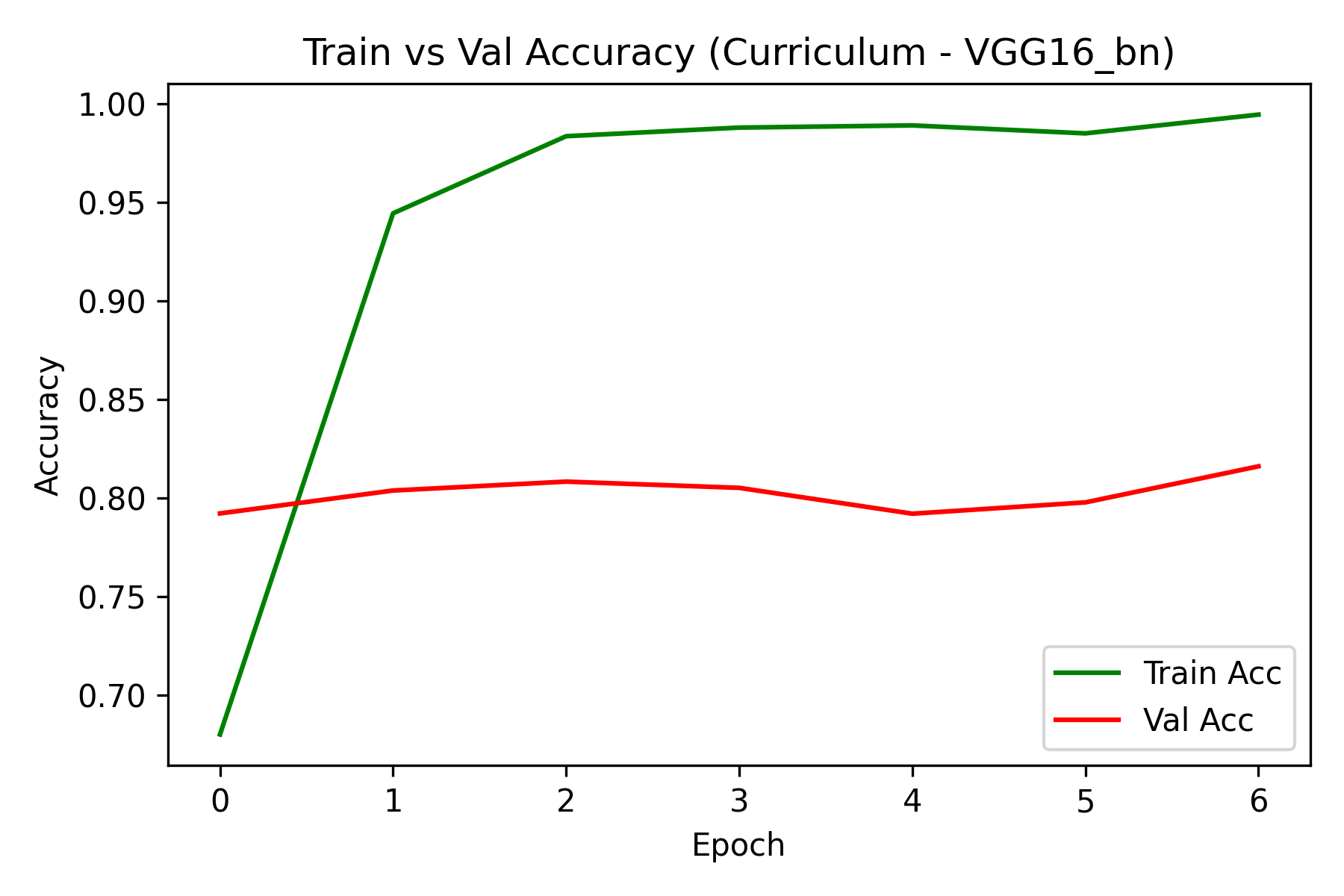}
    \caption{VGG16\_bn Curriculum - Accuracy}
    \end{subfigure}

    \begin{figure}[H]
    \centering
    \begin{subfigure}[t]{0.24\textwidth}
    \centering
    \includegraphics[width=\textwidth]{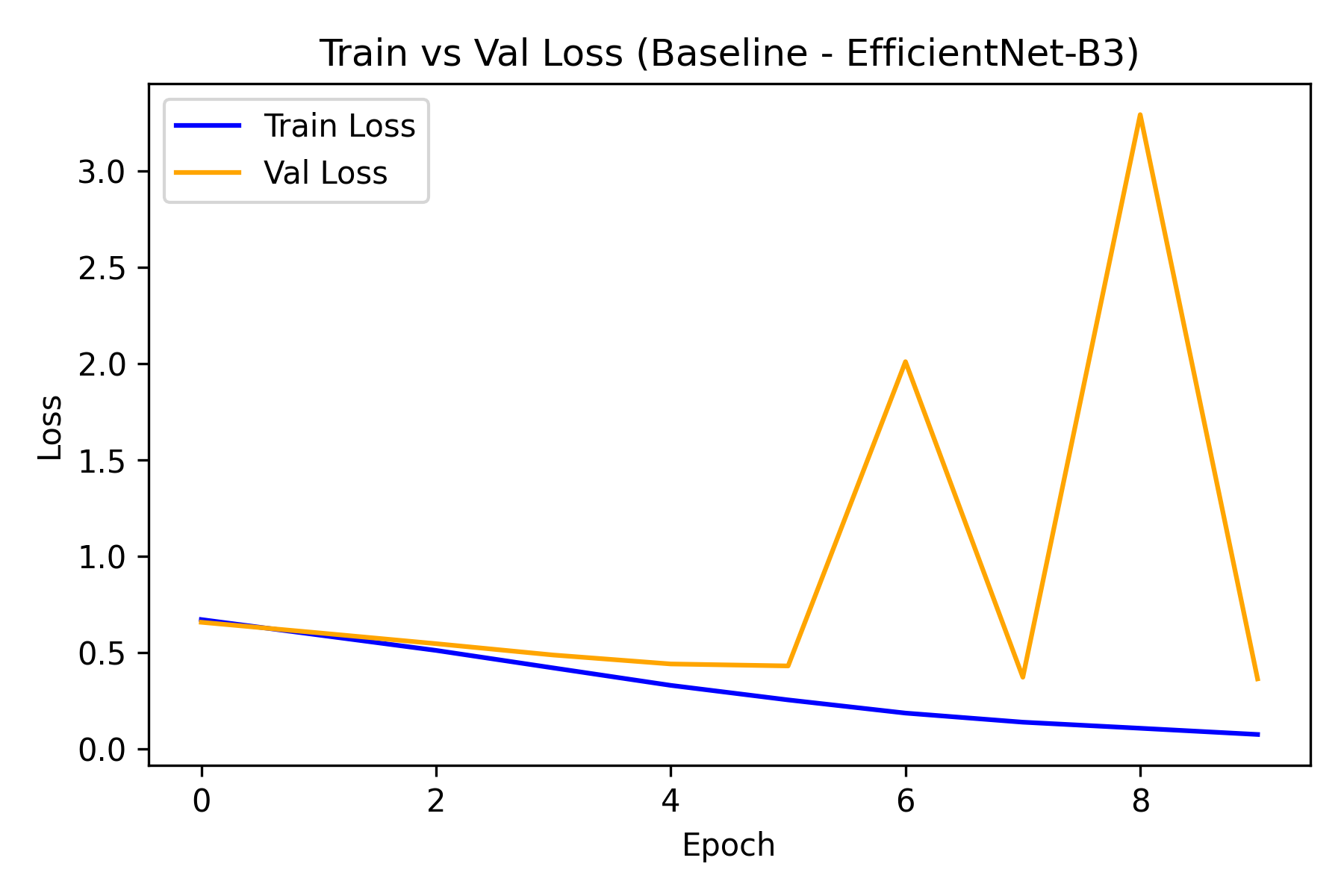}
    \caption{EfficientNetB3 Baseline - Loss}
    \end{subfigure}
    \hfill
    \begin{subfigure}[t]{0.24\textwidth}
    \centering
    \includegraphics[width=\textwidth]{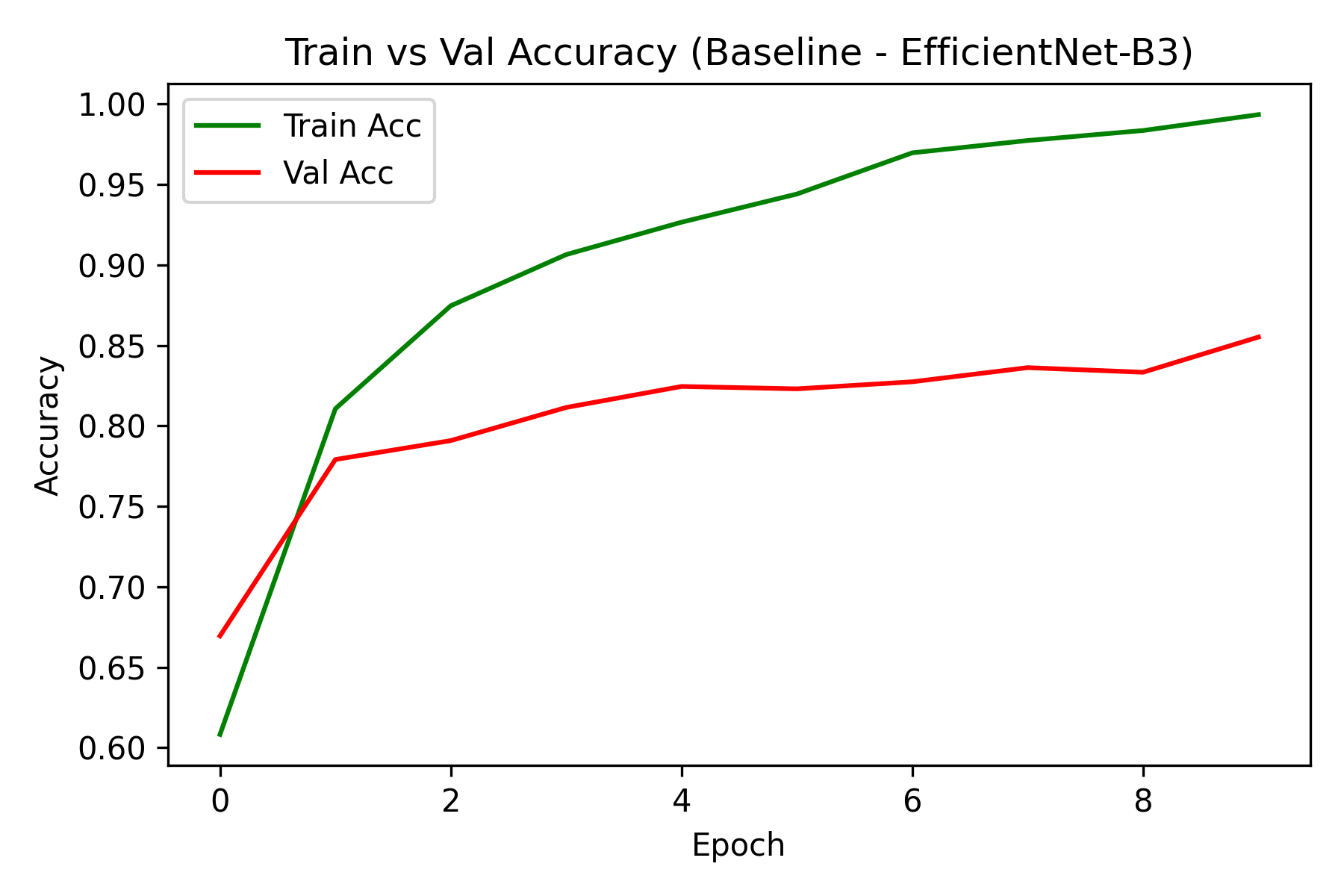}
    \caption{EfficientNetB3 Baseline - Accuracy}
    \end{subfigure}
    \hfill
    \begin{subfigure}[t]{0.24\textwidth}
    \centering
    \includegraphics[width=\textwidth]{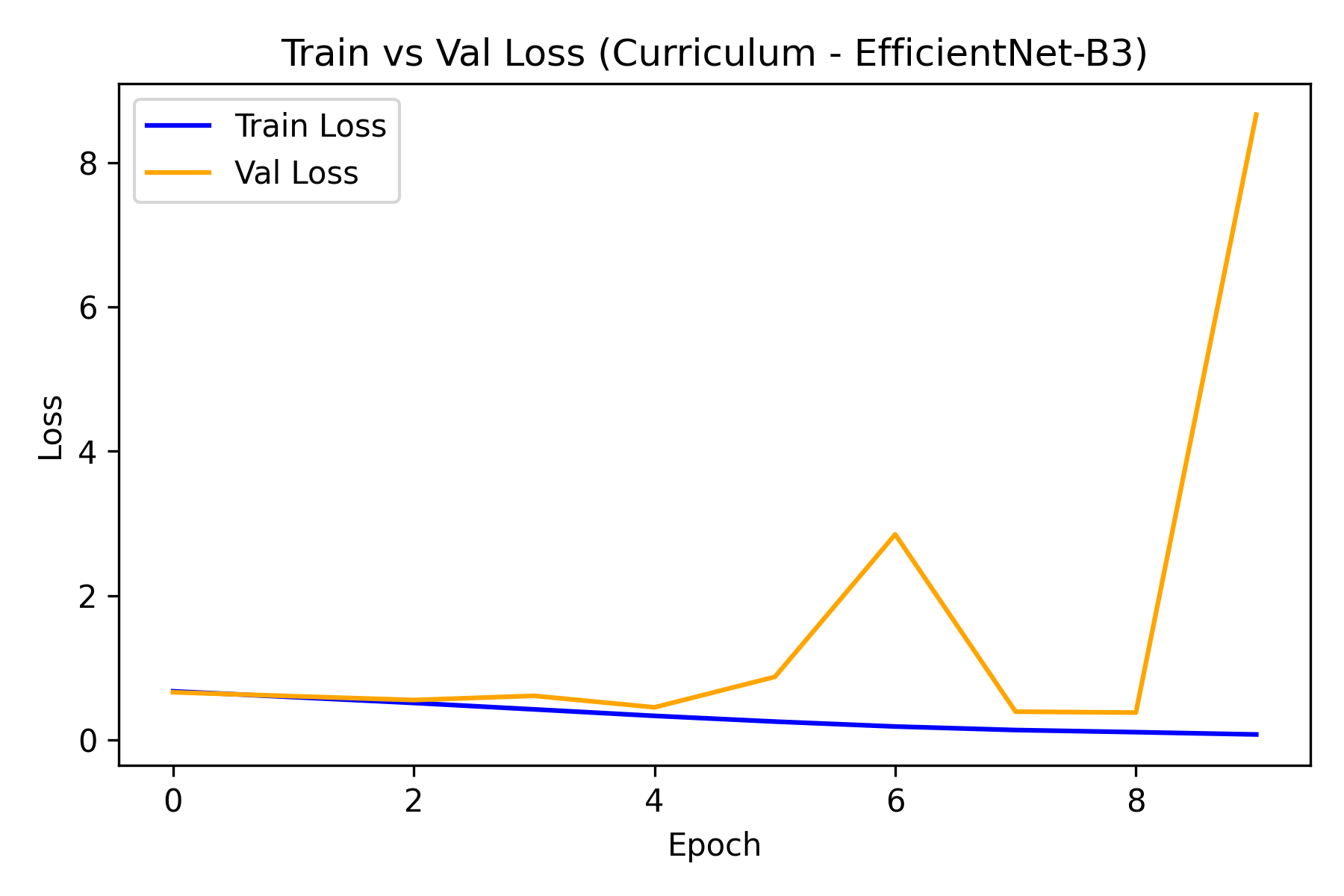}
    \caption{EfficientNetB3 Curriculum - Loss}
    \end{subfigure}
    \hfill
    \begin{subfigure}[t]{0.24\textwidth}
    \centering
    \includegraphics[width=\textwidth]{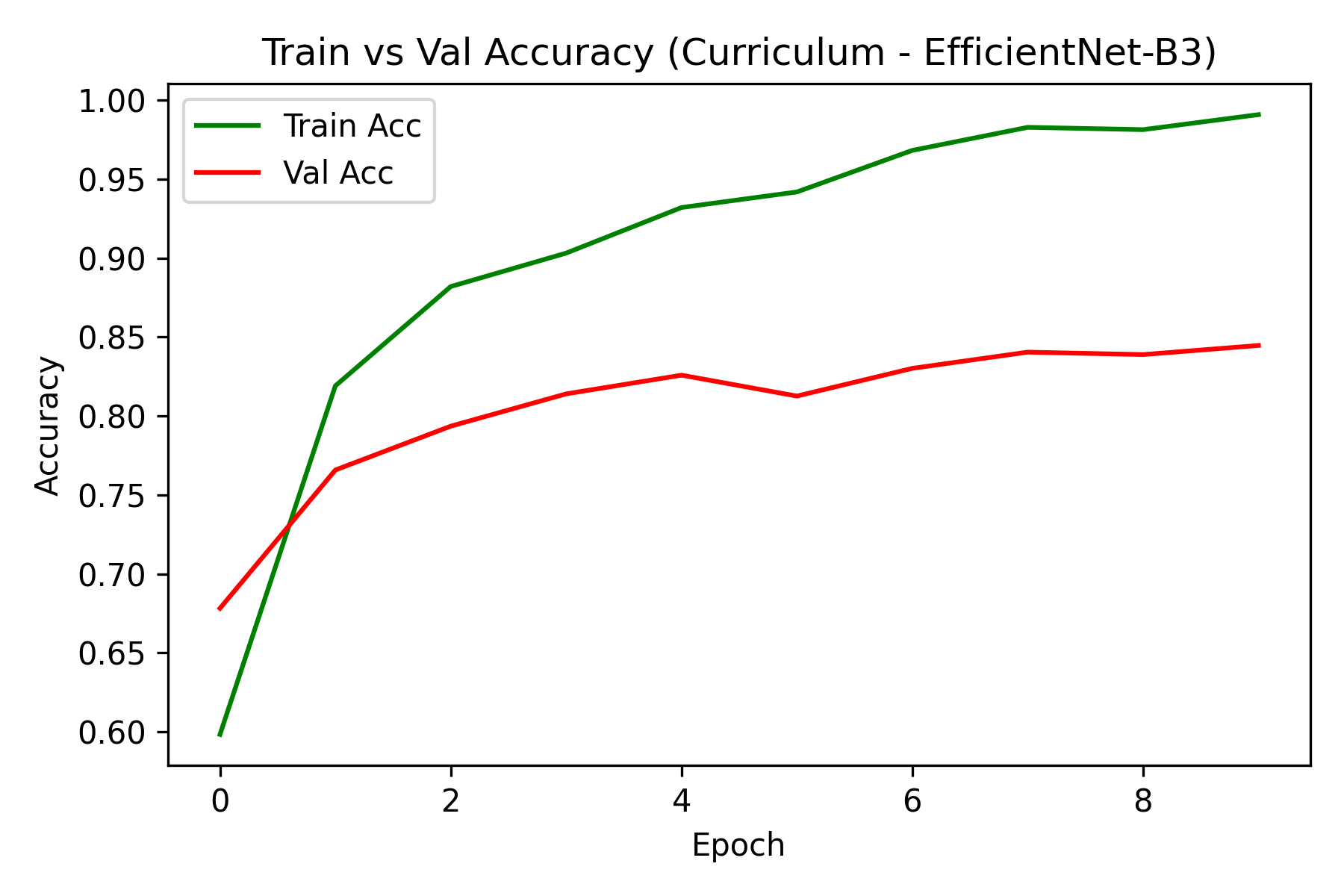}
    \caption{EfficientNetB3 Curriculum - Accuracy}
    \end{subfigure}
    \end{figure}
    \caption*{\textbf{Supplementary Figure S1A.} Training curves (Part 1 of 2).}
\end{figure}

\begin{figure}[H]
    \centering
    \begin{subfigure}[t]{0.24\textwidth}
        \centering
        \includegraphics[width=\textwidth]{efficientnet_b4_baseline_train_val_loss.png}
        \caption{EfficientNet-B4 Baseline - Loss}
    \end{subfigure}
    \hfill
    \begin{subfigure}[t]{0.24\textwidth}
        \centering
        \includegraphics[width=\textwidth]{efficientnet_b4_baseline_train_val_acc.png}
        \caption{EfficientNet-B4 Baseline - Accuracy}
    \end{subfigure}
    \hfill
    \begin{subfigure}[t]{0.24\textwidth}
        \centering
        \includegraphics[width=\textwidth]{efficientnet_b4_curriculum_train_val_loss.png}
        \caption{(aa) EfficientNet-B4 Curriculum - Loss}
    \end{subfigure}
    \hfill
    \begin{subfigure}[t]{0.24\textwidth}
        \centering
        \includegraphics[width=\textwidth]{efficientnet_b4_curriculum_train_val_acc.png}
        \caption{EfficientNet-B4 Curriculum - Accuracy}
    \end{subfigure}
    
    \begin{subfigure}[t]{0.24\textwidth}
        \centering
        \includegraphics[width=\textwidth]{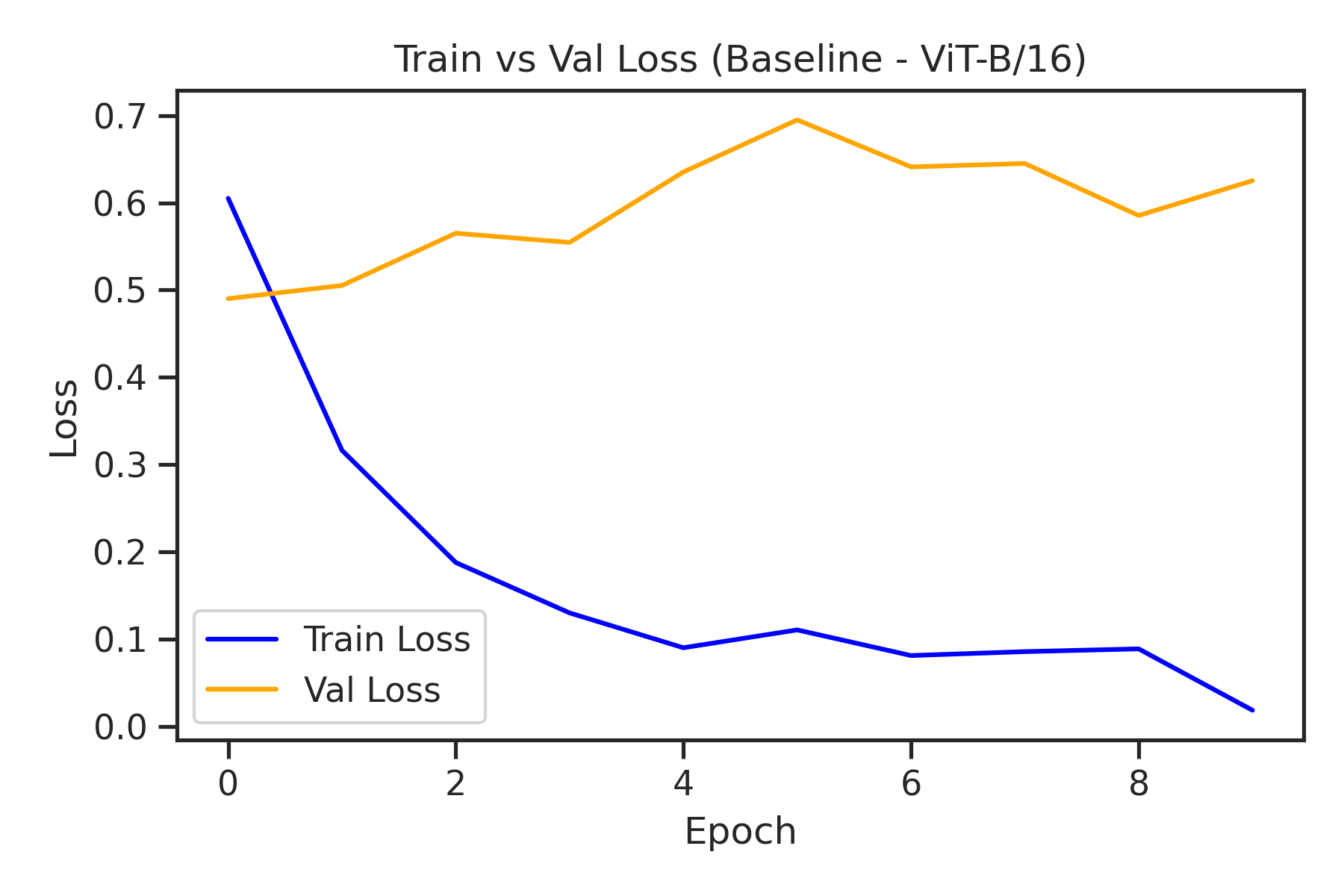}
        \caption{ViT-B/16 Baseline - Loss}
    \end{subfigure}
    \hfill
    \begin{subfigure}[t]{0.24\textwidth}
        \centering
        \includegraphics[width=\textwidth]{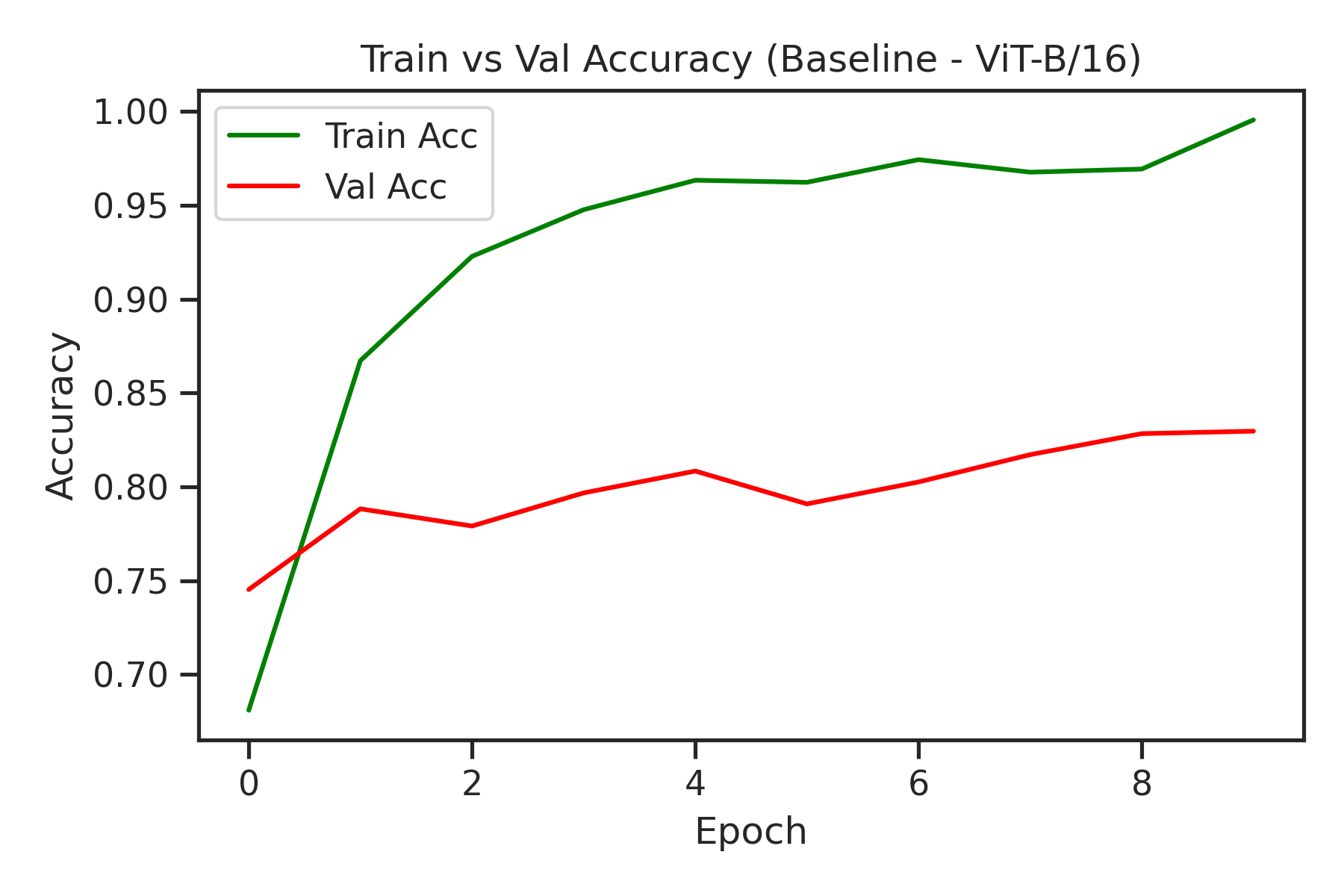}
        \caption{ViT-B/16 Baseline - Accuracy}
    \end{subfigure}
    \hfill
    \begin{subfigure}[t]{0.24\textwidth}
        \centering
        \includegraphics[width=\textwidth]{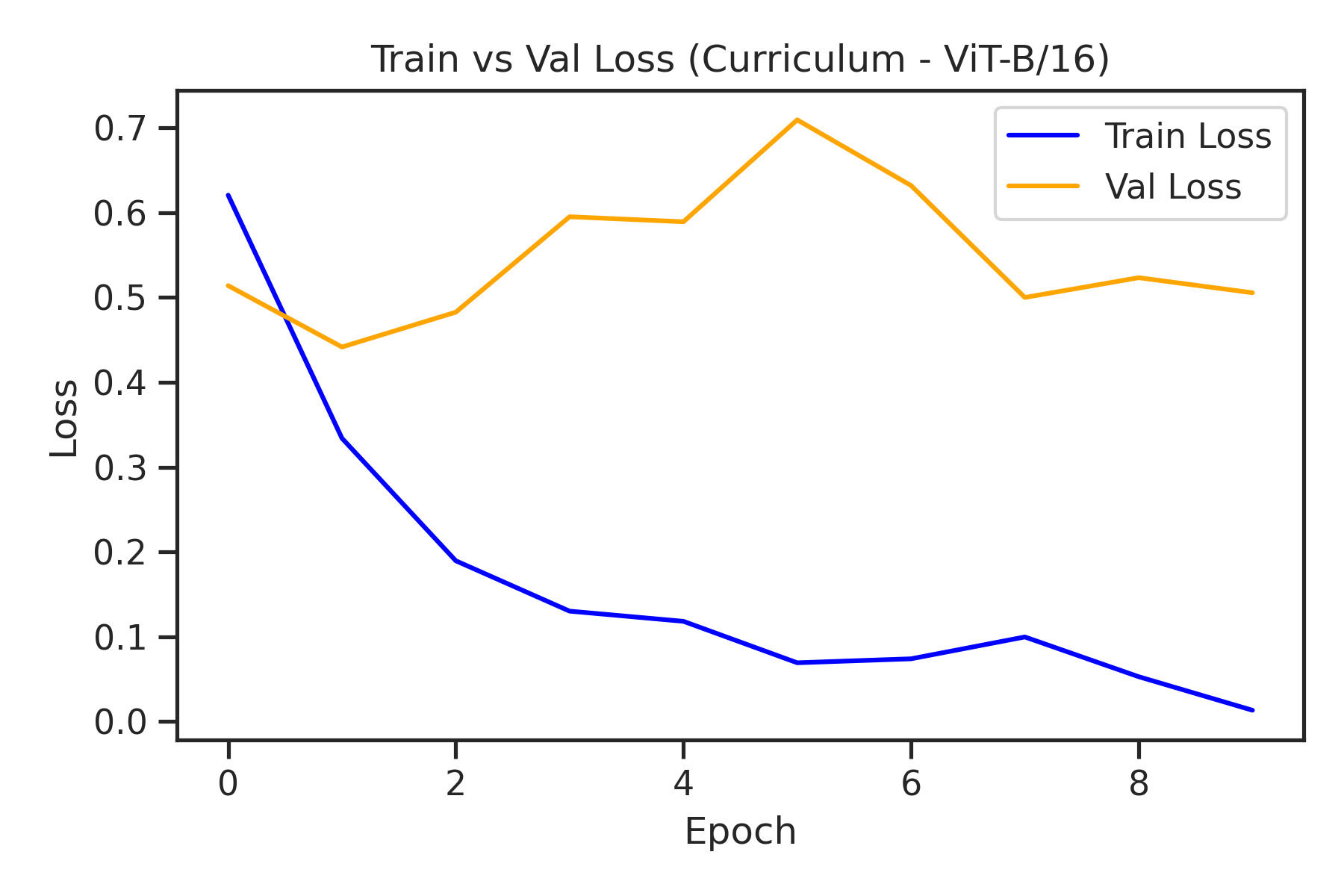}
        \caption{ViT-B/16 Curriculum - Loss}
    \end{subfigure}
    \hfill
    \begin{subfigure}[t]{0.24\textwidth}
        \centering
        \includegraphics[width=\textwidth]{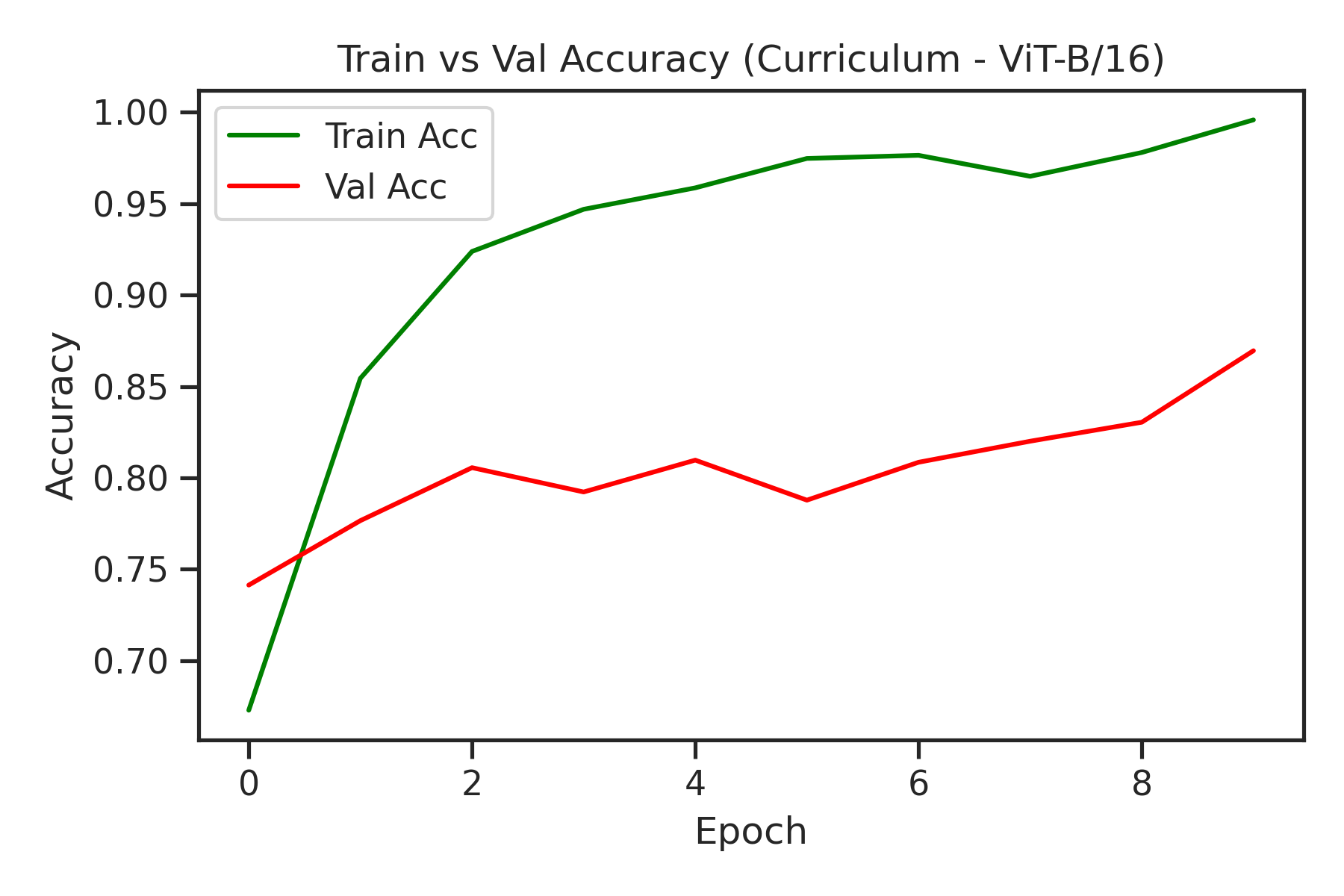}
        \caption{ViT-B/16 Curriculum - Accuracy}
    \end{subfigure}

    \begin{subfigure}[t]{0.24\textwidth}
        \centering
        \includegraphics[width=\textwidth]{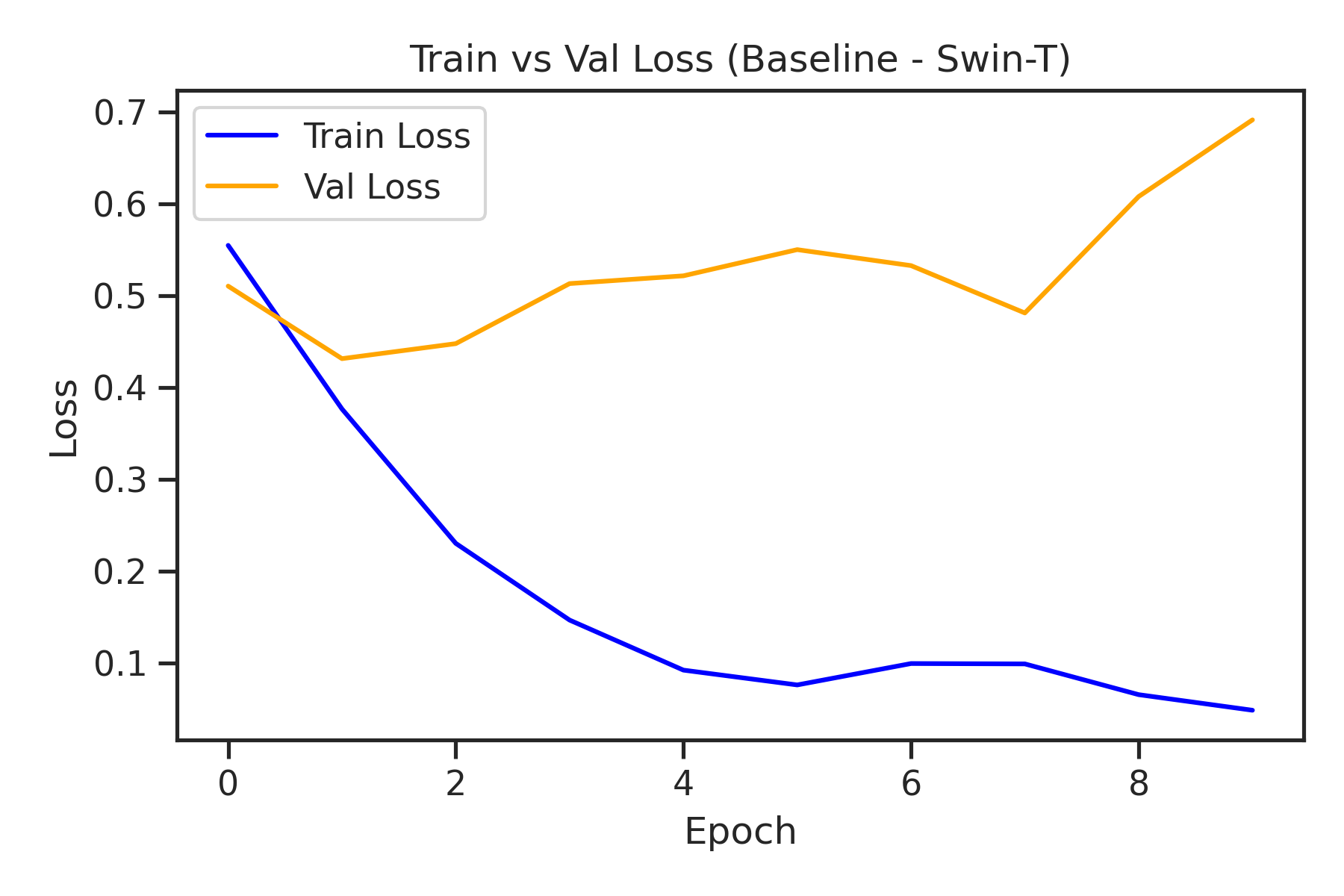}
        \caption{Swin-T Baseline - Loss}
    \end{subfigure}
    \hfill
    \begin{subfigure}[t]{0.24\textwidth}
        \centering
        \includegraphics[width=\textwidth]{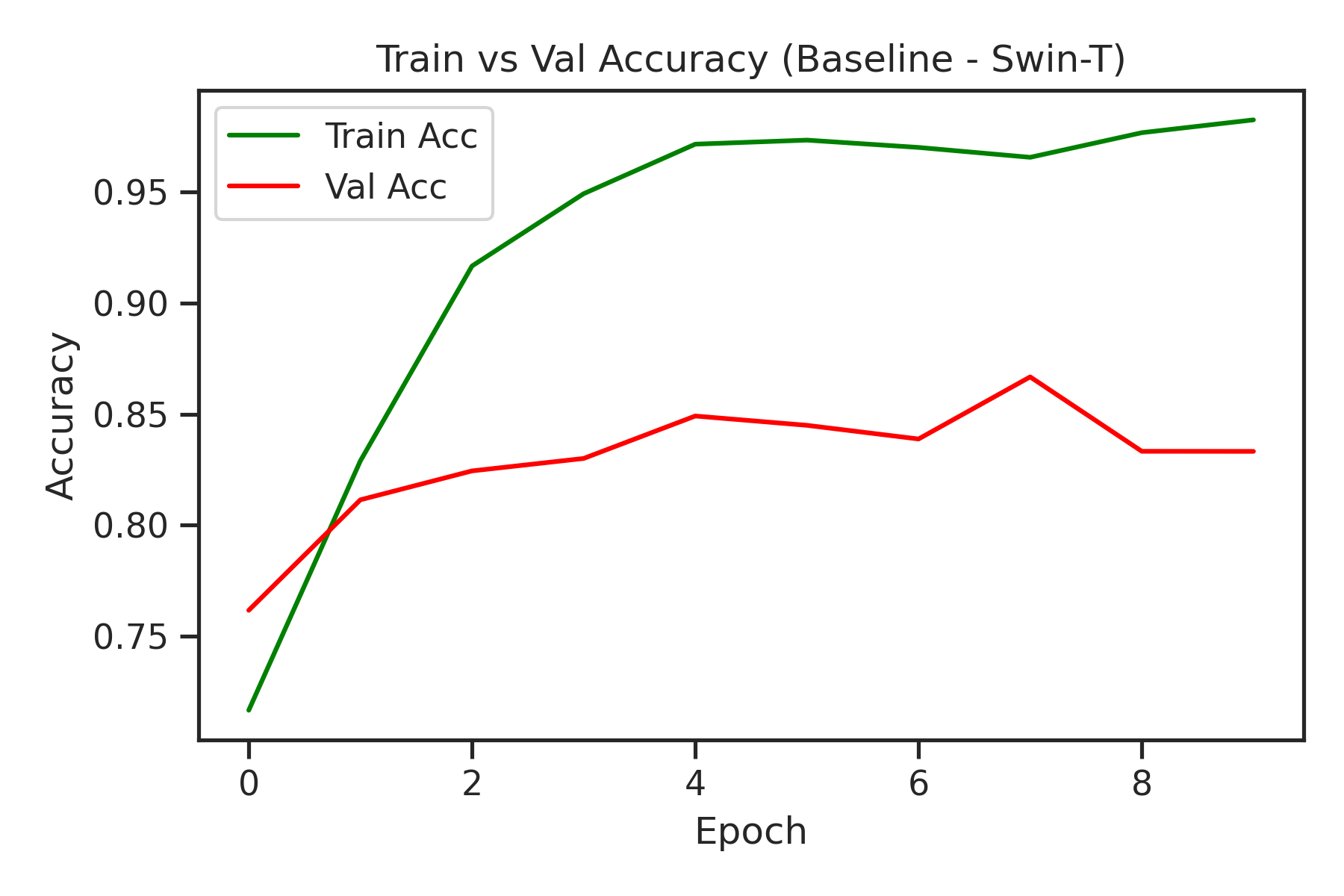}
        \caption{Swin-T Baseline - Accuracy}
    \end{subfigure}
    \hfill
    \begin{subfigure}[t]{0.24\textwidth}
        \centering
        \includegraphics[width=\textwidth]{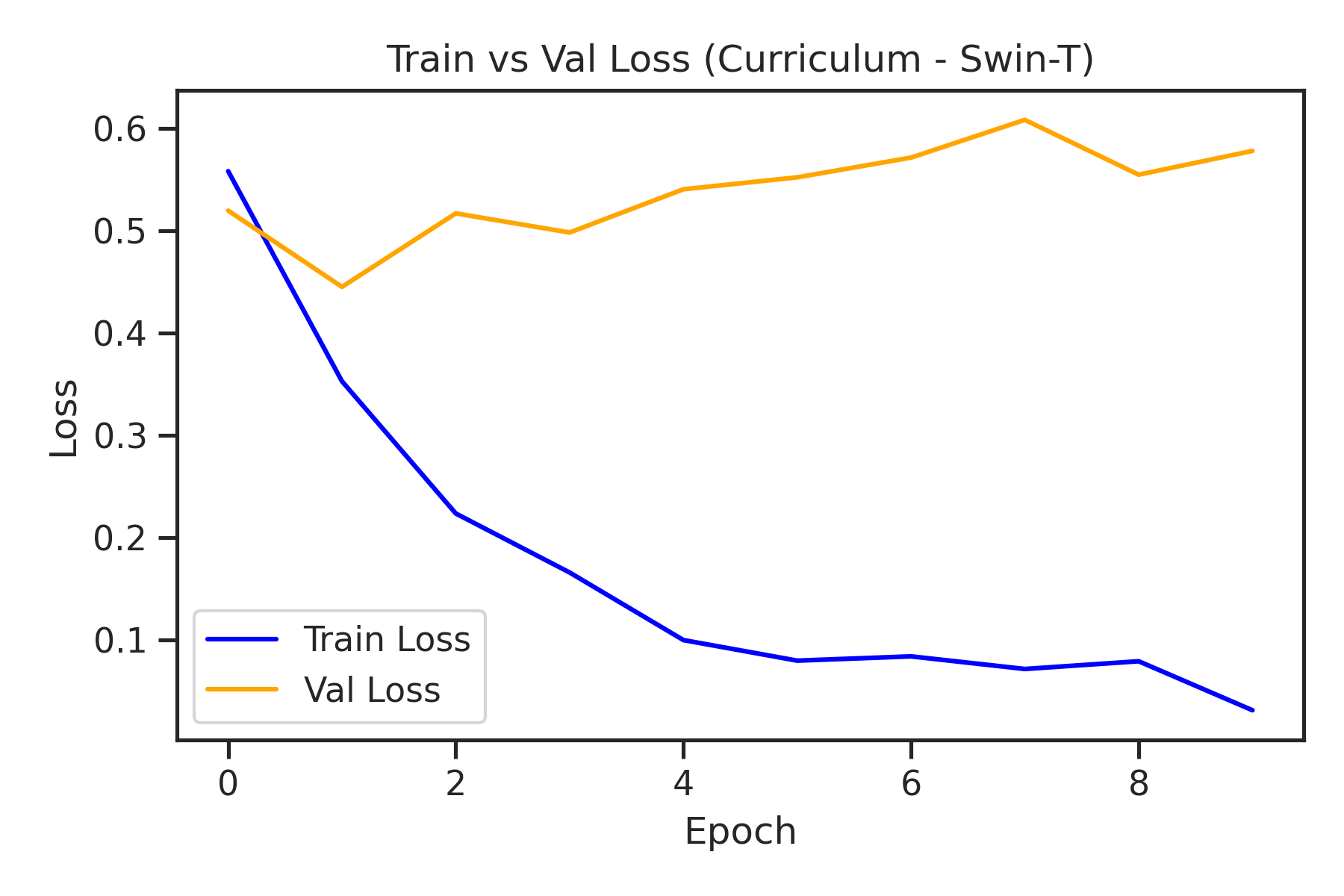}
        \caption{Swin-T Curriculum - Loss}
    \end{subfigure}
    \hfill
    \begin{subfigure}[t]{0.24\textwidth}
        \centering
        \includegraphics[width=\textwidth]{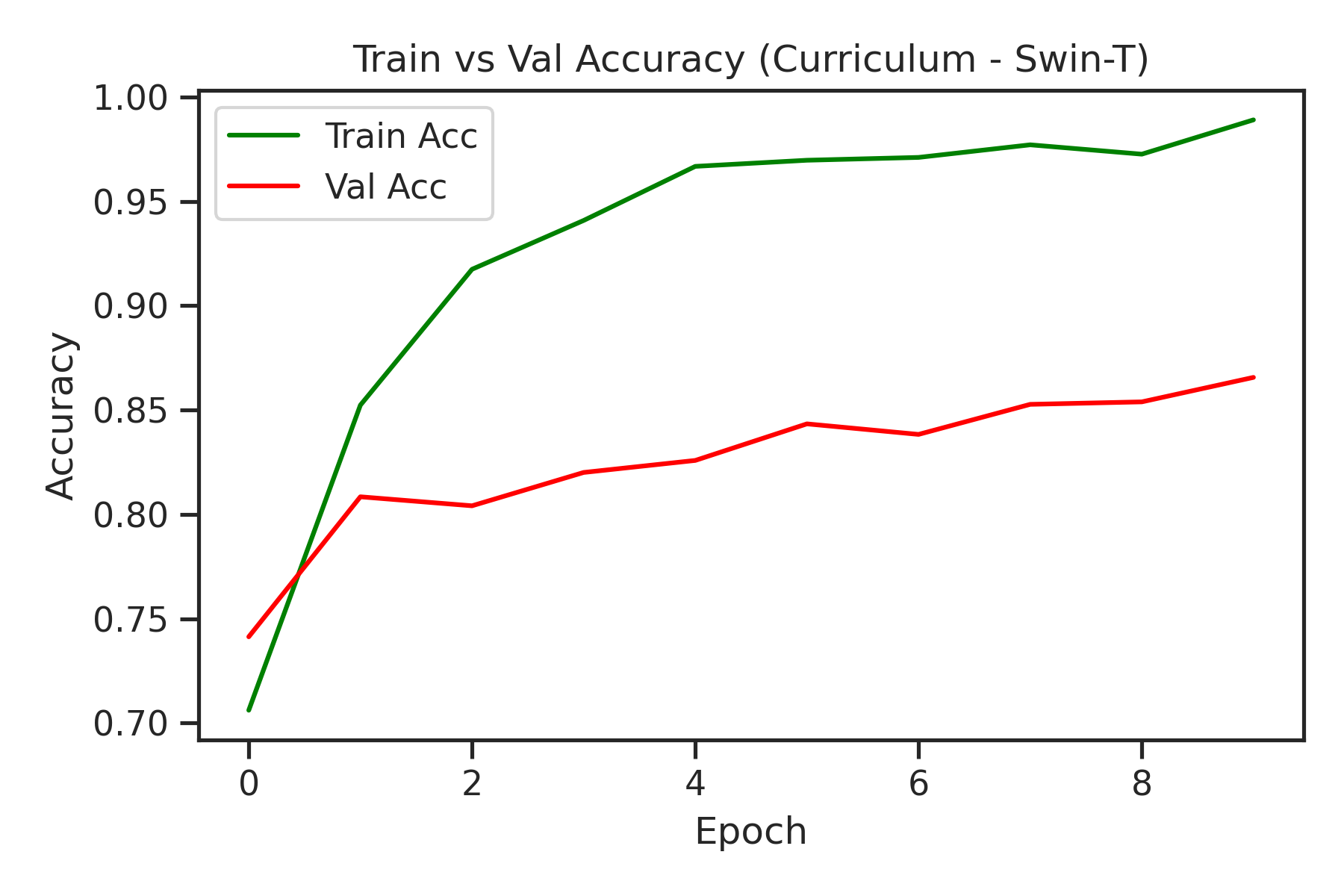}
        \caption{Swin-T Curriculum - Accuracy}
    \end{subfigure}
    
    \begin{subfigure}[t]{0.24\textwidth}
    \centering
    \includegraphics[width=\textwidth]{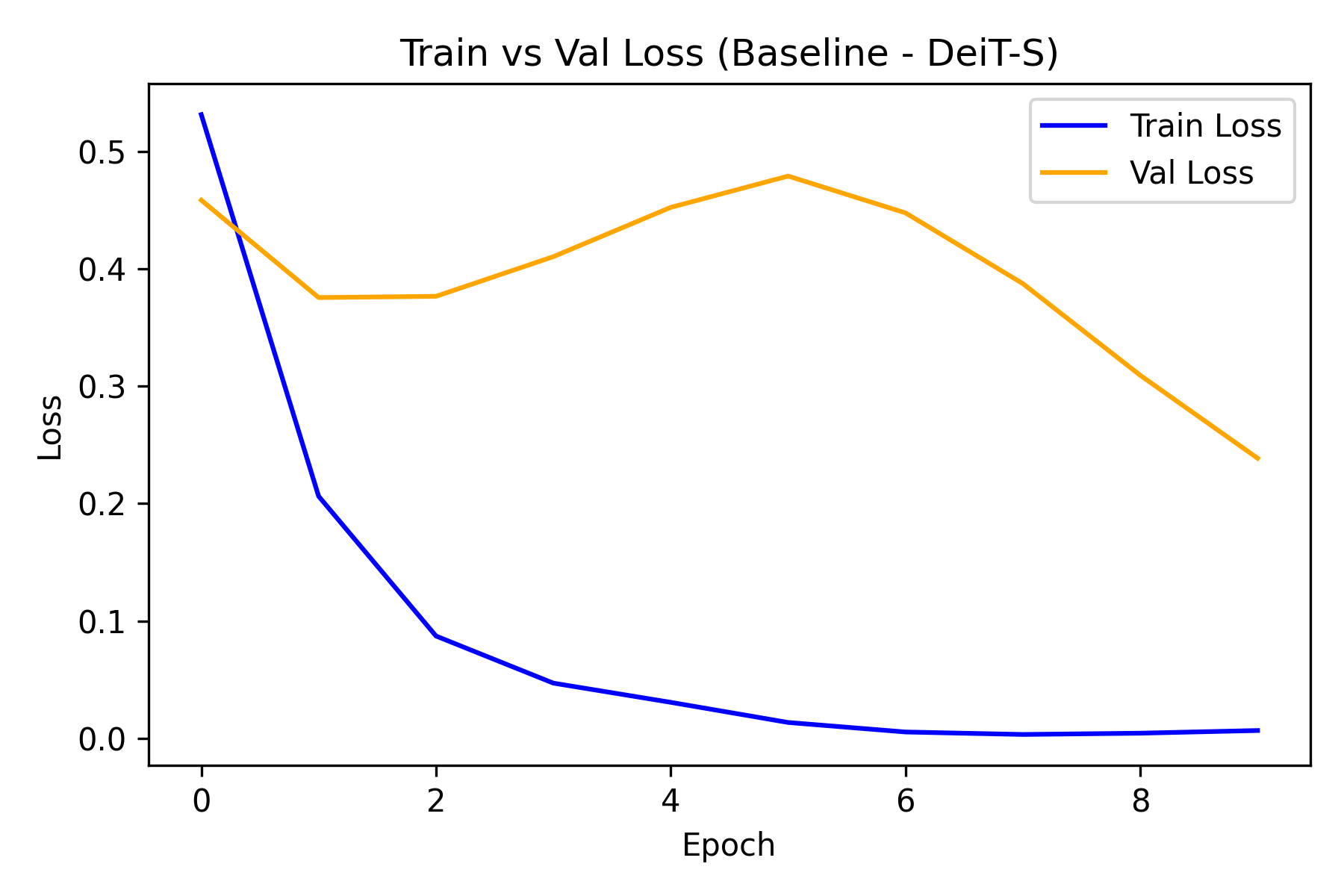}
    \caption{(ag) DeiT-S Baseline - Loss}
    \end{subfigure}
    \hfill
    \begin{subfigure}[t]{0.24\textwidth}
    \centering
    \includegraphics[width=\textwidth]{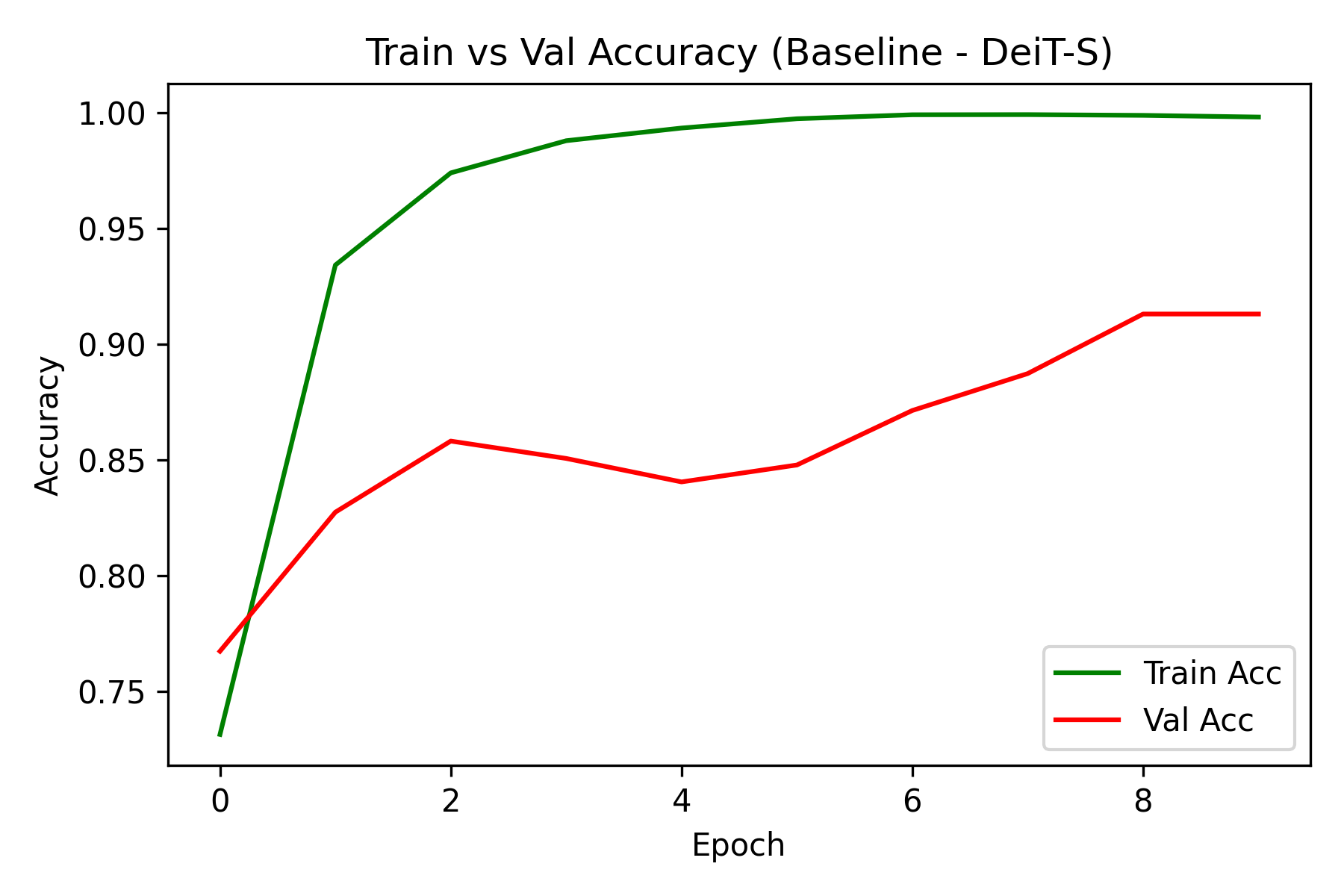}
    \caption{(ah) DeiT-S Baseline - Accuracy}
    \end{subfigure}
    \hfill
    \begin{subfigure}[t]{0.24\textwidth}
    \centering
    \includegraphics[width=\textwidth]{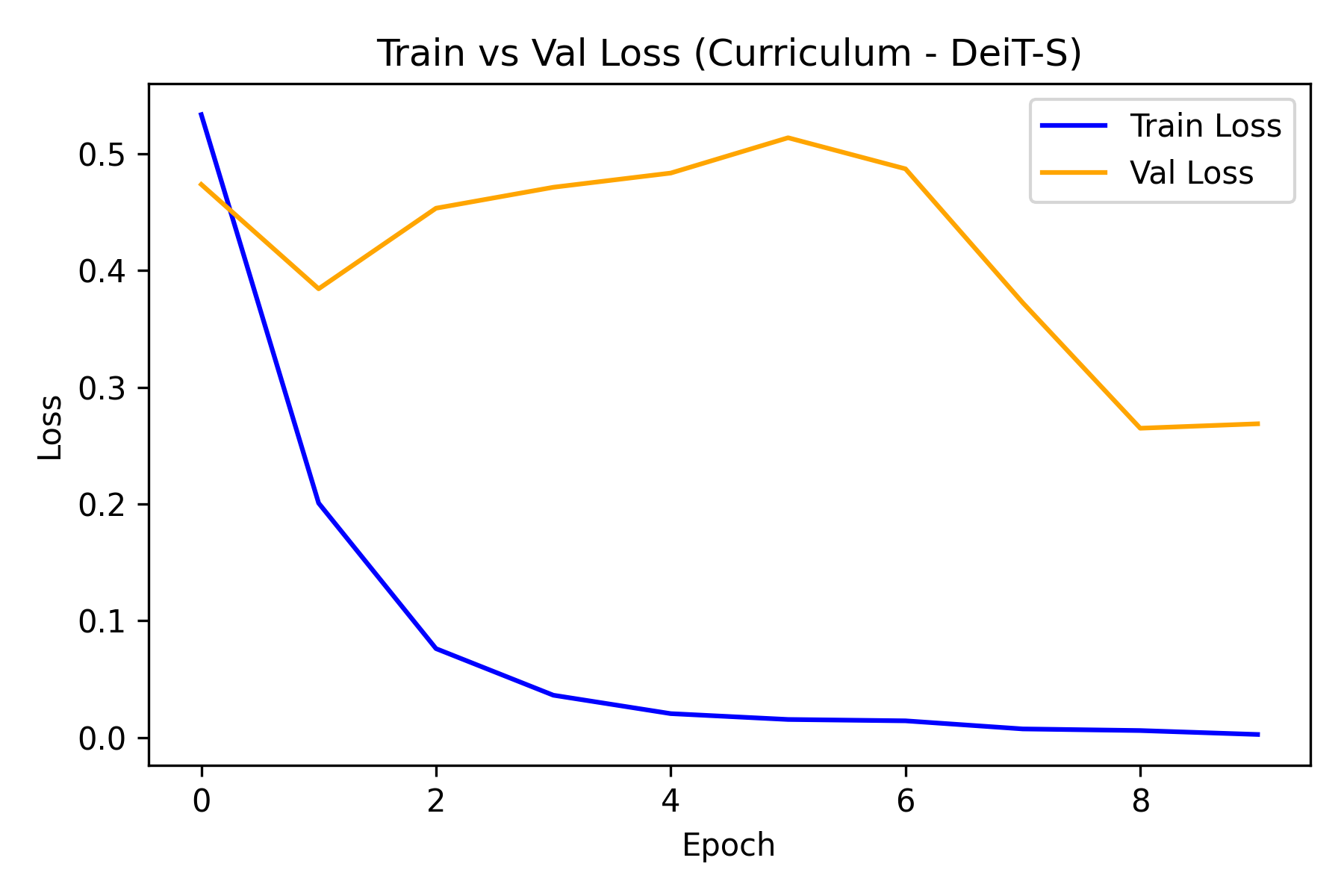}
    \caption{(ai) DeiT-S Curriculum - Loss}
    \end{subfigure}
    \hfill
    \begin{subfigure}[t]{0.24\textwidth}
    \centering
    \includegraphics[width=\textwidth]{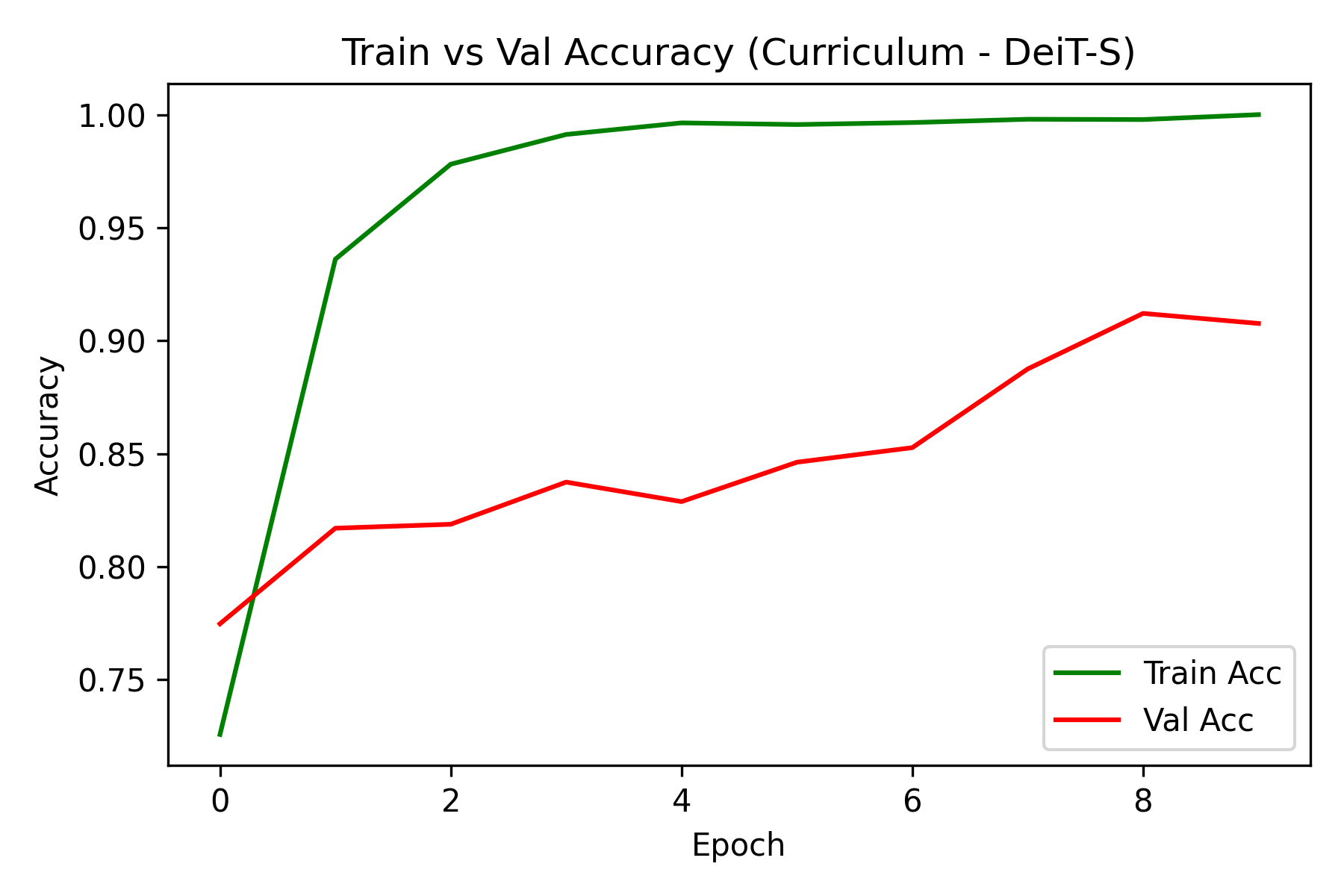}
    \caption{(aj) DeiT-S Curriculum - Accuracy}
    \end{subfigure}

    \begin{subfigure}[t]{0.24\textwidth}
        \centering
        \includegraphics[width=\textwidth]{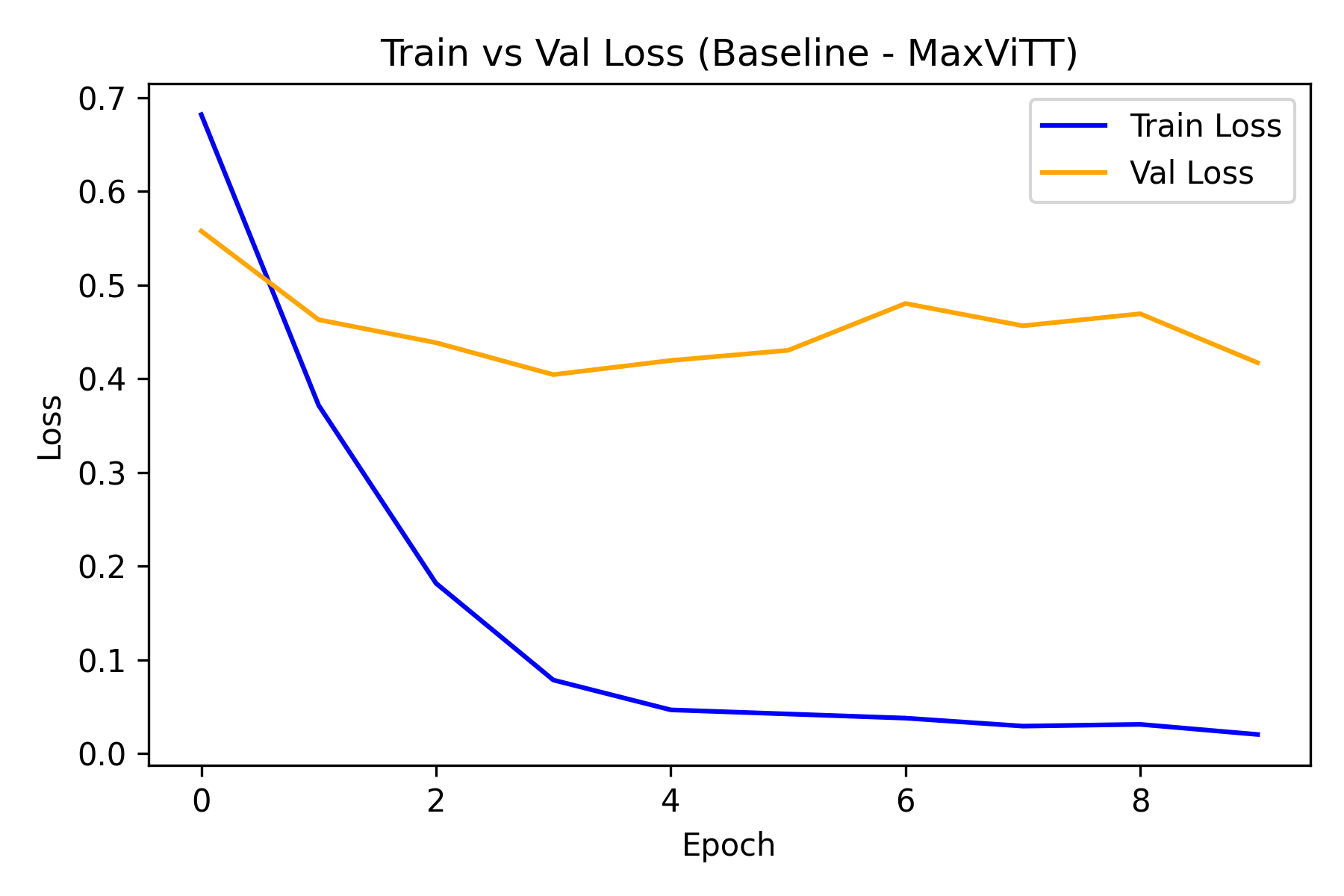}
        \caption{MaxViT-T Baseline - Loss}
    \end{subfigure}
    \hfill
    \begin{subfigure}[t]{0.24\textwidth}
        \centering
        \includegraphics[width=\textwidth]{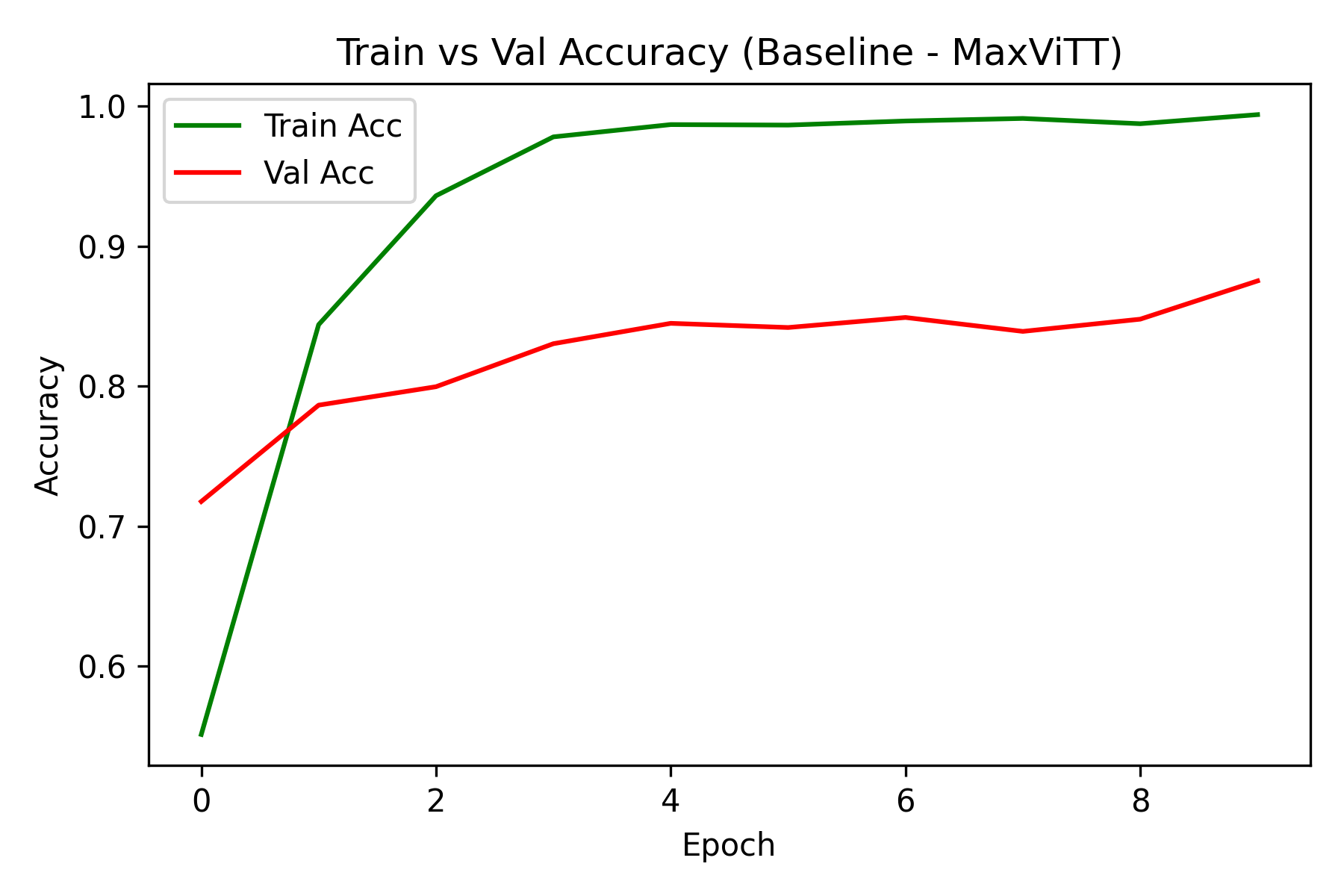}
        \caption{MaxViT-T Baseline - Accuracy}
    \end{subfigure}
    \hfill
    \begin{subfigure}[t]{0.24\textwidth}
        \centering
        \includegraphics[width=\textwidth]{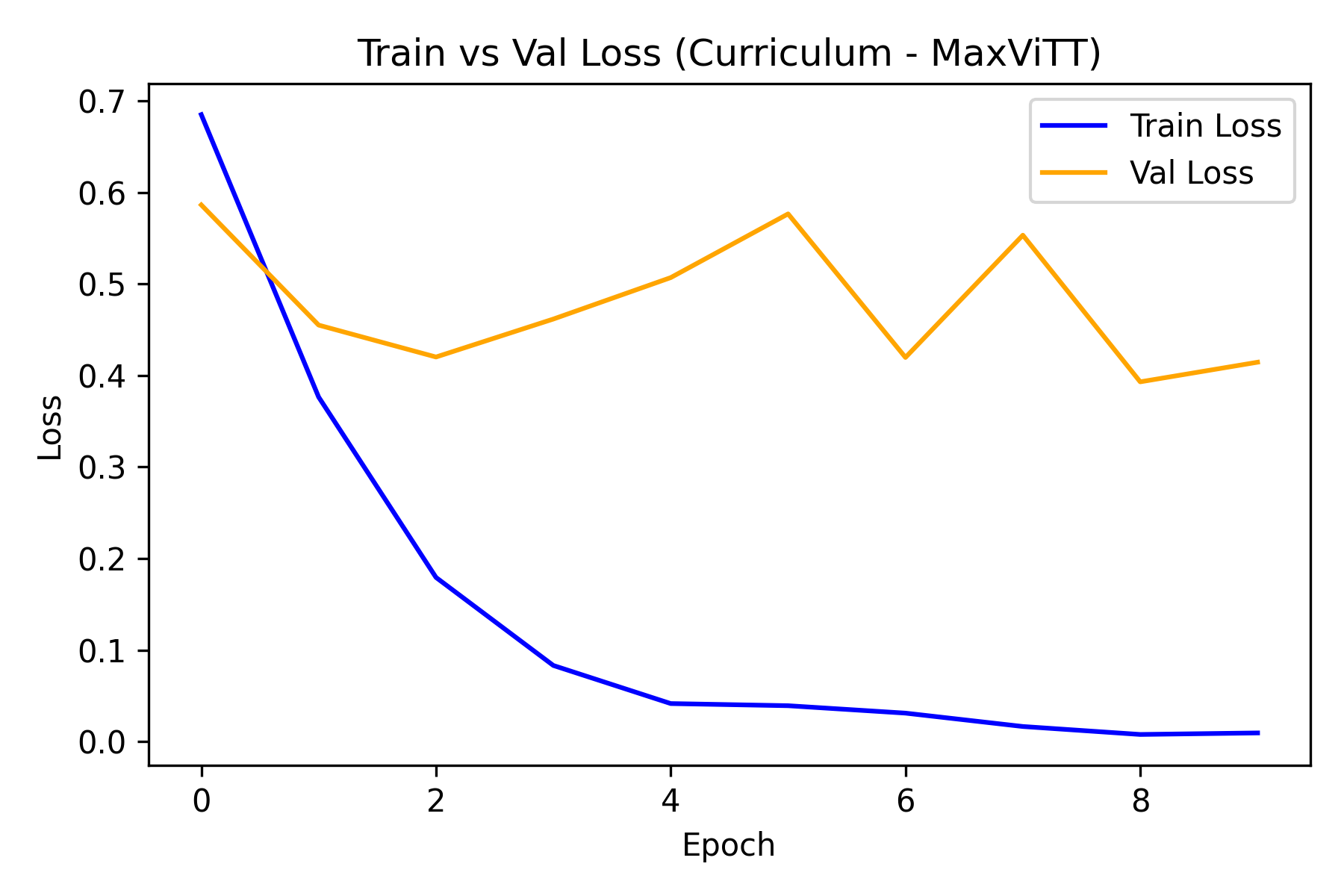}
        \caption{MaxViT-T Curriculum - Loss}
    \end{subfigure}
    \hfill
    \begin{subfigure}[t]{0.24\textwidth}
        \centering
        \includegraphics[width=\textwidth]{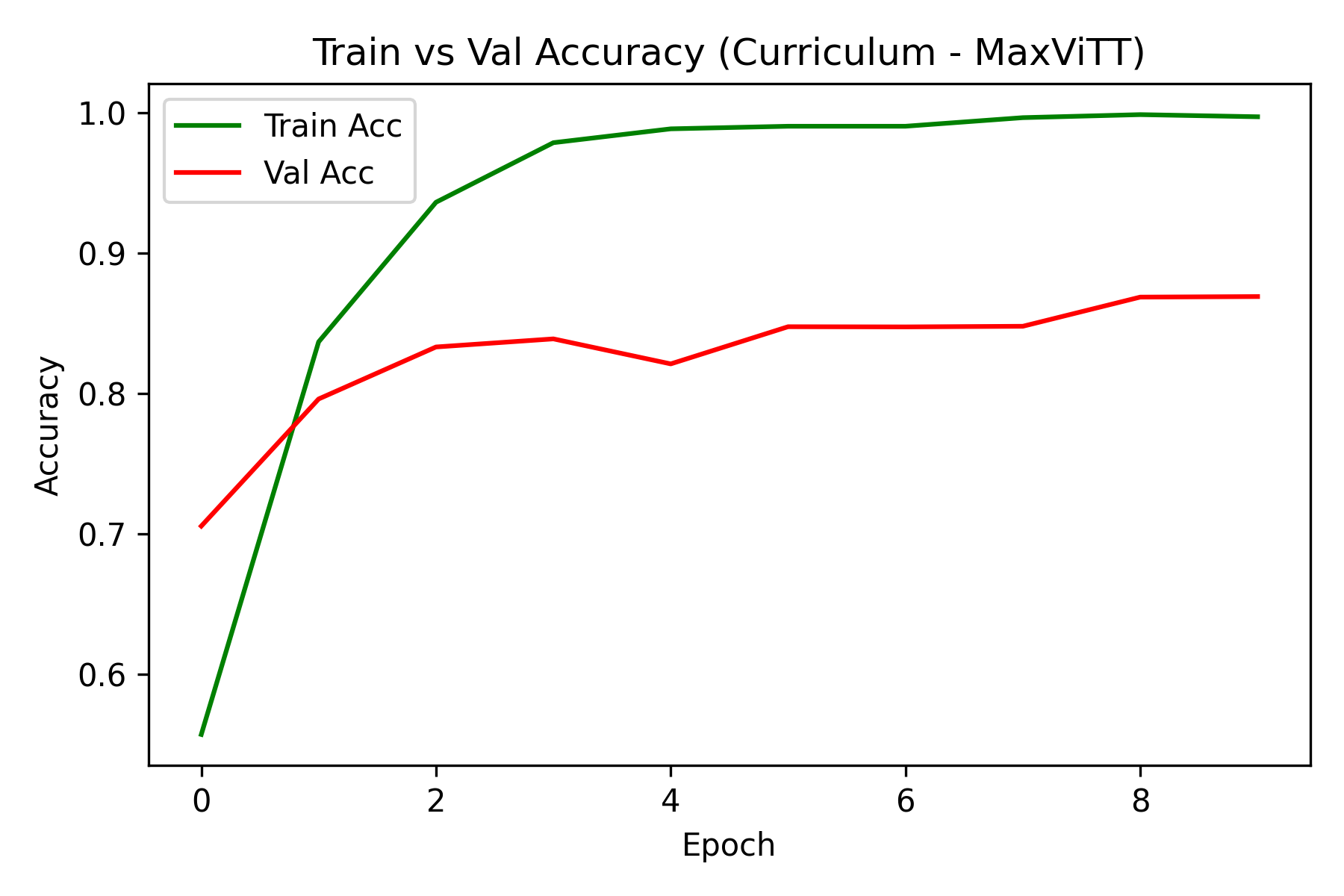}
        \caption{MaxViT-T Curriculum - Accuracy}
    \end{subfigure}
    \caption*{\textbf{Supplementary Figure S1B.} Training curves (Part 2 of 2).}
    
\captionof{figure}{\textbf{Supplementary Figure S1. Training curves under baseline and curriculum settings for selected models (excluding DenseNet121 and ConvNeXt-T shown in the main paper).} (S1A. a--d) MobileNetV2, (S1A. e--h) SqueezeNet1.1, (S1A. i--l) ResNet18, (S1A. m--p) VGG16\_bn, (S1A. q--t) EfficientNet-B3. (S1B. a--d) EfficientNet-B4, (S1B. e--h) ViT-B/16, (S1B. i--l) Swin-T, (S1B. m--p) DeiT-S, (S1B. q--t) MaxViT-T.}
\end{figure}

\clearpage
\subsection*{A.4. Statistical significance of FOSSIL learning effects}
\label{appendix:stats}

\begin{sidewaystable}[p]
\centering
\footnotesize
\caption{
Statistical comparison of FOSSIL versus FROZEN (FOSSIL + SAM for DeiT-S and MaxViT-T) across six models 
using paired t-tests and Wilcoxon signed-rank tests. 
Values are p-values; $p<0.05$ indicates statistical significance.
}
\label{tab:stat_test_fossil_frozen}
\renewcommand{\arraystretch}{0.95}
\begin{tabular}{lcccccc}
\toprule
\textbf{Metric} & \textbf{ConvNeXt-T} & \textbf{DenseNet121} & \textbf{DeiT-S\footnotemark[1]} & \textbf{MaxViT-T\footnotemark[1]} & \textbf{VGG16\_bn} & \textbf{MobileNetV2} \\
\midrule
\multicolumn{7}{l}{\textit{Paired t-test p-values}} \\
AUC           & $<$0.0001 & $<$0.0001 & $<$0.0001 & $<$0.0001 & 0.0212 & $<$0.0001 \\
Accuracy      & 0.0040 & $<$0.0001 & $<$0.0001 & $<$0.0001 & 0.3431 & $<$0.0001 \\
Sensitivity   & 0.0001 & $<$0.0001 & 0.0003 & 0.0542 & 0.0358 & 0.0001 \\
Specificity   & 0.1506 & 0.0285 & 0.0005 & 0.0142 & 0.6897 & 0.2192 \\
F1 Score      & 0.0013 & $<$0.0001 & $<$0.0001 & 0.0007 & 0.0946 & $<$0.0001 \\
PPV           & 0.7005 & 0.0002 & $<$0.0001 & 0.0032 & 0.9750 & 0.0399 \\
NPV           & $<$0.0001 & $<$0.0001 & $<$0.0001 & 0.0214 & 0.0261 & $<$0.0001 \\
ECE           & $<$0.0001 & $<$0.0001 & $<$0.0001 & $<$0.0001 & 0.3212 & $<$0.0001 \\
\midrule
\multicolumn{7}{l}{\textit{Wilcoxon signed-rank p-values}} \\
AUC           & 0.0007 & 0.0001 & 0.0001 & 0.0001 & 0.0355 & 0.0001 \\
Accuracy      & 0.0063 & 0.0007 & 0.0010 & 0.0001 & 0.4415 & 0.0010 \\
Sensitivity   & 0.0014 & 0.0008 & 0.0004 & 0.1240 & 0.0457 & 0.0017 \\
Specificity   & 0.1603 & 0.0142 & 0.0018 & 0.0130 & 0.9370 & 0.2603 \\
F1 Score      & 0.0006 & 0.0001 & 0.0001 & 0.0001 & 0.0962 & 0.0010 \\
PPV           & 0.9250 & 0.0009 & 0.0001 & 0.0006 & 0.9721 & 0.0480 \\
NPV           & 0.0002 & 0.0001 & 0.0001 & 0.0151 & 0.0258 & 0.0012 \\
ECE           & 0.0001 & 0.0001 & 0.0001 & 0.0002 & 0.0833 & 0.0001 \\
\bottomrule
\end{tabular}

\footnotetext[1]{DeiT-S and MaxViT-T were trained using FOSSIL and FOSSIL + SAM optimizers, respectively. 
No backbone freezing was applied; the ``FROZEN'' condition corresponds to the SAM setting.}
\end{sidewaystable}

\vspace{1em}
\noindent
Table~\ref{tab:stat_test_fossil_frozen} presents the statistical validation of FOSSIL’s learning effects 
across six representative architectures. Both the paired \textit{t}-test and Wilcoxon signed-rank test 
demonstrate that FOSSIL consistently yields statistically significant improvements over the FROZEN baseline 
($p<0.05$) in most metrics, particularly for AUC, F1 score, and calibration (ECE). 
CNN-based models such as ConvNeXt-T, DenseNet121, and MobileNetV2 
showed the strongest and most consistent gains ($p<0.0001$ across nearly all key metrics), 
confirming the robustness of FOSSIL’s sample-weighting under low-data and imbalanced settings.

For transformer-based architectures (DeiT-S and MaxViT-T), the differences were less pronounced, 
reflecting possible optimizer interaction effects when combined with Sharpness-Aware Minimization (SAM). 
Nonetheless, even in these cases, improvements in discrimination (AUC) and reliability (ECE) remained significant. 
Specificity and PPV exhibited weaker significance levels, suggesting that FOSSIL primarily improves 
recall-oriented behavior (e.g., sensitivity and NPV) by emphasizing harder or ambiguous examples during training.

Overall, these results statistically substantiate that the FOSSIL framework 
achieves consistent and meaningful gains in both discrimination and calibration performance. 
The combination of significant $p$-values across metrics and architectures 
demonstrates that FOSSIL’s benefits are not model-specific but rather represent 
a generalizable learning improvement mechanism applicable to diverse deep networks in medical imaging contexts.


\subsection*{A.5. Theoretical Note on FOSSIL Weightings}
\label{appendix:theory}

\renewcommand{\theequation}{A.\arabic{equation}}
\setcounter{equation}{0}

This appendix summarizes the theoretical properties of the 
FOSSIL (\textbf{F}lexible \textbf{O}ptimization via \textbf{S}ample 
\textbf{S}ensitive \textbf{I}mportance \textbf{L}earning) weighting framework, 
originally introduced and analyzed in Cha et al.~\cite{cha2025fossil}. 
FOSSIL provides a regret-minimizing weighting rule designed for robust learning 
under small and imbalanced data regimes, with formal guarantees on boundedness, 
monotonicity, and sublinear regret.

\paragraph{Setup.}
Let $\ell_t(w_t; x_i, y_i)$ denote the instantaneous convex loss at iteration $t$
for sample $i$, with gradient $g_t = \nabla_w \ell_t(w_t; x_i, y_i)$. 
The model parameters are updated with a weighted step size $\eta_t w_i$, where
\begin{equation}
    w_i = \exp\!\left(-\frac{d_i}{T}\right),
    \label{eq:A1_weight}
\end{equation}
and $d_i \in [0,1]$ represents sample difficulty while $T>0$ is a temperature parameter 
controlling the smoothness of the weighting distribution. 
Lower $T$ values prioritize easier samples during early iterations, while higher $T$ 
values yield a flatter weighting landscape.

\paragraph{Lemma A.5.1 (Boundedness).}
For all samples $i$ and temperature $T>0$,
\begin{equation}
    0 < w_i \le 1.
\end{equation}
\textit{Proof.} Since $d_i \ge 0$, $\exp(-d_i/T)$ lies strictly between 0 and 1. 
Each weight is therefore positive and bounded, ensuring numerical stability 
and preventing gradient explosion. \qed

\paragraph{Lemma A.52 (Monotonicity).}
If $T_t$ is non-increasing over iterations $t$, then for fixed $d_i$,
\begin{equation}
    w_i(t+1) \ge w_i(t).
\end{equation}
\textit{Proof.} Because $\partial w_i/\partial T = (d_i/T^2)\exp(-d_i/T) > 0$, 
reducing $T$ monotonically increases $w_i$. 
Hence, as training progresses and $T_t$ decays, harder samples gradually receive 
higher weight, creating a self-paced weighting schedule. \qed

\paragraph{Theorem A.5.1 (Sublinear Regret Bound).}
Assume $\ell_t(\cdot)$ is convex and $G$-Lipschitz continuous, 
and let $D$ denote the diameter of the feasible parameter space $\mathcal{W}$. 
If FOSSIL weights are applied in the update rule
\[
w_{t+1} = \Pi_{\mathcal{W}}\!\left(w_t - \eta_t w_i g_t\right),
\]
where $\Pi_{\mathcal{W}}(\cdot)$ denotes Euclidean projection, 
then the expected cumulative regret satisfies
\begin{equation}
    \mathbb{E}[R_T] 
    = \mathbb{E}\!\left[\sum_{t=1}^{T} 
        \ell_t(w_t) - \ell_t(w^*)\right]
    \le \frac{D^2}{2\eta_T} + 
    \frac{\eta_T G^2}{2}\sum_{t=1}^{T} w_i^2
    = \mathcal{O}(\sqrt{T}).
    \label{eq:A3_regret}
\end{equation}

\textit{Proof sketch.}
Following the standard online convex optimization framework 
of Hazan~\cite{hazan2016introduction} and Shalev-Shwartz~\cite{shalev2012online}, 
the cumulative regret bound 
$R_T \le D^2/(2\eta_T) + (\eta_T/2)\sum_t \|w_i g_t\|^2$ 
is minimized when the weights follow a Boltzmann distribution 
$w_i \propto \exp(-d_i/T)$. 
This exponential form ensures bounded gradients 
and smooth temporal decay, yielding sublinear cumulative regret 
$\mathbb{E}[R_T] = \mathcal{O}(\sqrt{T})$. 
Consequently, the weighted empirical risk converges to 
the optimal loss $\ell(w^*)$ with provable stability. \qed

\paragraph{Corollary A.5.1 (Smooth Convergence and Stability).}
Under mild convexity and bounded-gradient assumptions, 
FOSSIL achieves the same asymptotic convergence rate as mirror descent 
but with reduced gradient variance due to exponential damping in $w_i$. 
This provides theoretical grounding for the empirical robustness observed 
in the Mpox classification experiments (Section~\ref{sec:Results}).

\end{document}